\RequirePackage{amsthm}
\documentclass[pdflatex,sn-mathphys-num]{sn-jnl}

\usepackage{graphicx}%
\usepackage{multirow}%
\usepackage{amsmath,amssymb,amsfonts}%
\usepackage{mathrsfs}%
\usepackage[title]{appendix}%
\usepackage{xcolor}%
\usepackage{textcomp}%
\usepackage{manyfoot}%
\usepackage{booktabs}%
\usepackage{algorithm}%
\usepackage{algorithmicx}%
\usepackage{algpseudocode}%
\usepackage{listings}%

\usepackage{fix-cm}
\usepackage{mathrsfs}
\usepackage{calrsfs}

\usepackage{subfloat} 
\usepackage{tablefootnote}
\usepackage{xcolor,colortbl}
\usepackage{wrapfig}
\usepackage{xspace}
\newcommand{\Tref}[1]{Table~\ref{#1}}
\newcommand{\eref}[1]{Eq.~\eqref{#1}}

\newcommand{\fref}[1]{Fig.~\ref{#1}}
\newcommand{\Fref}[1]{Figure~\ref{#1}}

\newcommand{\ie}{\textit{i.e.}}
\newcommand{\eg}{\textit{e.g.}}
\newcommand{\etal}{\textit{et al.}}

\usepackage{xcolor}
\newcounter{todos}
\AtEndDocument{\ifnum\value{todos}>0 \PackageWarning{TODOS}{There are \arabic{todos} todos left in this paper! Fix them before submitting the paper!} \fi}

\newcommand{\V}[1]{\ensuremath{\mathbf{#1}}}

\usepackage{bm}

\usepackage{subfig}
\usepackage{bm}
\usepackage{tabularx} 

\newcommand{\constraint}{\ensuremath{{E_{tree}}}}
\newcommand{\func}{\ensuremath{\mathcal{F}}}

\newcommand{\Proj}{\ensuremath{\mathcal{P}}}
\newcommand{\Rep}{\ensuremath{\mathcal{R}}}
\newcommand{\Loss}{\ensuremath{\mathcal{L}}}

\newcommand{\img}{\ensuremath{I}}
\newcommand{\graph}{\ensuremath{{G}}}
\newcommand{\ijsub}{\ensuremath{(i,j)}}

\newcommand{\softmax}{\ensuremath{\V{\sigma}}}

\newcommand{\edge}{\ensuremath{{E}}}
\newcommand{\vertex}{\ensuremath{{V}}}
\newcommand{\REP}{SFS\xspace}

\newcommand{\real}{\mathbb{R}}
\newcommand{\unconst}[1]{\ensuremath{\hat{#1}}}

\newcommand{\bll}[1]{\cellcolor{red!40}#1} 
\newcommand{\sll}[1]{\cellcolor{orange!40}#1} 
\newcommand{\wll}[1]{\cellcolor{white}#1} 

\PassOptionsToPackage{pagebackref,breaklinks,colorlinks}{hyperref}
\usepackage{hyperref}

\usepackage[capitalize]{cleveref}
\crefname{section}{Sec.}{Secs.}
\Crefname{section}{Section}{Sections}
\Crefname{table}{Table}{Tables}
\crefname{table}{Tab.}{Tabs.}

\theoremstyle{thmstyleone}%

\theoremstyle{thmstyletwo}%

\theoremstyle{thmstylethree}%

\begin{document}

\title[PlantPose]{PlantPose: Universal Plant Skeleton Estimation via Tree-constrained Graph Generation}

\author[1]{\fnm{Xinpeng} \sur{Liu}}\email{liu.xinpeng@ist.osaka-u.ac.jp}
\author[1]{\fnm{Hiroaki} \sur{Santo}}\email{santo.hiroaki@ist.osaka-u.ac.jp}
\author[2,3]{\fnm{Yosuke} \sur{Toda}}\email{yosuke@phytometrics.jp}
\author*[1]{\fnm{Fumio} \sur{Okura}}\email{okura@ist.osaka-u.ac.jp}

\affil*[1]{\orgdiv{Graduate School of Information Science and Technology}, \orgname{Osaka University}, \orgaddress{\street{Suita}, \city{Osaka}, \postcode{5650871}, \country{Japan}}}
\affil[2]{\orgname{Phytometrics}, \orgaddress{\street{Hamamatsu}, \city{Shizuoka}, \postcode{4350036}, \country{Japan}}}
\affil[3]{\orgdiv{Institute of Transformative Bio-Molecules}, \orgname{Nagoya University}, \orgaddress{\street{Nagoya}, \city{Aichi}, \postcode{4648601}, \country{Japan}}}

\abstract{Accurate estimation of plant skeletal structures (\eg, branching structures) from images is essential for smart agriculture and plant science. Unlike human skeletons with fixed topology, plant skeleton estimation presents a unique challenge, \ie, estimating arbitrary tree graphs from images. To address this problem, we introduce \textbf{PlantPose}, a universal plant skeleton estimator via tree-constrained graph generation. PlantPose combines learning-based graph generation with traditional graph algorithms to enforce tree constraints during the training loop. To enhance the model’s generalization capability, we curate a large and diverse dataset comprising real-world and synthetic plant images, along with simplified representations (\eg, sketches and abstract drawings). This dataset enables the generalized model to adapt to diverse input styles and categories of plant images while preserving topological consistency. Our approach demonstrates robust and accurate plant skeleton estimation across multiple domains, including previously unseen out-of-domain scenarios. Further analyses highlight the method's strengths and limitations in handling complex, heterogeneous data distributions. All implementations and datasets are available at \url{https://github.com/huntorochi/PlantPose/}.}


\maketitle

\begin{figure}[t]
    \centering
    \includegraphics[width=\linewidth]{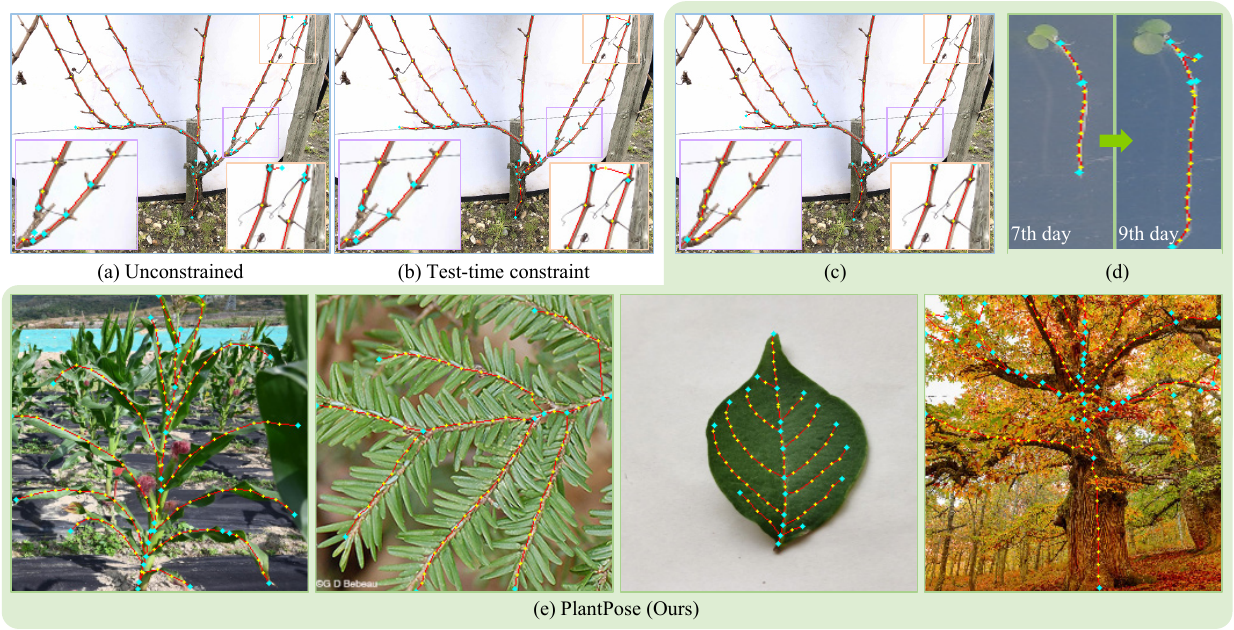}  
    \caption{We propose \textbf{PlantPose}, a method for single-image plant skeleton estimation combining learning-based graph generators with traditional graph algorithms (\ie, MST). The red lines show the predicted graph edges. Compared to (a) an \emph{unconstrained} graph generator and (b) a naive tree-graph constraint implementation, (c) our method naturally imposes the constraint during the graph generation models' training. Our method can be directly applied to plant science and agricultural applications, such as (d) time-series reconstruction of botanical roots. Furthermore, the PlantPose with large-scale training demonstrates strong generalization capabilities across diverse datasets, including significantly different out-of-domain plant species, as shown in (e).}
    \label{fig:intro}
\end{figure}

\section{Introduction}
\label{sec:intro}

Skeletal structures of plants (\eg, branches and roots) are key information for analyzing plant traits in agriculture and plant science.
In particular, single-view estimation of plant skeletons has potential benefits for various downstream tasks, such as high-throughput plant phenotyping~\cite{Guyot, skeleton3dMulitRUE, skeleton3dMaizePoint} and plant organ segmentation~\cite{skeletonSeg, 3dpointSeg}. 
As a similar task, single-view estimation of human poses has been widely studied, \eg, OpenPose~\cite{OpenPose}. However, unlike human skeletons, which have a fixed graph topology, the plant skeleton is not organized because the number of joints and their relationships are unknown, posing a unique problem of estimating an arbitrary tree graph from an image.

The estimation of graph structure from images has been studied to extract thin structures such as road networks in satellite images~\cite{Sat2Graph, SOTAWorkRNGDet, iCurbIL}.
Recent end-to-end models using recurrent neural networks (RNNs)~\cite{polygon-rnn++}, graph neural networks (GNNs)~\cite{StrucLandDetect, PolyBuildSeg, CurveGCN}, or transformers~\cite{GGT, Image2SMILES, SGTR, PolyTransform, Relationformer} show the ability to extract faithful \emph{unconstrained} graph structures from images. 
However, inferring tree-constrained graphs with the existing graph generators becomes a non-trivial problem, where the output graph often violates the required constraints, as shown in \fref{fig:intro}a.
One reason for this difficulty is that tree graph generation, which requires finding a set of graph edges that satisfy the constraint defined on the entire graph, naturally falls into combinatorial optimization. Although constrained optimization in neural networks has been studied by designing differentiable layers~\cite{kotary2021end}, using them to solve our problem is still not straightforward.
A simple way to impose constraints on the graph generation is to convert the inferred unconstrained graphs to the closest graph that satisfies the given constraint using traditional graph algorithms, such as Dijkstra's shortest path or minimum spanning tree (MST) algorithms.
Such post-processing can work; however, because the graph generators are trained without any constraints, the output may be unrealistic, as shown in \fref{fig:intro}b.

For tree graph generation from single images, we propose a simple yet effective way to integrate state-of-the-art learning-based graph generation methods, which achieve high-quality image-based graph estimation, and traditional graph algorithms, which compute strictly constrained tree graphs such as shown in Figs.~\ref{fig:intro}c--d. 
Specifically, we propose to \emph{project} an unconstrained graph into a tree graph by a non-differentiable MST algorithm during each training loop. 
Our selective feature suppression (SFS) layer then converts the inferred unconstrained graph to the MST-based tree graph in a differentiable manner, thereby naturally incorporating the constraints into the graph generation. 

By integrating our feature suppression layer with a state-of-the-art transformer-based graph generator, we develop \textbf{PlantPose}\footnote{This terminology is, of course, an analogy of OpenPose~\cite{OpenPose} for plants.}, which infers tree structures from images capturing plants. Beyond the majority of plant-oriented vision methods that use domain-specific (\ie, species-specific) datasets for the training, we achieve a generalized model that estimates the plant skeletons for varieties of plants, as highlighted in \fref{fig:intro}e. To this goal, we curate or collect $10$ datasets spanning real-world, synthetic, and web-sourced plant images, ensuring the model is exposed to various structural patterns and contexts. 

We evaluate the effectiveness of PlantPose on different classes of plant images, including out-of-domain datasets that are significantly different from those used for the training. The results show that our constraint-aware graph generator accurately estimates the target tree structures compared to baselines. 

This paper is an extended version of a conference paper~\cite{treeformer} presented in WACV 2025. Beyond \cite{treeformer} using the domain-specific training for each plant species, we achieve a significant advancement by focusing on the universal applicability of our method through the large-scale dataset creation and training, resulting in the plant skeleton estimation across various plant species by a single pre-trained model, as shown in \fref{fig:intro}e.  

\paragraph*{Contributions} 
Our contributions are threefold: 
\begin{itemize}
    \item We propose a novel method that tightly integrates learning-based graph generation methods with traditional graph algorithms using the newly proposed SFS layer, which modifies intermediate features in the network, effectively mimicking the behavior of the non-differentiable graph algorithms. 
    \item Building upon our constrained graph generation method, we develop the first end-to-end method inferring skeletal structures from a single plant image, which benefits the agriculture and plant science fields. 
    \item We collect or curate plant skeleton datasets, including diverse plant images and skeletons, contributing to the generalization of our method as well as future studies for plant skeleton analysis. 
\end{itemize}

\section{Related Work}
We propose constraining the graph structures given by image-based graph generators, whose primary goal is plant skeleton estimation.
We, therefore, introduce the related work of plant skeleton estimation, graph generation from images, and constrained optimization for neural networks.

\subsection{Plant skeleton estimation}
Plant skeleton estimation is actively studied since it becomes a fundamental technique for downstream tasks related to plant phenotyping and cultivation~\cite{Okura3D}.

\paragraph*{3D plant skeleton estimation}
Several methods are proposed to derive plant skeletons from 3D observations~\cite{Okura3D}. 
These methods often use point clouds acquired by LiDAR~\cite{skeleton3dTreeLiDAR,skeletonTreeLaserOld} or multi-view stereo (MVS)~\cite{MultiStereo3dPoint,skeleton3dMaizePoint}. Regardless of the 3D acquisition method, these works generally use a two-stage pipeline: Skeletonization~\cite{bucksch2014practical} followed by graph optimization using MST or Dijkstra's algorithm~\cite{AdTree,L1point,3dpointSeg,PointCloudSeg,StochasticSkeleton}, where the graph algorithms are required to convert a set of skeleton positions into a graph.

\paragraph*{2D plant skeleton estimation}
Compared to 3D methods, skeleton estimation from a single 2D image poses significant technical challenges due to the lack of depth information and severe occlusions despite the simplicity of data acquisition. Like 3D methods, existing 2D methods use a two-stage process involving skeletonization and graph optimization. To extract the skeleton regions on 2D images, plant region segmentation is often used for plants with relatively thin leaves~\cite{CenterDetection2D}.
Similarly, a neural network that converts an input image into a map representing 2D skeleton positions is used to mitigate the occlusions~\cite{isokane2018probabilistic}. To reason about the direction of intersecting branches, a recent work~\cite{Guyot} proposes to use vector fields representing branch direction instead of mask images, similar to the Part Affinity Fields (PAFs) used in OpenPose~\cite{OpenPose}. 

Unlike existing two-stage methods, we propose an end-to-end method that directly infers a tree graph representing plant skeletons in a single image.
Our experiments show that our end-to-end method achieves better accuracy than a recent two-stage method for 2D images.

\subsection{Graph generation from images}

Graph generation from images, sometimes called image-to-graph generation, is studied for extracting geometric structures (\eg, road networks) or relations (\eg, scene graphs) from images~\cite{polygon-rnn++, GGT, Image2SMILES, SGTR, StrucLandDetect, PolyBuildSeg, PolyTransform, CurveGCN, Relationformer, Sat2Graph, SOTAWorkRNGDet, iCurbIL, RelTR_DETR}. 

\paragraph{Geometric image-to-graph generation}
Recent learning-based methods for extracting geometric structures often use object detectors, which detect graph nodes (\eg, intersections in road networks) from images, and then aggregate the combinations of node features to predict the edges defined between two nodes as binary (\ie, existence of edges) or categorical (\eg, classification of edge relations) values. Some studies use external knowledge~\cite{2longtail,2recovering, Improving, glove} to improve the results. 

Graph generators have usually taken autoregressive methods (\eg,~\cite{Molecule_auto_father, Molecule_auto1, Molecule_auto2, GGT, deeptree}) that output a graph by starting at an initial node and estimating neighboring graph nodes. 
Recent GNNs and transformers enable non-autoregressive graph generators~\cite{kingma2013auto, GraphVAE, PSG_query_matching, SGTR, MoFlow, RelTR_DETR} simultaneously estimating the entire graph. Autoregressive methods are prone to errors during the estimation process, and the state-of-the-art non-autoregressive method, RelationFormer~\cite{Relationformer}, performs better than autoregressive methods, especially for medium to large graphs.

A few recent studies consider graph generation with tree-graph constraints in a different context.
For the molecule structure estimation~\cite{SpanningTreeMolecules, TreeMolecules_father, TreeMolecules_1}, these methods assume autoregressive graph generation, making it hard to work with complex and relatively large graphs like botanical plants. 

\paragraph{Scene graph generation}
Scene graph generation (SGG) represents another line of image-to-graph research, focusing on semantic relations between detected objects. Milestone works such as IMP~\cite{SGG_IMP} and VCTree~\cite{SGG_VCTree} rely on region-of-interest (RoI) features from bounding boxes and union regions, while Lang-SGG~\cite{SGG_Lang} augments RoI features with geometric cues and language-derived word embeddings. More recently, VSS~\cite{SGG_Lang_VSS} leverages vision–language alignment and text prompts via GLIP~\cite{GLIP} to obtain object nodes. Unlike SGG, our task deals with geometric graphs: Nodes represent junctions or endpoints defined by coordinates, and edges are thin connections between node pairs. 
SGG methods fundamentally assume region-based objects and semantic predicates; although we can technically use SGG methods on the extracted node positions in our case, the reasoning modules degrade into exhaustive pairwise classification with contextual encoders or become unusable without text prompts. In practice, the pipelines are reduced to those akin to RelationFormer with pairwise reasoning, but with heavier computation or additional semantic dependencies.

\subsection{Constrained optimization for neural networks}
Constrained optimization is crucial for machine learning. In particular, introducing constraints in neural networks has become a recent trend~\cite{kotary2021end}. 

\paragraph*{Designing differentiable layers}
The most direct way to introduce additional constraints to neural networks is to make the constraints differentiable.
In the continuous domain, it is known that a convex optimization can be implemented as a differentiable layer~\cite{agrawal2019differentiable}. 
However, the design of differentiable layers for combinatorial optimization poses a significant challenge due to the difficulty of differentiation. 
Wilder~\etal~\cite{wilder2019melding} propose a differentiable layer for linear programming (LP) problems using continuous relaxation. This method is extended to mixed integer linear programming (MILP)~\cite{ferber2020mipaal} by splitting the problem into multiple LPs. 
MST, which we want to use as constraints, is known to be transformed into the class of MILP~\cite{myung1995generalized,pop2020generalized}. However, using differentiable layers for these complex combinatorial problems requires exponential computation time~\cite{kotary2021end} to obtain the exact solution and is practically unrealistic.

\paragraph*{Reparameterization for constrained optimization}
If the constraint function is difficult to differentiate, a simple alternative is to \emph{project} unconstrained inferences or model parameters into constrained space, which can be considered a use of reparameterization~\cite{kingma2013auto}. 

In gradient descent optimization, methods projecting unconstrained optimization parameters (\eg, model parameters in neural networks) to the closest ones satisfying the given constraint are called projected gradient descent (PGD). PGD is often used for traditional optimization problems, directly optimizing the input variables~\cite{PGD-old, PGD-CT}. While PGD can be used for neural network optimization, such as for generating adversarial examples~\cite{madry2018towards}, designing projection functions for neural networks is challenging. It requires mapping a large number of model parameters into a space satisfying complex constraints, where the constraints are often more naturally defined on the model output.

Instead of designing a projection function for the model parameters, the model's output can be projected onto the subspace that satisfies the constraints during the training loop. 
Since reparameterization for model output can be easily integrated with existing neural network models, they are often used for domain-specific applications such as coded aperture optimization with hardware constraint~\cite{CameraConstraints} and internal organ segmentation with given parametric shape models~\cite{SurveyMedicalConstraints, MedicalSegmentationConstraints}.
We take this approach in our SFS layer, easily plugging it into off-the-shelf end-to-end graph generation methods without preparing the differentiable implementation of the constraints (\ie, MST algorithm).

\section{Tree-constrained Graph Generation} 
\begin{figure*}[t]
	\centering
	\includegraphics[width=\linewidth]{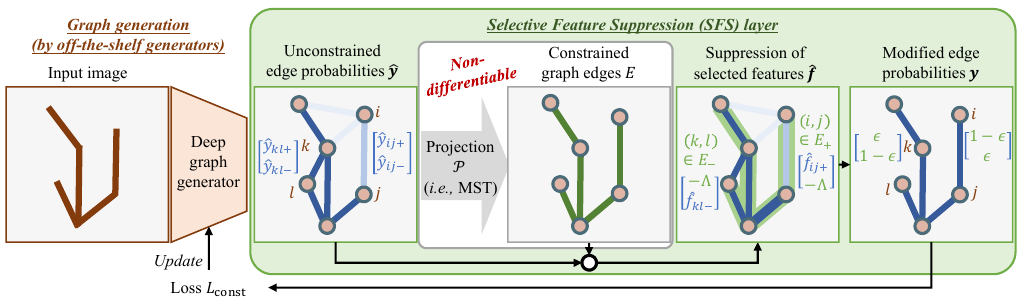}
	\caption{Overview of reparameterization layer that can be easily plugged into off-the-shelf graph generators. Given unconstrained edge predictions by graph generators, our method \emph{projects} it to the closest constrained graph (\ie, tree) using a non-differentiable MST algorithm. Comparing constrained and unconstrained edges, unwanted edge features are selectively suppressed so that the graph becomes a tree.}
	\label{fig:overview}
\end{figure*}

We here describe the constrained graph generation method. \Fref{fig:overview} summarizes the proposed \REP layer, which casts the original unconstrained edge probabilities to the constrained domain.

\subsection{Problem statement}
We here consider a simple setting of neural-network-based graph generation, where the model outputs the prediction of the edge probabilities (\ie, the edge exists or not) defined for a pair of nodes, while this can be extended to a multi-class classification setting straightforwardly.

Our goal is to design a tree-constrained graph generator $\func$ that converts a given image $\img$ to a tree graph $\graph$ as
\begin{equation}
\label{eq:graph_const}
\graph = (\vertex,\edge) = \func(\img) \quad \mathrm{s.t.} \quad \edge \in \constraint,
\end{equation}
where the graph $\graph$ consists of a set of nodes (or objects) $\vertex$ and edges (or relations) $\edge$. Here, $\constraint$ denotes all possible edge patterns forming a tree graph given the set of nodes $\vertex$.

We consider a (non-differentiable) projection function $\Proj$ that maps an unconstrained graph predicted by graph generators to the constrained graph $G$. Let the edge probabilities defined between each node pairs as $\{\unconst{\V{y}}_{\ijsub}\}_{(i,j)\in{V\times V}}$.
\begin{equation}
\label{eq:probs}
\unconst{\V{y}}_{\ijsub}=[\unconst{y}^{+}_{\ijsub}, \unconst{y}^{-}_{\ijsub}]^\top \quad \mathrm{s.t.} \quad \|\unconst{\V{y}}_{\ijsub}\|_1=1,   
\end{equation}
in which $\unconst{y}^{+}_{\ijsub}$ and $\unconst{y}^{-}_{\ijsub}$ respectively denote the edge existence and non-existence probabilities.
The projection function $\Proj$ then read as
\begin{equation}
(\vertex,\edge) = \Proj_{\edge \in \constraint}(\vertex,\{\unconst{\V{y}}_{\ijsub}\}),
\label{eq:P_I2G}
\end{equation}
which are given by traditional graph algorithms with combinatorial optimization.
We assume the projection function $\Proj$ converts the existence probability (or category prediction) of graph edges while leaving the graph nodes $V$ unchanged.
A typical example of $\Proj$ can be designed using the MST algorithm we use in our PlantPose implementation, which projects an arbitrary graph into a tree structure by modifying the existence of graph edges based on the costs defined between each pair of nodes.

We aim to develop a differentiable function $\Rep$ that mimics the non-differentiable projection $\Proj$. Plugging with the \emph{unconstrained} graph generator $\unconst{\func}$, \eref{eq:graph_const} is rewritten as
\begin{equation}
\begin{array}{ll}
\graph = (\vertex,\edge) = \Rep_{E \in \constraint} (\vertex,\{\unconst{\V{y}}_{\ijsub}\}), 
\quad (\vertex,\{\unconst{\V{y}}_{\ijsub}\}) = \unconst{\func}(\img),
\end{array}
\label{eq:rep}
\end{equation}
where the whole process is differentiable.

\subsection{SFS layer}
\paragraph{Overview} 
We first briefly summarize the intuition of the SFS layer. Naive neural graph generators predict unconstrained edges between node pairs, which may include cycles or disconnected parts. The MST projection converts these predictions into a valid tree by adding or removing edges. Our key idea is to mimic this projection inside the network in a differentiable way.

We here suppose the multi-head binary classification of edge existence between two given nodes, \ie, the features of the last layer outputs both edge and non-edge features, followed by a softmax layer deciding whether the node pair has an edge.
Our SFS layer \emph{selectively} suppresses the features in the last layer, \ie, for graph edges that should be added, we suppress the non-edge feature so that the softmax activation function outputs the edge; for edges that should be removed, we suppress the edge feature instead. All other edges remain unchanged. This selective suppression ensures that the resulting edge existence matches the tree structure given by the MST, while still allowing gradients to flow back to the backbone network.

\paragraph{Formulations}
Here, we describe a detailed implementation of the SFS layer for constrained graph generation. As described in \eref{eq:probs}, the unconstrained graph generator $\unconst{\func}$ computes the probability of \emph{unconstrained} edge existence between the $i$-th and $j$-th nodes, $\unconst{\V{y}}_{\ijsub}=[\unconst{y}^{+}_{\ijsub}, \unconst{y}^{-}_{\ijsub}]^\top$.
In neural networks, $\unconst{\V{y}}_{\ijsub}$ is usually computed through the softmax activation $\softmax$ applied to the output feature vector of the final layer $\unconst{\V{f}}_{\ijsub}=[\unconst{f}^{+}_{\ijsub},\unconst{f}^{-}_{\ijsub}]^\top\in\real^2$ as
\begin{equation}
\unconst{\V{y}}_{\ijsub} = \softmax(\unconst{\V{f}}_{\ijsub}). 
\label{eq:feat}
\end{equation}
The set of unconstrained graph edges $\unconst{E}$ is then obtained by comparing the edge existence probabilities as
\begin{equation}
\label{eq:thresholding_main}
\unconst{\edge} = \{ (i, j) \mid \unconst{y}^{+}_{\ijsub} > \unconst{y}^{-}_{\ijsub} \},
\end{equation}
in which $\unconst{E}$ records node pairs where the edge exists.

Suppose the projection function $\Proj$ converts the set of unconstrained edge probabilities $\{\unconst{\V{y}}_{\ijsub}\}$ to a set of constrained edges $E$. 
Let the difference of two sets be $E^+ = E - \unconst{E}$ and $E^- = \unconst{E} - E$, denoting the sets of edges newly added and removed by the projection.
To mimic discrete (and non-differentiable) inferences by $\Proj$ in differentiable end-to-end learning, we modify the edge features corresponding to $E^+ \cup E^-$ in the differentiable forward process. 

Specifically, what we want to get is the edge probabilities that approximate the constrained edges $E$, denoted as
\begin{equation}
\label{eq:naive}
\V{y}_{\ijsub} = \left\{
\begin{array}{l@{\hspace{2em}}l}
\left[1-\epsilon,\:\:\:\:\epsilon\:\:\:\: \right]^\top           & ((i,j)\in E^+)\\
\left[\:\:\:\:\epsilon\:\:\:\:,  1-\epsilon \right]^\top           & ((i,j)\in E^-)   \\
\left[\unconst{{y}}^{+}_{\ijsub},  \unconst{{y}}^{-}_{\ijsub} \right]^\top       & \mathrm{(otherwise)}.
\end{array}
\right.
\end{equation}

When $\epsilon$ is small enough, the constrained output $\V{y}_{\ijsub}$ perfectly mimics the output by the projection function $\Proj$. 
However, the direct modification of the edge probabilities naturally disconnects the computation graph. 
Therefore, we modify the unconstrained feature vector $\unconst{\V{f}}_{\ijsub}$ so that the corresponding edge probabilities $\V{y}_{\ijsub}$ follows \eref{eq:naive}. Specifically, since $\V{y}_{\ijsub}$ is computed through the softmax function $\softmax$, it is achieved via the following minimal modification that \emph{selectively} suppresses the feature values by replacing them with a constant\footnote{See the appendices for the derivation.} as
\begin{equation}
\begin{array}{l@{\hspace{2em}}l}
f^{-}_{\ijsub} := -\Lambda & ((i,j) \in E^+) \\
f^{+}_{\ijsub} := -\Lambda & ((i,j) \in E^-),
\end{array}
\label{eq:rep2}
\end{equation}
where $\Lambda$ is assumed to be large enough to make $\exp(-\Lambda) \sim 0$. Given modified features $\V{f}_{\ijsub} = [f^{+}_{\ijsub},f^{-}_{\ijsub}]^\top$, the softmax activation $\softmax$ normalizes and converts them to edge probability $\V{y}_{\ijsub}$. 

In summary, from Eqs.~(\ref{eq:feat}) and (\ref{eq:rep2}), the constrained edge prediction between $i$-th and $j$-th nodes,~\hbox{$\V{y}_{ij} = [y_{ij+},y_{ij-}]^\top$}, is obtained as
\begin{equation}
\label{eq:rep_layer}
\V{y}_{\ijsub} = \left\{
\begin{array}{l@{\hspace{2em}}l}
\softmax( [\unconst{{f}}^{+}_{\ijsub}, \:\:-\Lambda\:]^\top)  \quad  & ((i,j)\in E^+)   \\
\softmax( [\:\:-\Lambda\:,      \unconst{{f}}^{-}_{\ijsub}]^\top)       & ((i,j)\in E^-)    \\
\softmax( [\unconst{{f}}^{+}_{\ijsub},  \unconst{{f}}^{-}_{\ijsub}]^\top)       & \mathrm{(otherwise)}.
\end{array}
\right.
\end{equation}
After the reparameterization, the set of edges computed from $\{\V{y}_{\ijsub}\}$ in the same way as \eref{eq:thresholding_main} is guaranteed to be equal to $E$ inferred by the discrete projection function $\Proj$ when $\Lambda$ is large enough.

\subsection{Analysis}
The common auto differentiation libraries automatically compute the gradient of the SFS layer.
Although it can disconnect the computation path at a feature, since we keep at least one of the original features (either $\unconst{{f}}^{+}_{\ijsub}$ or $\unconst{{f}}^{-}_{\ijsub}$), the backpropagation path to the backbone graph generation network is not disconnected\footnote{This is akin to the dropout layer often used in neural networks.}.
Here, we briefly analyze the behavior of the SFS layer. The appendices provide a detailed analysis, including mathematical proofs. 

When using the cross-entropy loss $\Loss_\text{CE}$ to evaluate the availability of the graph edges, the derivative to be backpropagated to the backbone graph generator is approximated as\footnote{We omit the subscript $\ijsub$ for simplicity.}
\begin{align}
\label{eq:rep_gradient}
\frac{\partial \Loss_\text{CE}}{\partial \unconst{\V{f}}} &\sim
\left\{
\begin{array}{l@{\hspace{2em}}l}
\left[\:\:1-t^{+}\:, \:\:\:\:\:\:0\:\:\:\:\:\:\:\right]^\top & \hspace{-3mm}((i,j)\in E^+)    \\
\left[\:\:\:\:\:\:\:0\:\:\:\:\:\:, \:\:1-t^{-}\:\right]^\top & \hspace{-3mm}((i,j)\in E^-)    \\
\left[y^{+}-t^{+}, y^{-}-t^{-}\right]^\top & \hspace{-3mm}(\mathrm{otherwise}),
\end{array}
\right.
\end{align}
where $\V{t}=[t^{+},t^{-}]^\top\in\{0,1\}^2$ denotes the ground truth edge existence and non-existence for the node pair $\ijsub$. 
Our method modifies the computation graph of the network when the MST algorithm disagrees with the output of graph generation model (\ie, $(i,j) \in E^+\cup E^-$), but in different ways for derivatives of each feature value $\frac{\partial \Loss_\text{CE}}{\partial \unconst{{f^+}}}$ or~$\frac{\partial \Loss_\text{CE}}{\partial \unconst{{f^-}}}$.

Without loss of generality, we consider the case when the MST algorithm \emph{adds} an edge, \ie, $(i,j) \in E^+$. 
When the MST \emph{correctly} modify the edge availability (\ie, $t^+=1$), the gradient vector becomes small, $\frac{\partial \Loss_\text{CE}}{\partial \unconst{\V{f}}} \sim \V{0}$, which is the behavior we expect.
On the other hand, if the MST \emph{incorrectly} adds the edge (\ie, $t^+=0$), the gradient becomes $[1,0]^\top$, which strongly penalizes the positive edge probability, where the norm of the gradient vector is always larger than unconstrained ones\footnote{See appendices for the mathematical proof.}.
Therefore, the behavior of our simple reparameterization strategy is reasonable in practice.

\section{PlantPose: A Plant Skeleton Estimator}
\label{sec:PlantPose}

We develop PlantPose, an implementation of the SFS layer for a state-of-the-art graph generator. 
This section first recaps the graph generator we use~\cite{Relationformer} and then details how we introduce a tree structure constraint.

\subsection{RelationFormer: A brief recap}
RelationFormer~\cite{Relationformer} is the state-of-the-art non-autoregressive graph generation method. 
This method uses an end-to-end architecture that combines an object (node) detector and relation (edge) predictor, which shows superior performance for unconstrained graph generation.
The object detection part is based on deformable DETR~\cite{Deformable_DETR}, which is trained to extract graph nodes (\eg, objects) and global features from a given image. 
Specifically, given the extracted image features, the transformer decoder outputs a fixed number of object queries ([obj]-tokens) representing each of the nodes and a relation query ([rtn]-token) describing the global features, including node relations.

The relation prediction head outputs the relationship (\ie, edge existence or category) from the detected pairs of objects (\ie, [obj]-tokens) and the global relation (\ie, [rtn]-tokens). This module is implemented as a multi-layer perceptron (MLP) headed by layer normalization~\cite{LayerNorm}. 
RelationFormer is trained using the sum of loss functions related to object detection and edge (relation) estimation, where edge (relation) loss $\Loss_\text{edge}$\footnote{Denoted as $\Loss_\text{rln}$ in the original paper~\cite{Relationformer}, we use $\Loss_\text{edge}$ for generality.} evaluates the edge existence or category between node pairs using cross-entropy loss.

\subsection{Tree-constrained graph generation}
To introduce the tree structure constraint, we use Kruskal's MST algorithm~\cite{Kruskal} implemented in NetworkX\footnote{\url{https://networkx.org/}, last accessed on December 20, 2024.}. To extract a tree from an unconstrained graph predicted by RelationFormer, we use the edge non-existence probabilities $\{\unconst{y}^{-}_{\ijsub}\}$ as the edge cost for the MST algorithm to span the tree on edges with higher existence probabilities.

We implement the SFS layer on top of the relation prediction head in the RelationFormer. 
Specifically, the output features from the MLP after layer normalization are regarded as unconstrained features $\{\V{\unconst{f}}\}$.
In our experiments, we use $\Lambda = 10$ during training, where~\hbox{$\exp(-\Lambda) = 4.5\times 10^{-5}$}. We show an ablation study changing $\Lambda$ in the appendices.

\paragraph*{Loss function} Our SFS layer affects the evaluation of the edge loss $L_\text{edge}$ in the graph generator, while the computation of other loss functions, such as for node detection, remains the same as in the original implementation. 
Our implementation uses both loss functions for original (unconstrained) and constrained edges. Denoting the ground-truth edges as $E_\text{GT}=\{(i,j) \mid t^{+}_{\ijsub} > t^{-}_{\ijsub}\}$, where $\V{t}_{\ijsub} = [t^{+}_{\ijsub},t^{-}_{\ijsub}]^\top\in\{0,1\}^2$,
the loss function for edge availability $\Loss_\text{edge}$ is modified as follows
\begin{equation}
\Loss_\text{edge} = \underbrace{\sum_{(i,j)}\Loss_\text{CE}(\unconst{\V{y}}_{\ijsub},\V{t}_{\ijsub})}_{\Loss_\text{unconst}} + \underbrace{\sum_{(i,j)}\Loss_\text{CE}({\V{y}}_{\ijsub},\V{t}_{\ijsub})}_{\Loss_\text{const}},
\end{equation}
where $\Loss_\text{CE}$ denotes the cross-entropy loss.

\section{Datasets}

To validate the effectiveness of our method, we assess the proposed method using two scenarios, namely, domain-specific and generalized training. The former scenario is still useful for domain or species-specific analysis, which is often used in practical applications in plant science and agriculture, and we use one synthetic and two real-world datasets.
Beyond the domain-specific training, we also propose to train our PlantPose model with a large-scale dataset, which comprises six datasets that span various species and attributes. For the validation of the generalized model, we also introduce out-of-domain test datasets sourced from various domains that are significantly different from those used for the training. 

\subsection{Domain-specific datasets}
\begin{figure}[t]
	\centering
        \includegraphics[width=0.8\linewidth]{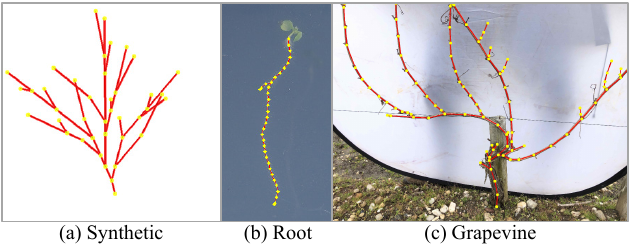} 
	\caption{Example images from the domain-specific dataset. Annotated graphs are superimposed. Yellow dots and red lines indicate nodes and edges.}
	\label{fig:specific_dataset_sample}
\end{figure}

We use one synthetic and two real datasets, where examples are shown in \fref{fig:specific_dataset_sample} and \Tref{tab:domain_dataset}.
Appendices describe the details of the datasets.

\begin{table}[t]
\centering
\renewcommand{\arraystretch}{1.0} 
\caption{Domain-specific datasets. The table summarizes the key properties of each dataset used in our experiments.}
\label{tab:domain_dataset}
\begin{tabularx}{0.8\textwidth}{c|c|c|c|c|c}
\toprule
\textbf{Dataset}     & \textbf{Resolution} & \textbf{Max Nodes} & \textbf{Train} & \textbf{Val} & \textbf{Test} \\ \midrule
L-system~\cite{L-system} & $512 \times 512$ & $100$     & $100,000$        & $1,000$           & $20,000$     \\
Root                     & $570 \times 190$ & $117$     & $62,500$         & $78$              & $78$     \\
Grapevine~\cite{Guyot}   & $504 \times 378$ & $205$     & $118,500$        & $63$              & $255$     \\ \bottomrule
\end{tabularx}
\end{table}

\bmhead*{L-system synthetic dataset}
To systematically demonstrate the performance of our method, we perform an experiment using a large synthetic dataset.
We automatically generate images of tree patterns using pre-defined rules of Lindenmayer systems (L-system) \cite{L-system, L-system-papa}, which generate structural patterns using recursive processes.
We add randomness of branching patterns, branch length, and joint angles to increase the dataset variation.
The number of nodes in the graph is controlled at less than $100$.
The resolution of the generated images is $512\times 512$ pixels.
We generated $100000$ images for training, $1000$ for validation, and $20000$ for testing.

\bmhead*{Root dataset}
We use photographs of early-growing roots of Arabidopsis, which are often important targets of analysis in plant science.
In this dataset, the graph structures are manually annotated.
The dataset contains $781$ root images, and we randomly divide them into $625$ training, $78$ validation, and $78$ test images. 
Each graph contains up to $117$ nodes. The image resolution is $570\times 190$ pixels.
We use data augmentation involving rotation, flipping, and cropping for the training dataset, which collectively expands the training dataset to $62,500$ images.

\bmhead*{Grapevine dataset~\textnormal{\cite{Guyot}}} 
We use the 3D2cut Single Guyot Dataset~\cite{Guyot} containing grapevine tree images captured in an agricultural field with annotated branch patterns. 
The dataset contains relatively complex structures; the graph contains up to $205$ nodes.
The resolution is $504\times 378$ pixels.
The dataset contains $1503$ images, and we use the dataset split the same as \cite{Guyot}, where $1185$ images are for training, and $63$ and $255$ images are for validation and testing, respectively.
We use data augmentation in the same manner as the root dataset, resulting in $118,500$ involving rotation, flipping, and cropping for the training dataset.

\subsection{Generalized datasets}
\begin{figure}[tp]
    \centering
    \includegraphics[width=1.0\linewidth]{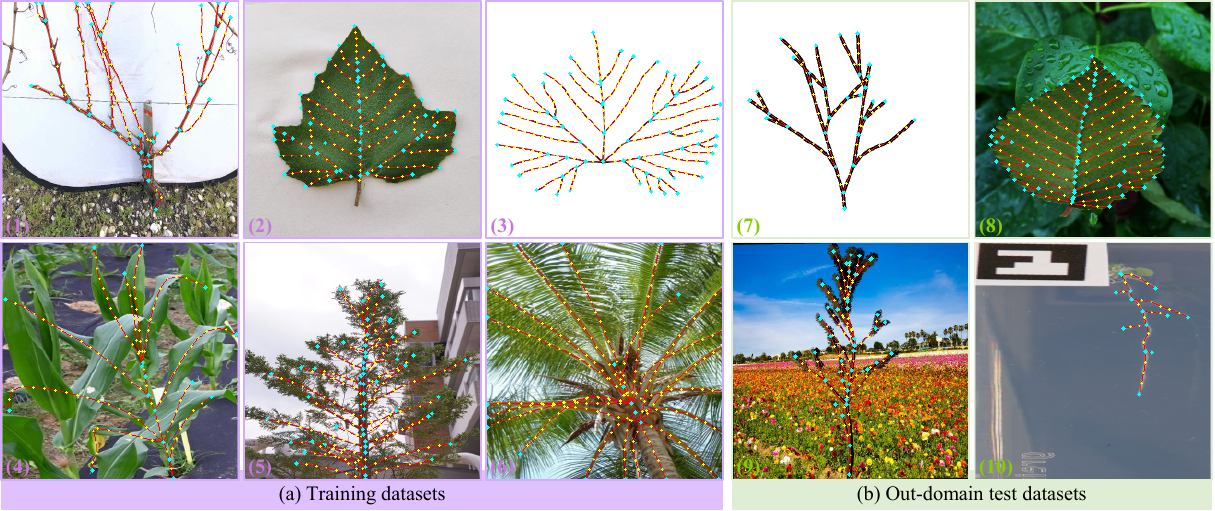}
    \caption{
        Sample images from the generalized datasets. Annotated graphs are superimposed, where yellow dots represent nodes and red lines represent edges.  
        (a) \textbf{Training datasets}:
        (1) grapevine, (2) LVD2021 leaf image, (3) LVD2021 vein mask, (4) MIPDB, (5) self-captured, (6) web-sourced.  
        (b) \textbf{Out-of-domain test datasets}:  
        (7) thickened L-system, (8) LVD with background, (9) tree synthetic, (10) root.  
        The datasets include diverse plant structures and imaging conditions to improve the model's generalization and robustness across in-domain and out-of-domain scenarios.}
    \label{fig:generalized_dataset_sample}
\end{figure}

To expand the model's applicability, we curate and process six training datasets and four out-of-domain test datasets. An overview of these datasets is provided in~\fref{fig:generalized_dataset_sample} and \Tref{tab:generalized_dataset}. 
Across the six training datasets, we collected and processed over $22,000$ individual plant instances (before augmentation), capturing diverse morphological, environmental, and taxonomic variations.

These datasets encompass real-world, synthetic, and web-sourced images, thus covering diverse structural variations and environmental contexts. We increased the maximum node count to 384 and standardized all images to $256 \times 256$ pixels to ensure consistent training and evaluation conditions. For all datasets, we constructed tree graphs by identifying key branch points and sampling them at uniform intervals of $13$ pixels, thereby establishing a standardized representation of structural patterns. The dataset details are described in the appendix.

\begin{table}[tp]
\centering
\renewcommand{\arraystretch}{1.0} 
\caption{Generalized dataset composition. Training datasets are marked in white, while out-of-domain test datasets are shaded in gray. Augmentation is applied only to training datasets to improve model generalization.}
\label{tab:generalized_dataset}
\begin{tabularx}{0.915\textwidth}{c|c|c|c|c}
\toprule
\textbf{Dataset} & \textbf{Max Nodes} & \textbf{Samples (Train/Val/Test)} & \textbf{Aug.} & \textbf{Total} \\ \midrule
Grapevine ~\cite{Guyot}                & 205 & 1202 / 150 / 151 & 10x   & 1503  \\ 
LVD2021 leaf image~\cite{LVD2021}      & 326 & 1858 / 232 / 232 & 10x  & 2322 \\
LVD2021 vein mask~\cite{LVD2021}       & 326 & 1858 / 232 / 232 & 1x   & 2322 \\
MIPDB~\cite{MIPDB}                     & 210 & 12992 / 249 / 250& 1x   & 13491 \\
Self-captured                          & 378 & 652 / 82 / 83    & 20x  & 817 \\
Web-sourced                            & 382 & 1320 / 166 / 168 & 10x  & 1654 \\ \midrule
\rowcolor[gray]{0.9} Thickened L-system     & 211 & -- / -- / 500   & 1x   & 500   \\ 
\rowcolor[gray]{0.9} LVD w/ background & 285 & -- / -- / 500   & 1x   & 500   \\ 
\rowcolor[gray]{0.9} Tree synthetic    & 384 & -- / -- / 500   & 1x   & 500   \\ 
\rowcolor[gray]{0.9} Root              & 71  & -- / -- / 782   & 1x   & 782   \\ 
\bottomrule
\end{tabularx}
\end{table}

\section{Experiments}
We conduct experiments to validate the effectiveness of the proposed method in both domain-specific and generalized scenarios using synthetic and real-world plant image datasets.

\subsection{Baselines}
Since the constrained graph generation task is new in this paper, there are few established baseline methods.
We compare our method with the state-of-the-art methods for 2D plant structure estimation and unconstrained graph generation.
Also, as an ablation study, we compare a simpler alternative to our method. Appendices provide additional comparisons with other baseline methods, including autoregressive graph generation.

\bmhead*{Two-stage~\textnormal{\cite{Guyot}}} We implement a 2D plant skeleton estimation method based on a two-stage method involving skeletonization and graph optimization with reference to~\cite{Guyot}. Specifically, vector fields of branch directions are generated by a neural network, followed by graph optimization to generate branch structure, in which we find our implementation outperforms the naive re-implementation of the existing method~\cite{Guyot}. Specific implementations and analyses are described in the appendices.

\bmhead*{Unconstrained~\textnormal{\cite{Relationformer}}} We compare the state-of-the-art (unconstrained) graph generation method, RelationFormer~\cite{Relationformer}. This method is identical to our method without applying the tree structure constraint.

\bmhead*{Test-time constraint} As a straightforward implementation of constrained graph generation, we apply MST only in the inference phase, where the graph generator is trained using the same procedure as the unconstrained method.

\subsection{Evaluation metrics}
We use different metrics to capture spatial similarity alongside the topological similarity of the predicted graphs.

\bmhead*{Street mover's distance (SMD)~\textnormal{\cite{GGT}}} SMD is a metric to assess the accuracy of the positions of graph edges, which is computed as the Wasserstein distance between the predicted and the ground truth edges. In our implementation, the distance is computed between densely sampled points on the edges, which is the same procedure as in the original paper proposing the SMD~\cite{GGT}.

\bmhead*{TOPO score~\textnormal{\cite{he2018roadrunner}}} We compute the TOPO scores to evaluate the topological mismatch of the output graph. This metric consists of the precision, recall, and F1 scores of the graph nodes, which are evaluated considering the edge topology. We use the implementation used in the Sat2Graph paper~\cite{Sat2Graph}, while we only evaluate the nodes with the degree $\neq 2$ that affect the tree structure, \ie, we only evaluate joint and leaf nodes in the graphs.

\bmhead*{Tree rate} To evaluate how well the output graph satisfies the constraint, we calculate the probability that the output graph forms a tree structure. While it is obvious that the tree rate becomes $100~\%$ for constrained methods, including ours, we are interested in how well the output of the unconstrained graph generation model can reflect the constraint by training on datasets that contain only tree graphs.

\subsection{Implementation details}
For RelationFormer in our method and the baseline comparison, we use the official implementation\footnote{\url{https://github.com/suprosanna/relationformer}, last accessed on December 20, 2024.} on PyTorch.
For other hyperparameters, we follow the original RelationFormer implementation used for road network extraction. 
We used early stopping for all datasets and methods by selecting the model with the best validation performance and terminating training after $30$ epochs without improvement.
The training of our method takes approximately $141$ hours for the synthetic dataset, $10$ hours for the root dataset, and $98$ hours for the grapevine dataset, all conducted on eight NVIDIA RTX A100 GPUs.

\begin{figure}[tp]
	\centering
	\includegraphics[width=\linewidth]{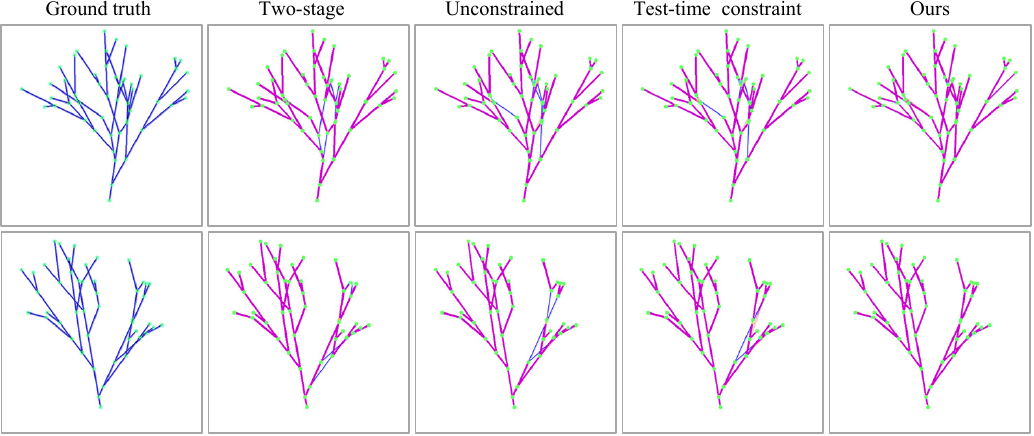}
	\caption{Visual results for the domain-specific training with the synthetic tree pattern dataset. For the synthetic tree pattern dataset: We translucently overlay the estimated and ground truth edges with red and blue lines, respectively. While all methods accurately detect nodes, only our method accurately predicts the availability of edges from given images compared to the baseline methods.}
	\label{fig:synthetic}
\end{figure}

\begin{figure}[tp]
	\centering
    \subfloat[][Visual results for the root dataset.]{
	  \includegraphics[width=0.98\linewidth]{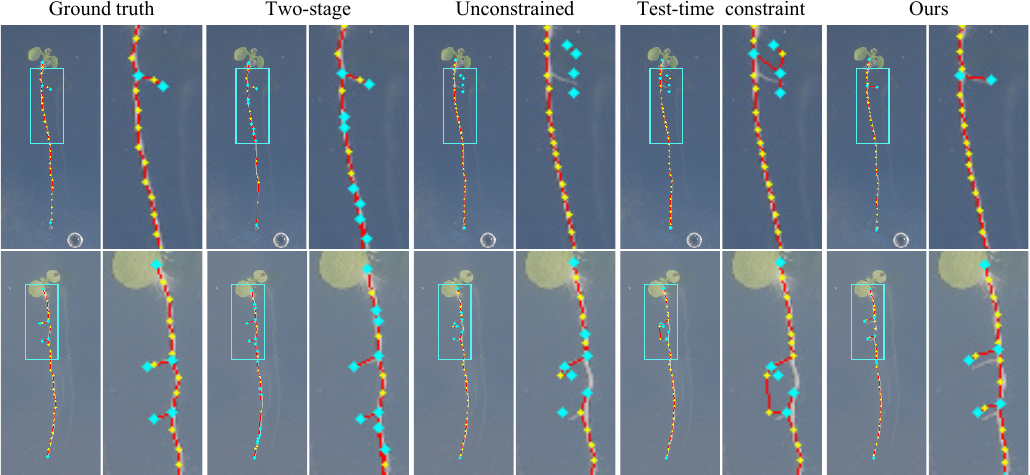}    
        \label{fig:root}
    }\\ 
    \subfloat[][Visual results for the grapevine dataset.]{
    	\includegraphics[width=0.98\linewidth]{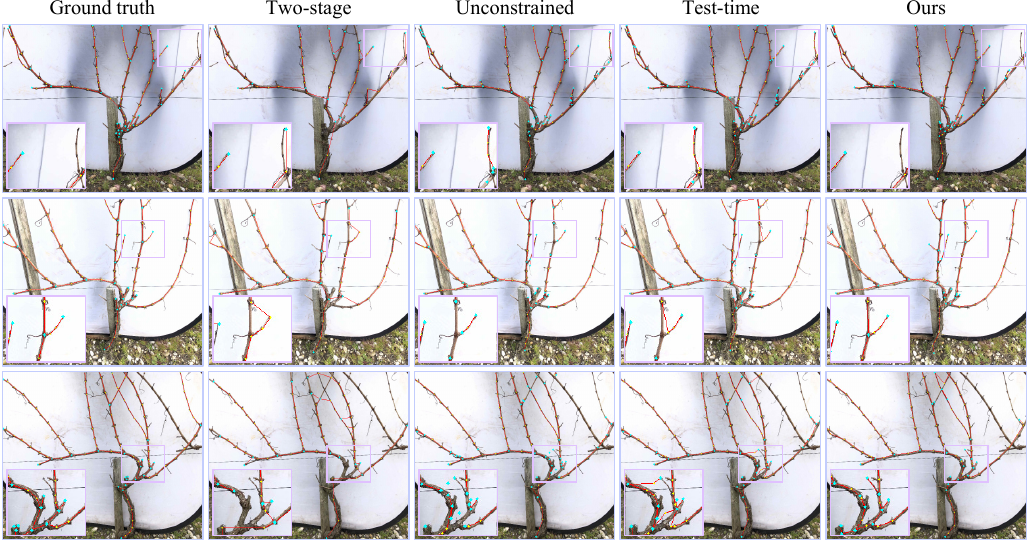}    
    	\label{fig:guyot}    
    }    
	\caption{Visual results for the domain-specific training with the real image datasets: Red lines, yellow dots, and cyan dots indicate the predicted graph edges, nodes, and keypoints (\ie, joints and leaf nodes). Our method accurately estimates the target plant structures compared with baseline methods, demonstrating the applicability of our method for practical uses in plant science and agriculture.}
    \label{fig:real}
\end{figure}

\begin{table}[tp]
\centering
\caption{Quantitative results on domain-specific training. Our method significantly improves both the shape and topology of the predicted graph while enforcing the given constraints. The best scores are highlighted in \textbf{bold}.}
\label{tab:specific_results}
\begin{tabular}{c|c|c|ccc|c}
\toprule
\multirow{2}{*}{\textbf{Dataset}}   & \multirow{2}{*}{\textbf{Method}}      & \multirow{2}{*}{\textbf{SMD $\downarrow$}} & \multicolumn{3}{c|}{\textbf{TOPO score $\uparrow$}} & \textbf{Tree rate} \\
                          &                                    &                      & \textbf{Prec.}     & \textbf{Rec.}    & \textbf{F1}       & \textbf{[\%]}   \\ \midrule

\multirow{4}{*}{Synthetic} & Two-stage~\cite{Guyot}             & $1.91 \times 10^{-3}$     & 0.940         & 0.886        & 0.912        &\textbf{100.0}     \\
                           & Unconstrained~\cite{Relationformer}& $1.43 \times 10^{-5}$     & 0.978         & 0.929        & 0.953        & 36.2      \\
                           & Test-time constraint               & $6.26 \times 10^{-6}$     & 0.977         & 0.953        & 0.965        &\textbf{100.0}     \\
                           & Ours                               & $\bm{4.78 \times 10^{-6}}$& \textbf{0.986}&\textbf{0.968}&\textbf{0.977}&\textbf{100.0}     \\ \midrule
\multirow{4}{*}{Root}      & Two-stage~\cite{Guyot}             & $4.83 \times 10^{-4}$     & 0.767         &0.732         &0.749         &\textbf{100.0}     \\
                           & Unconstrained~\cite{Relationformer}& $1.19 \times 10^{-4}$     & 0.831         & 0.633        & 0.719        & 35.9     \\
                           & Test-time constraint               & $1.52 \times 10^{-4}$     & 0.829         & 0.771        & 0.799        &\textbf{100.0}     \\
                           & Ours                               & $\bm{8.82 \times 10^{-5}}$&\textbf{0.861} &\textbf{0.807}&\textbf{0.833}&\textbf{100.0}     \\ \midrule
\multirow{4}{*}{Grapevine} & Two-stage~\cite{Guyot}             & $4.24 \times 10^{-4}$     & 0.677         & 0.589        & 0.630        &\textbf{100.0}     \\
                           & Unconstrained~\cite{Relationformer}& $1.45 \times 10^{-4}$     & \textbf{0.963}& 0.559        & 0.708        & 0.0      \\
                           & Test-time constraint               & $1.47 \times 10^{-4}$     & 0.896         & 0.840        & 0.867        &\textbf{100.0}     \\
                           & Ours                               & $\bm{1.03 \times 10^{-4}}$& 0.899         &\textbf{0.843}&\textbf{0.870}&\textbf{100.0}     \\ \bottomrule
\end{tabular}
\end{table}

\subsection{Results}

\subsubsection{Results on domain-specific training}
\paragraph*{Results on synthetic dataset}
\Fref{fig:synthetic} shows visual results for the synthetic dataset, where the red and blue lines indicate the predicted and ground truth edges, respectively. 
Since they are shown translucently, if the estimated edge overlaps the true edge, it is displayed in purple. 
Similarly, cyan and yellow dots indicate the nodes, which merge into green if correctly estimated. From the results, all methods correctly estimate the node positions.
The existing unconstrained method outputs isolated edges and cycles. Although the two-stage and test-time constraint methods enforce the tree structure constraint, they often produce incorrect edges. 
Compared to the baselines, our method accurately generates the graph edges. 

The above trend can be quantitatively confirmed in \Tref{tab:specific_results}. The unconstrained method produces tree structures with only about $30$~\% probability, even though all the training graphs form tree structures. Although introducing the test-time constraint and two-stage methods improves the shape and topology, there are still many incorrect estimates. Compared to those baseline methods, our method significantly improves both edge positions and graph topology.

\paragraph*{Results on real datasets}
\Fref{fig:real} shows the results of skeleton estimation for two real-world datasets. 
For these figures, red lines, yellow dots, and cyan dots indicate the edges, nodes, and keypoints (\ie, joints and leaf nodes), respectively.
In agreement with the synthetic results, our method predicts visually better structures, while the unconstrained model hardly produces tree structures.
The method with the test-time constraint clearly produces false edges, as shown in the results for the grapevine images. 
The two-stage method is often sensitive to the node detection error, leading to unnecessary ({\it cf}.~\fref{fig:root}) or missing ({\it cf}.~\fref{fig:guyot}) keypoints.
In these practical settings, our end-to-end pipeline especially benefits from the simultaneous optimization of edge and node detection, resulting in faithful predictions at both nodes and edges for real-world datasets.

The quantitative results in \Tref{tab:specific_results} confirm the advantage of our method for real-world scenes.
Our method shows particularly compelling results on grapevine datasets with relatively complex branching structures, outperforming the second-best method (test-time constraint) by approximately $30~\%$ improvement on the edge accuracy evaluated by SMD.

\begin{figure}[tp]
	\centering
        \includegraphics[width=\linewidth]{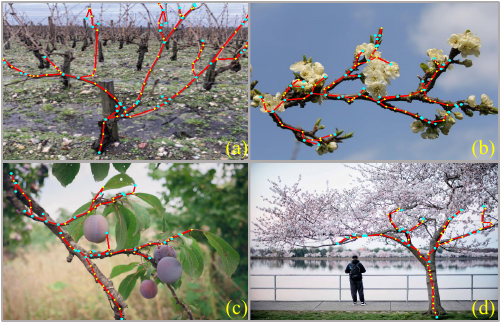}
	\caption{Visual results highlighting generalization ability of domain-specific training using the grapevine dataset for (a) a grapevine tree under a natural background and (b--d) other tree species.}
	\label{fig:sakura}
\end{figure}

\paragraph*{Generalization ability of domain-specific training}
Although the model is trained in a domain-specific manner, our model has a certain ability of generalization.
Although the model is trained with the Grapevine dataset with few background textures, it successfully works for grapevine images with background textures (\fref{fig:sakura}(a)) and for other tree species (\fref{fig:sakura}(b--d)).
These results highlight the generalizability of our method.

\begin{table}[tp]
\setlength{\tabcolsep}{4pt} 
\centering
\caption{Results on generalized training. Our method significantly improves both the shape and topology of the predicted graph while enforcing the given constraints, especially for out-of-domain scenarios.}
\label{tab:generalized_results}
\begin{tabular}{c|c|c|ccc|c}
\toprule
\multirow{2}{*}{\parbox{3.5cm}{\centering \textbf{Inner-domain datasets}}} & \multirow{2}{*}{\textbf{Method}} & \multirow{2}{*}{\textbf{SMD $\downarrow$}} & \multicolumn{3}{c|}{\textbf{TOPO score $\uparrow$}} & \textbf{Tree rate} \\
                         &                         &                                         & \textbf{Prec.}             & \textbf{Rec.}             & \textbf{F1}     & \textbf{[\%]}      \\ \midrule
\multirow{4}{*}{\parbox{3.5cm}{\centering Grapevine}} & Two-stage & \wll{$1.52 \times 10^{-2}$} & \sll{0.898}    & \bll{0.817}      & \bll{0.855}      & \bll{100.0}       \\
                                   & Unconstrained         & \wll{$3.58 \times 10^{-4}$}     & \bll{0.956}       & \wll{0.398}      & \wll{0.562}      & \wll{0.0}       \\
                                   & Test-time constraint  & \bll{$1.29 \times 10^{-4}$}     & \wll{0.876}       & \wll{0.792}      & \wll{0.832}      & \bll{100.0}       \\
                                   & Ours                  & \sll{$1.57 \times 10^{-4}$}     & \wll{0.885}       & \sll{0.798}      & \sll{0.839}      & \bll{100.0}      \\ \midrule
\multirow{4}{*}{\parbox{3.5cm}{\centering LVD2021 leaf image}}& Two-stage  & \wll{$6.36 \times 10^{-3}$}  & \sll{0.924}  & \bll{0.772} & \bll{0.842}   & \bll{100.0}       \\
                                   & Unconstrained         & \wll{$5.62 \times 10^{-4}$}     & \bll{0.963}       & \wll{0.269}      & \wll{0.421}      & \wll{0.0}       \\
                                   & Test-time constraint  & \sll{$5.44 \times 10^{-4}$}     & \wll{0.881}       & \wll{0.731}      & \wll{0.799}      & \bll{100.0}       \\
                                   & Ours                  & \bll{$5.41 \times 10^{-4}$}     & \wll{0.890}       & \sll{0.755}      & \sll{0.817}      & \bll{100.0}       \\ \midrule
\multirow{4}{*}{\parbox{3.5cm}{\centering LVD2021 vein mask}} & Two-stage & \wll{$1.81 \times 10^{-2}$} & \wll{0.943} & \wll{0.802} & \sll{0.867}      & \bll{100.0}       \\
                                   & Unconstrained         & \wll{$4.20 \times 10^{-4}$}     & \bll{0.979}       & \wll{0.309}      & \wll{0.470}      & \wll{0.0}       \\
                                   & Test-time constraint  & \sll{$3.86 \times 10^{-4}$}     & \wll{0.931}       & \sll{0.803}      & \wll{0.862}      & \bll{100.0}       \\
                                   & Ours                  & \bll{$3.75 \times 10^{-4}$}     & \sll{0.948}       & \bll{0.836}      & \bll{0.888}      & \bll{100.0}       \\ \midrule
\multirow{4}{*}{\parbox{3.5cm}{\centering MIPDB}} & Two-stage  & \wll{$7.45 \times 10^{-1}$}  & \wll{0.011}      & \wll{0.012}      & \wll{0.011}      & \wll{39.2}      \\
                                   & Unconstrained         & \wll{$1.74 \times 10^{-3}$}     & \bll{0.843}       & \wll{0.357}      & \wll{0.502}      & \wll{0.0}       \\
                                   & Test-time constraint  & \bll{$1.55 \times 10^{-3}$}     & \wll{0.703}       & \sll{0.623}      & \sll{0.660}      & \bll{100.0}       \\
                                   & Ours                  & \sll{$1.61 \times 10^{-3}$}     & \sll{0.723}       & \bll{0.653}      & \bll{0.686}      & \bll{100.0}       \\ \midrule
\multirow{4}{*}{\parbox{3.5cm}{\centering Self-captured}} & Two-stage & \wll{$8.55 \times 10^{-2}$} & \wll{0.486} & \wll{0.358}     & \wll{0.413}      & \wll{95.2}       \\
                                   & Unconstrained         & \wll{$2.12 \times 10^{-3}$}     & \bll{0.912}       & \wll{0.354}      & \wll{0.510}      & \wll{1.2}       \\
                                   & Test-time constraint  & \bll{$1.68 \times 10^{-3}$}     & \sll{0.747}       & \bll{0.648}      & \bll{0.694}      & \bll{100.0}       \\
                                   & Ours                  & \sll{$1.92 \times 10^{-3}$}     & \sll{0.747}       & \sll{0.642}      & \sll{0.691}      & \bll{100.0}       \\ \midrule
\multirow{4}{*}{\parbox{3.5cm}{\centering Web-sourced}} & Two-stage & \wll{$3.36 \times 10^{-1}$} & \wll{0.331}   & \wll{0.188}      & \wll{0.240}     & \wll{72.0}       \\
                                   & Unconstrained         & \wll{$9.00 \times 10^{-3}$}     & \bll{0.821}       & \wll{0.304}      & \wll{0.443}      & \wll{1.8}       \\
                                   & Test-time constraint  & \bll{$7.92 \times 10^{-3}$}     & \sll{0.656}       & \bll{0.564}      & \bll{0.606}      & \bll{100.0}       \\
                                   & Ours                  & \sll{$8.26 \times 10^{-3}$}     & \wll{0.634}       & \sll{0.559}      & \sll{0.594}      & \bll{100.0}   \\ \midrule\midrule
\multirow{2}{*}{\parbox{3.5cm}{\centering \textbf{Out-of-domain datasets}}} & \multirow{2}{*}{\textbf{Method}} & \multirow{2}{*}{\textbf{SMD $\downarrow$}} & \multicolumn{3}{c|}{\textbf{TOPO score $\uparrow$}} &  \textbf{Tree rate} \\
                                   &                                  &                                             & \textbf{Prec.}     & \textbf{Rec.}     & \textbf{F1}    & [\%]               \\ \midrule

\multirow{4}{*}{\parbox{3.5cm}{\centering Thickened L-system}}& Two-stage & \wll{$2.02 \times 10^{-2}$} & \wll{0.716} & \wll{0.578} & \wll{0.640}     & \bll{100.0}       \\
                                   & Unconstrained         & \wll{$6.61 \times 10^{-5}$}     & \bll{0.955}       & \wll{0.389}      & \wll{0.553}     & \wll{0.0}       \\
                                   & Test-time constraint  & \sll{$5.61 \times 10^{-5}$}     & \wll{0.843}       & \sll{0.754}      & \sll{0.796}     & \bll{100.0}       \\
                                   & Ours                  & \bll{$5.59 \times 10^{-5}$}     & \sll{0.892}       & \bll{0.795}      & \bll{0.841}     & \bll{100.0}       \\ \midrule
\multirow{4}{*}{\parbox{3.5cm}{\centering LVD w/ background}}& Two-stage & \wll{$2.99 \times 10^{-2}$} & \bll{0.892} & \bll{0.596}  & \bll{0.715}     & \wll{98.8}       \\
                                   & Unconstrained         & \wll{$2.40 \times 10^{-2}$}     & \sll{0.834}       & \wll{0.213}      & \wll{0.340}     & \wll{0.8}       \\
                                   & Test-time constraint  & \sll{$2.34 \times 10^{-2}$}     & \wll{0.678}       & \wll{0.450}      & \wll{0.541}     & \bll{100.0}       \\
                                   & Ours                  & \bll{$2.22 \times 10^{-2}$}     & \wll{0.725}       & \sll{0.502}      & \sll{0.594}     & \sll{99.8}       \\ \midrule
\multirow{4}{*}{\parbox{3.5cm}{\centering Tree synthetic}} & Two-stage & \wll{$4.97 \times 10^{-2}$} & \sll{0.796} & \wll{0.216}    & \wll{0.340}     & \wll{97.4}       \\
                                   & Unconstrained         & \wll{$1.16 \times 10^{-2}$}     & \bll{0.929}       & \wll{0.085}      & \wll{0.156}     & \wll{0.1}       \\
                                   & Test-time constraint  & \sll{$8.57 \times 10^{-3}$}     & \wll{0.767}       & \sll{0.347}      & \sll{0.478}     & \sll{99.8}       \\
                                   & Ours                  & \bll{$7.34 \times 10^{-3}$}     & \wll{0.772}       & \bll{0.352}      & \bll{0.483}     & \bll{100.0}       \\ \midrule
\multirow{4}{*}{\parbox{3.5cm}{\centering Root}} & Two-stage & \wll{$9.50 \times 10^{-1}$}    & \wll{0.032}      & \wll{0.005}      & \wll{0.008}     & \wll{5.0}       \\
                                   & Unconstrained         & \wll{$6.17 \times 10^{-2}$}     & \bll{0.694}       & \wll{0.482}      & \sll{0.569}     & \wll{8.2}       \\
                                   & Test-time constraint  & \sll{$6.06 \times 10^{-2}$}     & \sll{0.585}       & \bll{0.599}      & \bll{0.592}     & \sll{94.1}       \\
                                   & Ours                  & \bll{$2.70 \times 10^{-2}$}     & \wll{0.563}       & \sll{0.548}      & \wll{0.555}     & \bll{98.2}       \\ \bottomrule
\end{tabular}
\end{table}

\begin{figure}[tp]
	\centering
        \includegraphics[width=1.0\linewidth]{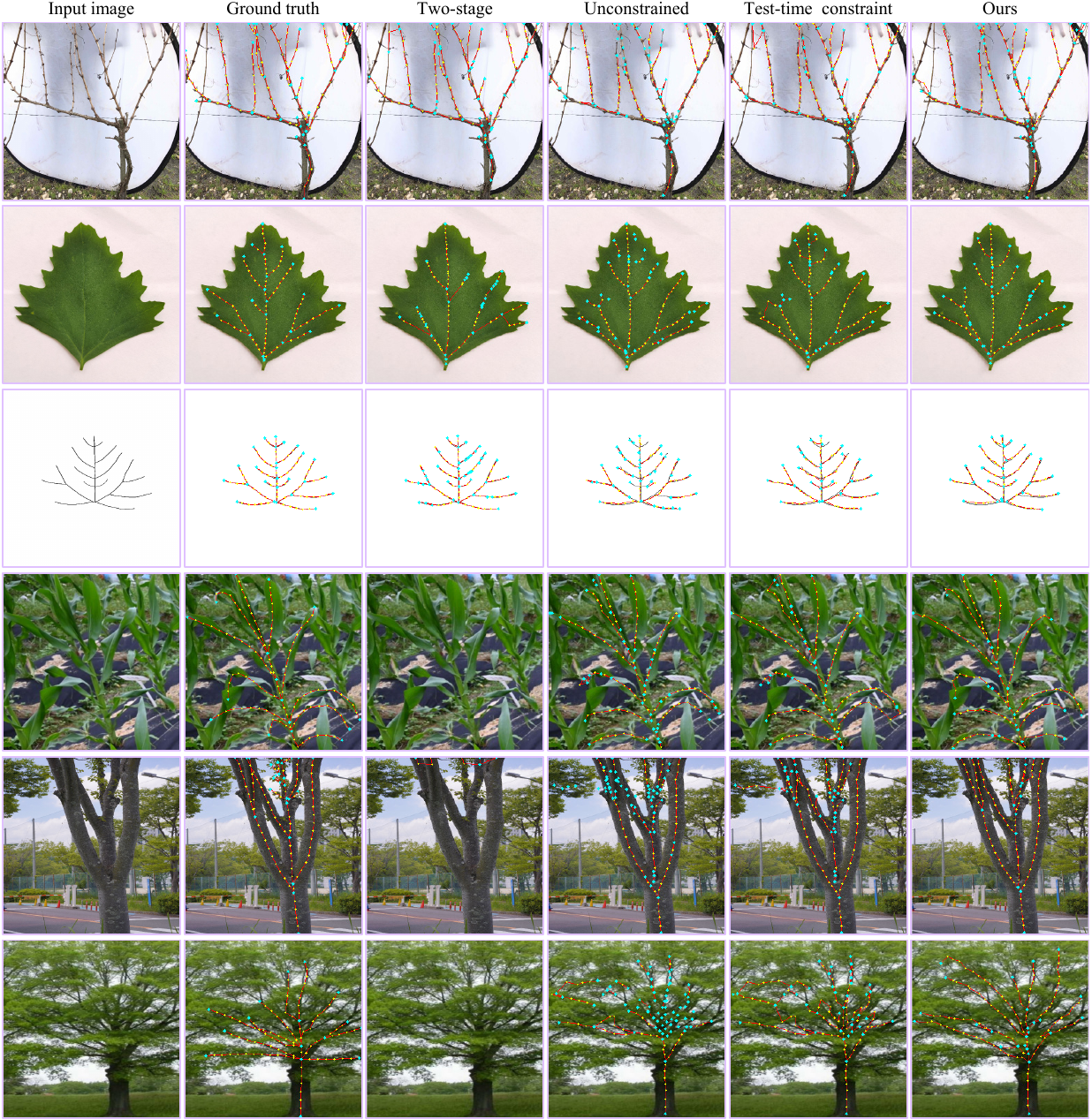}
	\caption{Visual results on the inner-domain datasets. From top to bottom: grapevine, LVD2021 leaf image, LVD2021 vein mask, MIPDB, self-captured, and web-sourced datasets. The two-stage method sometimes produces very few nodes (\eg, on the Self-captured dataset) or even fails to generate any nodes (\eg, on MIPDB and web-sourced dataset). On the grapevine and LVD2021 datasets, it frequently outputs multiple redundant keypoints around actual nodes (visible as dense clusters of blue markers), thereby inflating TOPO scores but also increasing SMD. In contrast, our method achieves a more balanced performance, maintaining stable TOPO scores and low SMD values.}
	\label{fig:training_dataset_result}
\end{figure}

\subsubsection{Results on generalized datasets}
The evaluation of generalized datasets examines the model's performance across diverse plant structures. We group these results into three aspects: (1) \textit{Inner-domain results}, evaluating performance on datasets resembling the training distribution; (2) \textit{Out-of-domain results}, assessing adaptability to novel conditions; and (3) \textit{Generalization ability}, highlighting performance on unseen and diverse data distributions.

\paragraph*{Inner-domain results}
\Tref{tab:generalized_results} compares results across training-related datasets, including grapevine, LVD2021 leaf image, LVD2021 vein mask, MIPDB, self-captured, and web-sourced datasets. We evaluate each method using the SMD, TOPO, and tree rate metrics.

The \textbf{two-stage} method often generates a small number of redundant nodes near actual branching points during the node detection stage. These redundant nodes are typically close to the actual keypoints. Since TOPO evaluates correctness by searching within a neighborhood around ground truth keypoints, these redundant nodes increase the likelihood of matching. As a result, both precision and recall remain high. However, these redundant nodes introduce short edges that misalign with the actual graph structure, significantly increasing the SMD value.

In contrast, the \textbf{unconstrained} method generates numerous independent nodes in areas where keypoints might exist. While this increases the chances of matching predicted nodes to ground truth keypoints, many of these nodes fall outside TOPO's search range. Consequently, TOPO precision appears high due to the excess of predicted nodes, but recall remains low as not all ground truth keypoints are matched.

The \textbf{test-time constraint method} provides a tree structure at inference, improving TOPO scores and spatial coherence compared to unconstrained predictions. However, because it applies constraints \emph{only at inference time}, it does not fully leverage structural information to refine the model's ability. Still, the test-time constraint method is a strong baseline that ensures tree structure while maintaining reasonable topological fidelity.

\textbf{Our method (ours)} combines structural priors directly into the learning process. Our method consistently outperforms or closely matches the performance of the test-time constraint method in a wide range of datasets. 
 
In summary, ours and the test-time constraint method are the top contenders on the training datasets. The test-time constraint method provides a simple baseline solution that ensures tree structure by design. At the same time, ours further refines the predictions by integrating tree constraints into the inference process more naturally.

\begin{figure}[tp]
	\centering
        \includegraphics[width=1.0\linewidth]{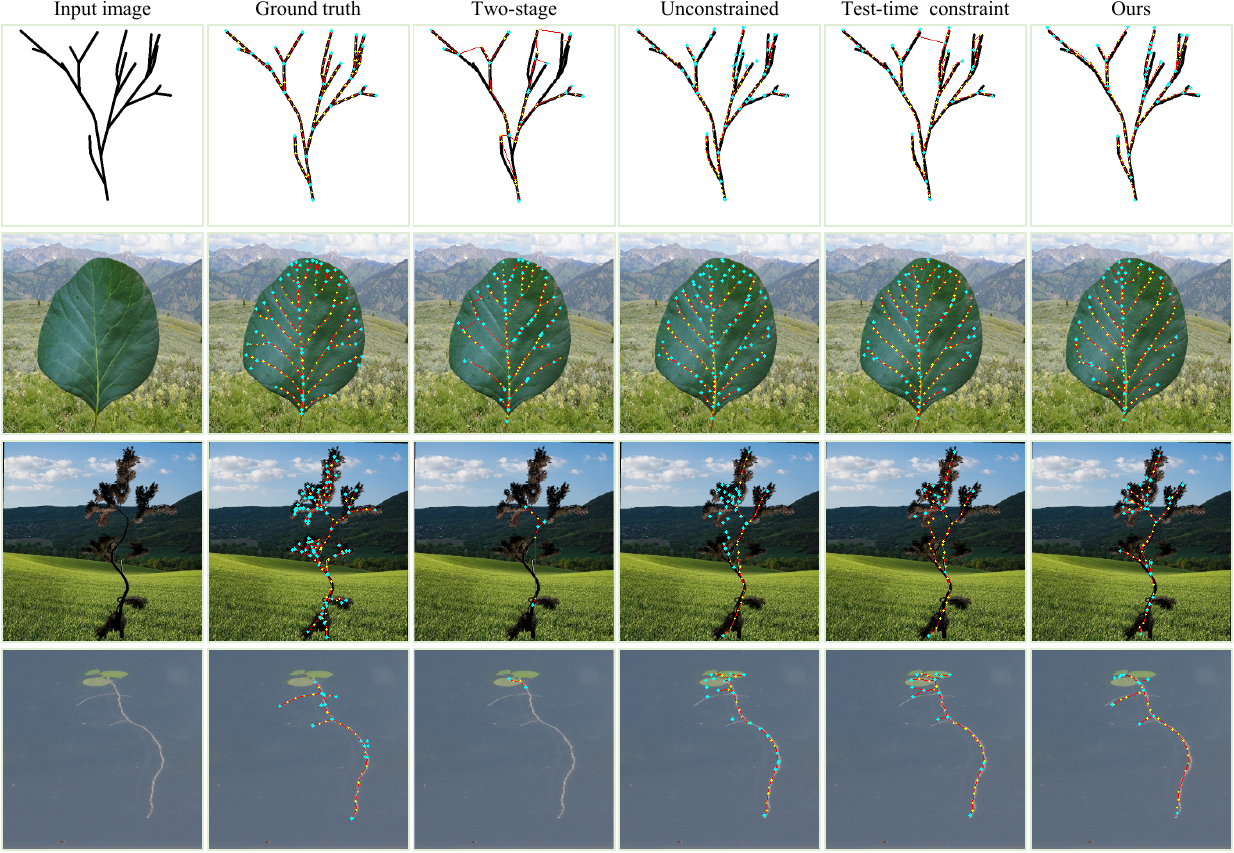}
	\caption{Visual results on the out-of-domain test datasets: thickened L-system, LVD with background, tree synthetic, and root datasets (top to bottom). The unconstrained method generates excessive isolated points and disconnected components, inflating TOPO precision but reducing recall due to poor keypoint coverage. In contrast, our method preserves structural consistency, balancing TOPO precision and recall, and produces coherent tree structures. The root dataset highlights the difficulty of detecting short and thin lateral roots, which remains challenging for all methods.}
	\label{fig:outdomain_dataset_result}
\end{figure}

\paragraph*{Out-of-domain results}
The out-of-domain test datasets, including thickened L-system, LVD with background, tree synthetic, and root datasets, introduce additional complexity and distributional shifts, testing the model's generalization beyond training conditions. Visual examples are shown in \fref{fig:outdomain_dataset_result}, and detailed metrics are provided in \Tref{tab:generalized_results}.

On the \textbf{thickened L-system} dataset, both the test-time constraint method and ours achieve near-perfect tree rates. Our method achieves a higher TOPO F1 score and lower SMD, showing that internally integrated structural priors can better handle such domain shifts, producing more topologically coherent and spatially accurate outputs.

For the \textbf{LVD with background} dataset, which merges natural leaves and environmental variations, as discussed, the two-stage method achieves a relatively high TOPO F1 but suffers from poor SMD due to redundant nodes near branching points. Both the test-time constraint and our method significantly improve SMD, with our approach achieving the lowest value. This proves our method provides the most geometrically accurate tree structures while balancing spatial precision and topological correctness.

Our method outperforms all baselines on the \textbf{tree synthetic} dataset, achieving the highest F1, lowest SMD, and a 100 \% tree rate. The lowest SMD further highlights the robustness of our combined structural constraints and MST-based pipeline in generating predictions that closely match the ground truth.

Finally, on the \textbf{root} dataset, ours and the test-time constraint method perform well, achieving high tree rates. Notably, our method achieves the lowest SMD, signifying better spatial alignment with the ground truth while maintaining comparable F1 scores and consistent topological accuracy.

\begin{figure}[tp]
	\centering
        \includegraphics[width=1.0\linewidth]{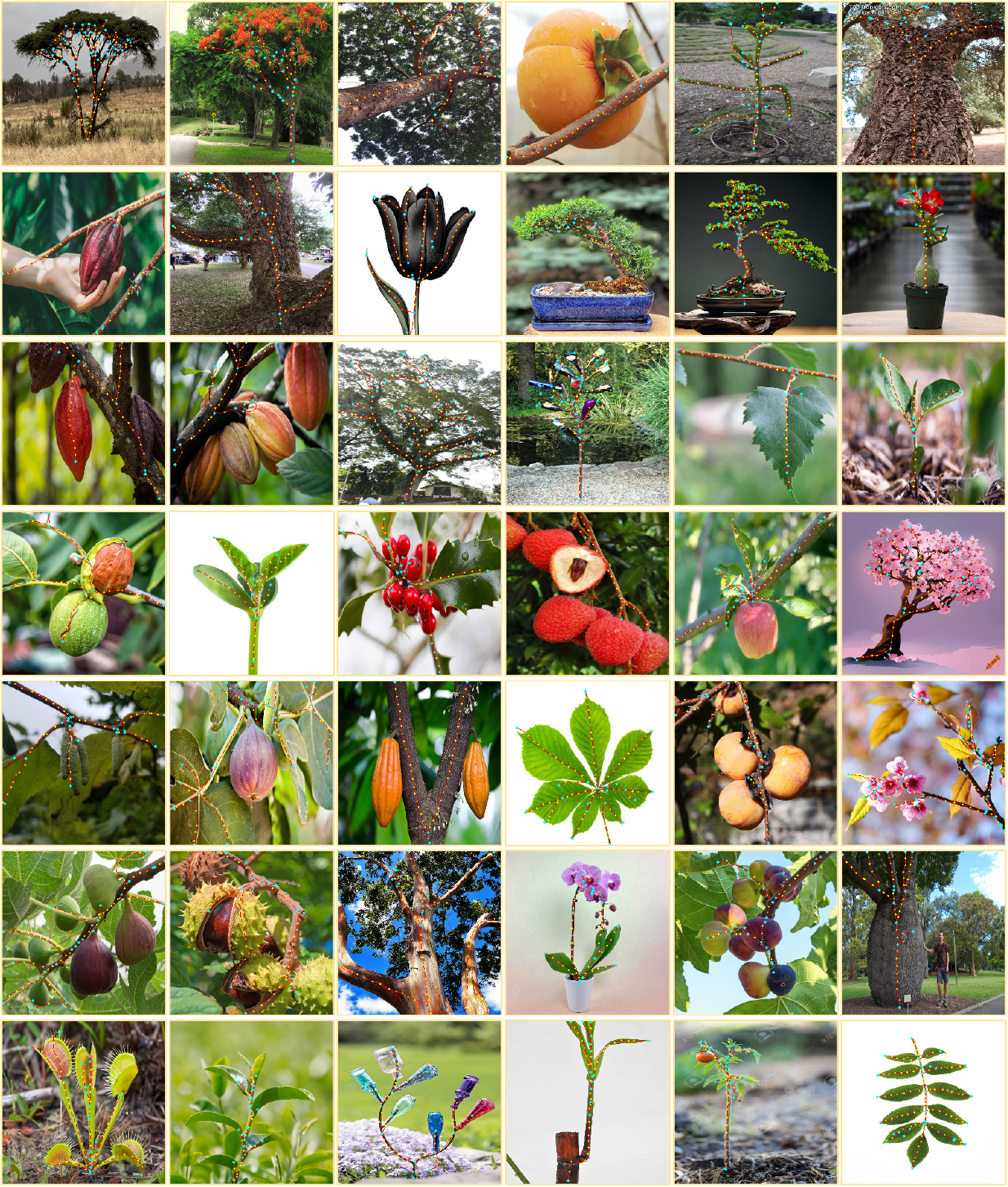}
	\caption{Visual results on the unlabeled web-sourced dataset, including flowers, rare plants, potted plants, and hand-drawn trees. These results highlight the diversity of test cases and demonstrate our method's generalization ability across different data domains and styles.}
	\label{fig:outdomain_google_result}
\end{figure}

\paragraph*{Generalization ability}
To evaluate the generalization ability of our method, we qualitatively test our method on the unlabeled web-sourced test dataset. We search and download diverse datasets using an extensive list of keywords, which include various plant species, unique ferns, crops, and plants with distinct appearances. These datasets feature significant variability, ranging from rare species like \textit{Dionaea muscipula} to major crops such as soybean and sugarcane, as well as abstract hand-drawn tree structures. 

Some visual results are shown in \fref{fig:outdomain_google_result}. 
Our method consistently performs well across these datasets, achieving accurate graph reconstruction even in unusual morphological structures or abstract representations. For instance, the model performed well on flowers, rare species, potted plants, and hand-drawn trees. The appendix shows more examples.

\section{Conclusion}
We present the first attempt at tree-constrained graph generation from a single image, especially useful for plant skeleton estimation.
Our method combines modern learning-based graph generators and traditional graph algorithms, where we project the unconstrained predictions onto constrained graphs during the training loop. 
Our SFS layer behaves to selectively suppress the gradient to be backpropagated, which exploits the constraint cues from the non-differentiable graph algorithm (\ie, MST).
Our method can be easily integrated with off-the-shelf graph generators, as we develop PlantPose based on RelationFormer. 

Through extensive experiments on six training datasets and four out-of-domain test datasets, we demonstrate that PlantPose achieves state-of-the-art results on a wide range of plant images. The strong performance of our model across generalized datasets highlights the effectiveness and versatility of the SFS layer in promoting robust generalization to diverse and previously unseen data distributions. The compelling results on both quantitative metrics and qualitative evaluations underscore its practical utility for plant science and agriculture. 

\paragraph*{Limitations} 
We use graph algorithms during each training iteration, taking a longer training time than unconstrained methods, where fast GPU-based MST implementations (\eg,~\cite{vineet2009fast}) can improve computational performance.
The success of our method depends on the accuracy of the underlying graph generation model, as we see a few undetected nodes in the visual results.

\section*{Declarations}

\bmhead*{Funding}
This work was partly supported by JSPS KAKENHI Grant Numbers JP23H05491, JP25K03140, and JP25K00142, and JST FOREST Grant Number~\hbox{JPMJFR206F}.

\bmhead*{Competing interests}
The authors declare that they have no competing interests.

\bmhead*{Data availability}
Our own data used and/or analyzed during the current study is publicly available in the repository at \url{https://github.com/huntorochi/PlantPose}. The curated datasets can be provided by the original authors.

\bmhead*{Code availability}
The source code supporting the findings of this study is publicly available in the repository at \url{https://github.com/huntorochi/PlantPose}.

\bmhead*{Author contribution}
\textbf{Conceptualization:} \author{\fnm{Xinpeng} \sur{Liu}}, \author{\fnm{Fumio} \sur{Okura}}. \\
\textbf{Methodology:} \author{\fnm{Xinpeng} \sur{Liu}}, \author{\fnm{Fumio} \sur{Okura}}. \\
\textbf{Formal analysis and investigation:} \author{\fnm{Xinpeng} \sur{Liu}}. \\
\textbf{Writing – original draft preparation:} \author{\fnm{Xinpeng} \sur{Liu}}, \author{\fnm{Hiroaki} \sur{Santo}}, \author{\fnm{Fumio} \sur{Okura}}. \\
\textbf{Writing – review and editing:} \author{\fnm{Xinpeng} \sur{Liu}}, \author{\fnm{Fumio} \sur{Okura}}. \\
\textbf{Funding acquisition:} \author{\fnm{Fumio} \sur{Okura}}. \\
\textbf{Resources:} \author{\fnm{Yosuke} \sur{Toda}}. \\
\textbf{Supervision:} \author{\fnm{Fumio} \sur{Okura}}.

\section*{Acknowledgments}
We thank Professor Yasuyuki Matsushita for insightful discussions throughout the study. We also thank Momoko Takagi, Manami Okazaki, and Professor Kei Hiruma for providing us with root images.

\newpage

\begin{appendices}

\section{Details of SFS layer}
\label{sec:supp_analysis}

\subsection{Motivation}
Our method infers a tree graph via MST as formulated in Eq.~(1)--Eq.~(4) in the main paper. Our task's goal is to optimize the graph generation network so that the \textbf{final output (\ie, tree graph via MST) becomes similar to the ground-truth tree graph}. 
Since MST modifies the edge availability in \emph{unconstrained} inferences, the unconstrained methods evaluating the unconstrained graph edges are indirect. 
Instead, our method \textbf{directly evaluates the quality of the final output tree} by mimicking MST. While experiments highlight our method's benefit, the following theoretical analysis also supports this intuition.

\subsection{Derivation}
This section details the derivation of Eq.~(10) in the main paper. To make this material self-contained, we repeat several descriptions in the main paper.

As discussed in the main paper, we consider the edge probabilities $\unconst{\V{y}}_{\ijsub} =[\unconst{y}^{+}_{\ijsub},\unconst{y}^{-}_{\ijsub}]^\top$ is usually computed through the softmax activation $\softmax$ applied to the output feature vector of the final layer $\unconst{\V{f}}_{\ijsub}=[\unconst{f}^{+}_{\ijsub},\unconst{f}^{-}_{\ijsub}]^\top$ as

\begin{align}
\begin{array}{ll}
\unconst{\V{y}}_{\ijsub} = \softmax(\unconst{\V{f}}_{\ijsub}) \\
= \Bigg[\frac{\exp(\unconst{f}^{+}_{\ijsub})}{\exp(\unconst{f}^{+}_{\ijsub})+\exp(\unconst{f}^{-}_{\ijsub})},
\frac{\exp(\unconst{f}^{-}_{\ijsub})}{\exp(\unconst{f}^{+}_{\ijsub})+\exp(\unconst{f}^{-}_{\ijsub})}\Bigg]^\top.
\end{array}
\label{eq:feat_softmax}
\end{align}

The set of unconstrained graph edges $\unconst{E}$ is then obtained by comparing the edge existence probabilities as
\begin{equation}
\label{eq:thresholding}
\unconst{\edge} = \{ (i, j) \mid \unconst{y}^{+}_{\ijsub} > \unconst{y}^{-}_{\ijsub} \},
\end{equation}
in which $\unconst{E}$ records node pairs where the edge exists. 

Suppose the projection function $\Proj$ converts the set of unconstrained edge probabilities $\{\unconst{\V{y}}_{\ijsub}\}$ to a set of constrained edges $E$. 
Let the difference of two sets be $E^+ = E - \unconst{E}$ and $E^- = \unconst{E} - E$, denoting the sets of edges newly added and removed by the projection.
To mimic the discrete (and non-differentiable) inferences by $\Proj$ in the differentiable end-to-end learning, we modify the edge features corresponding to $E^+ \cup E^-$ in the differentiable forward process. 
Here, we want to get the edge probabilities that approximate the constrained edges $E$, which can be denoted as
\begin{align}
\V{y}_{\ijsub} 
= \left\{
\begin{array}{l@{\hspace{2em}}l}
\label{eq:hard}
\left[\:\:\:\:1\:\:\:\:,\:\:\:\:0\:\:\:\: \right]^\top           & ((i,j)\in E^+)\\
\left[\:\:\:\:0\:\:\:\:,\:\:\:\:1\:\:\:\: \right]^\top           & ((i,j)\in E^-)   \\
\left[\unconst{{y}}^{+}_{\ijsub},  \unconst{{y}}^{-}_{\ijsub} \right]^\top       & \mathrm{(otherwise)}.
\end{array}
\right. \\
\label{eq:epsilon}
\sim \left\{
\begin{array}{l@{\hspace{2em}}l}
\left[1-\epsilon,\:\:\:\:\epsilon\:\:\:\: \right]^\top           & ((i,j)\in E^+)\\
\left[\:\:\:\:\epsilon\:\:\:\:,  1-\epsilon \right]^\top           & ((i,j)\in E^-)   \\
\left[\unconst{{y}}^{+}_{\ijsub},  \unconst{{y}}^{-}_{\ijsub} \right]^\top       & \mathrm{(otherwise)}.
\end{array}
\right.
\end{align}
When $\epsilon$ is small enough, the constrained output $\V{y}_{\ijsub}$ perfectly mimics the output by the projection function $\Proj$. Our goal is to modify the feature vector $\unconst{\V{f}}_{\ijsub}$ so that it makes the probabilities as \eref{eq:epsilon} through the softmax activation.

In the SFS layer, we replace the features as 
\begin{equation}
\begin{array}{l@{\hspace{2em}}l}
f^{-}_{\ijsub} := -\Lambda & ((i,j) \in E^+) \\
f^{+}_{\ijsub} := -\Lambda & ((i,j) \in E^-),
\end{array}
\label{eq:rep2_supp}
\end{equation}
where $\Lambda$ is assumed to be large enough. Given modified features $\V{f}_{\ijsub} = [f^{+}_{\ijsub},f^{-}_{\ijsub}]^\top$, the softmax activation $\softmax$ normalizes and converts them to edge probability $\V{y}_{\ijsub}$ as
\begin{equation}
\label{eq:rep_layer_supp}
\V{y}_{\ijsub} = \left\{
\begin{array}{l@{\hspace{2em}}l}
\softmax( [\unconst{{f}}^{+}_{\ijsub}, \:\:-\Lambda\:]^\top)  \quad  & ((i,j)\in E^+)   \\
\softmax( [\:\:-\Lambda\:,      \unconst{{f}}^{-}_{\ijsub}]^\top)       & ((i,j)\in E^-)    \\
\softmax( [\unconst{{f}}^{+}_{\ijsub},  \unconst{{f}}^{-}_{\ijsub}]^\top)       & \mathrm{(otherwise)}.
\end{array}
\right.
\end{equation}

Without loss of generality, we discuss the case in $(i,j)\in E^+$. Substituting \eref{eq:rep_layer_supp} into \eref{eq:feat_softmax} yields
\begin{align}
\begin{array}{ll}
  \V{y}_{\ijsub} &=\left[\frac{\exp(\unconst{f}^{+}_{\ijsub})}{\exp(\unconst{f}^{+}_{\ijsub})+\exp(-\Lambda)},
\frac{\exp(-\Lambda)}{\exp(\unconst{f}^{+}_{\ijsub})+\exp(-\Lambda)}\right]^\top \nonumber \\
&= \left[\frac{\exp(\unconst{f}^{+}_{\ijsub})}{\exp(\unconst{f}^{+}_{\ijsub})+\epsilon'},
\frac{\epsilon'}{\exp(\unconst{f}^{+}_{\ijsub})+\epsilon'}\right]^\top
\quad  ((i,j)\in E^+),
\end{array}
\end{align}
where $\epsilon'=\exp(-\Lambda)\sim 0$ when $\Lambda$ is large enough, leading to $\V{y}_{\ijsub}=[1-\epsilon, \epsilon]^\top$ as in \eref{eq:epsilon} by denoting $\epsilon = \frac{\epsilon'}{\exp(\unconst{f}^{+}_{\ijsub})+\epsilon'}\sim 0$.

\subsection{Detailed analysis}

We describe a detailed analysis of our reparameterization layer. As described in \eqref{eq:rep_layer}, the unconstrained edge feature between $i$ and $j$-th nodes $\unconst{\V{f}}_{\ijsub} = [\unconst{{f}}^{+}_{\ijsub},\unconst{{f}}^{-}_{\ijsub}]^\top$ is converted to constrained prediction of the edge availability $\V{y}_{\ijsub} = [y_{\ijsub}^+,y_{\ijsub}^-]^\top$
by selectively suppressing unwanted feature values.

When using the cross-entropy loss $\Loss_\text{CE}$ to evaluate the availability of the graph edges, the derivative to be backpropagated to the backbone graph generator is\footnote{We omit the subscript $\ijsub$ for simplicity.}
\begin{align}
\label{eq:rep_gradient_supp}
\frac{\partial \Loss_\text{CE}}{\partial \unconst{\V{f}}} &= 
\left\{
\begin{array}{l@{\hspace{2em}}l}
\left[(1-\epsilon)-t^{+}, \:\:\:\:\:\:\:\:\:\:0\:\:\:\:\:\:\:\:\:\:\:\right]^\top & ((i,j)\in E^+)    \\
\left[\:\:\:\:\:\:\:\:\:\:0\:\:\:\:\:\:\:\:\:\:\:, (1-\epsilon)-t^{-}\right]^\top & ((i,j)\in E^-)    \\
\left[\:\:\:y^{+}-t^{+}\:\:\:\:\:, \:\:\:y^{-}-t^{-}\:\:\:\:\:\right]^\top & (\mathrm{otherwise}),
\end{array}
\right.\\
\label{eq:rep_gradient_supp2}
&\sim
\left\{
\begin{array}{l@{\hspace{2em}}l}
\left[\:\:1-t^{+}\:, \:\:\:\:\:\:0\:\:\:\:\:\:\:\right]^\top & ((i,j)\in E^+)    \\
\left[\:\:\:\:\:\:\:0\:\:\:\:\:\:, \:\:1-t^{-}\:\right]^\top & ((i,j)\in E^-)    \\
\left[y^{+}-t^{+}, y^{-}-t^{-}\right]^\top & (\mathrm{otherwise}),
\end{array}
\right.
\end{align}
where $\V{t}=[t^{+},t^{-}]^\top$ denotes the ground truth edge existence and non-existence for the node pair $\ijsub$. 
Our method modifies the computation graph of the network when the MST algorithm does not agree with the output of the graph generation model (\ie, $(i,j) \in E^+\cup E^-$), but in different ways for derivatives of each feature value $\frac{\partial \Loss_\text{CE}}{\partial \unconst{{f^+}}}$ or $\frac{\partial \Loss_\text{CE}}{\partial \unconst{{f^-}}}$.

\begin{table}[t]
\centering
\caption{Case-by-case analyses of our reparameterization layer. For the columns of unconstrained features $\unconst{\V{f}}$, constrained prediction $\V{y}$, and the ground truth edge availability $\V{t}$, the table shows the index of the larger element. For example, the column $\unconst{\V{f}}$ will be $+$ when the edge feature for the positive edge availability is larger, \ie, $\unconst{{f}^+}>\unconst{{f}^-}$. The column $(i,j)$ displays $E^+$ or $E^-$ if the MST algorithm modifies the edge availability (in which the rows are also highlighted). For the remaining columns, $\uparrow$ and $\downarrow$ denote each value becoming (relatively) large or small, respectively.}
\label{tab:supp_analysis}
    \begin{tabular}{c|ccc|c|c|cccc|c}
    \toprule
    \multirow{2}{*}{Case}& \multicolumn{3}{c|}{Feats \& probs} & GT & Loss &  \multicolumn{4}{c|}{Approx. derivatives} & \multirow{2}{*}{\begin{tabular}{c}Modified\\by MST\end{tabular}} \\ 
            & $\unconst{\V{f}}$ & $(i,j)$ & $\V{y}$ & $\V{t}$ &  $\Loss_\text{CE}$ 
            & $\frac{\partial \Loss_\text{CE}}{\partial \unconst{f}^{+}}$ 
            & $\frac{\partial \Loss_\text{CE}}{\partial \unconst{f}^{-}}$
            & $\left|\frac{\partial \Loss_\text{CE}}{\partial \unconst{f}^{+}}\right|$ 
            & $\left|\frac{\partial \Loss_\text{CE}}{\partial \unconst{f}^{-}}\right|$ &\\ 
            \midrule
           1 & $+$    &       & $+$ & $[1,0]^\top$ & $\downarrow$  &$y^+-1$&$y^-$& $\downarrow$  & $\downarrow$ & Unmodified\\
           2 & $+$    &       & $+$ & $[0,1]^\top$ & $\uparrow$    &$y^+$&$y^--1$& $\uparrow$    & $\uparrow$  & Unmodified \\ 
    \rowcolor{lightgray}
           3 & $+$    & $E^-$ & $-$ & $[1,0]^\top$ & $\uparrow$    &$0$&$1$&  $\downarrow$ & $\uparrow$ & \emph{Incorrect}\\ 
    \rowcolor{lightgray}
           4 & $+$    & $E^-$ & $-$ & $[0,1]^\top$ & $\downarrow$  &$0$&$0$&  $\downarrow$ & $\downarrow$ & \emph{Correct}\\ 
    \midrule                                
           5 & $-$    &       & $-$ & $[1,0]^\top$ & $\uparrow$    &$y^+-1$&$y^-$&  $\uparrow$   & $\uparrow$ &Unmodified \\ 
           6 & $-$    &       & $-$ & $[0,1]^\top$ & $\downarrow$  &$y^+$&$y^--1$&  $\downarrow$ & $\downarrow$ &Unmodified\\ 
    \rowcolor{lightgray}
           7 & $-$    & $E^+$ & $+$ & $[1,0]^\top$ & $\downarrow$  &$0$&$0$&  $\downarrow$ & $\downarrow$ & \emph{Correct}\\
    \rowcolor{lightgray}
           8 & $-$    & $E^+$ & $+$ & $[0,1]^\top$ & $\uparrow$    &$1$&$0$& $\uparrow$    & $\downarrow$ & \emph{Incorrect}\\ 
    \bottomrule
    \end{tabular}
\end{table}

\Tref{tab:supp_analysis} summarizes the case-by-case behavior, in which we can categorize the behaviors of the SFS layer into eight cases. Hereafter, we use $E^* \triangleq E^+\cup E^-$. 

\vspace{2mm}\noindent\textit{Case $(i,j)\notin E^*$ (Cases 1, 2, 5, 6)\quad}
When the graph algorithm (\ie, MST) does not modify the edge availability (\ie, cases 1, 2, 5, and 6 in the table), the behavior is the same as the usual cross-entropy loss for unconstrained edges. 

\vspace{2mm}\noindent\textit{Case $(i,j)\in E^* \;\; \& \;\; \V{y}\sim \V{t}$ (Cases 4, 7)\quad}
In these cases, the MST algorithm \emph{correctly} suppresses the unwanted features, where the constrained prediction $\V{y}$ becomes the approximation of the ground-truth edge availability $\V{t}$. The loss value becomes small, and the derivative is $\frac{\partial \Loss_\text{CE}}{\partial \unconst{\V{f}}} \sim \V{0}$. This is natural since our constrained graph generator produces correct predictions.

\vspace{2mm}\noindent\textit{Case $(i,j)\in E^* \;\; \& \;\; \V{y}\nsim \V{t}$ (Cases 3, 8)\quad}
In these cases, MST \emph{incorrectly} modifies the edge availability, \ie, the node pair $(i,j)$ belongs to $E^+$ or $E^-$, but the constrained prediction $\V{y}$ does not fit the ground truth $\V{t}$. 
Here, we mathematically discuss the behavior in these cases. Without loss of generality, we focus on Case 3, where MST \emph{incorrectly} removes an edge and compares the methods with and without tree-graph constraints using the SFS layer.

\paragraph*{Case 3 (MST \emph{incorrectly} removes an edge)}
The following discussions can be straightforwardly extended to Case 8, where MST \emph{incorrectly} adds an edge.

\paragraph*{Conditions:}
\begin{itemize}
    \item Unconstrained prediction (edge exists): $\unconst{f}^+ > \unconst{f}^-$, 
    \item MST \emph{removes} the edge: $(i,j)\in E^-$,
    \item GT edge availability (edge exists): $[t^+,t^-] = [1,0]$.
\end{itemize}

\paragraph*{Unconstrained method (without SFS layer)}
The gradient at $\unconst{f}^+$ by the unconstrained method is 
\begin{align}
    \frac{\partial \Loss_\text{unconst}}{\partial \unconst{f}^{+}} = \frac{\exp(\unconst{f}^+)}{\exp(\unconst{f}^+)+\exp(\unconst{f}^-)}-1.
\end{align}
Since $\unconst{f}^+>\unconst{f}^-$, it takes the value in the range of $(-0.5,0)$. 
Similarly,
\begin{align}
    \frac{\partial \Loss_\text{unconst}}{\partial \unconst{f}^{-}} = \frac{\exp(f^-)}{\exp(f^-)+\exp(f^+)}-0,
\end{align} 
thus the gradient at $\unconst{f}^-$ is inside $(0, 0.5)$. 
Thus, the gradient vector $\frac{\partial \Loss_\text{unconst}}{\partial \unconst{\V{f}}}$ is always shorter than $[-0.5,0.5]^\top$ (corresponding to the special case $\unconst{f}^+=\unconst{f}^-$).

\paragraph*{Constrained method (Ours)}
As described in Eqs.~\eqref{eq:rep_gradient_supp} and \eqref{eq:rep_gradient_supp2}, our method yields the gradient as 
\begin{align}
    \frac{\partial \Loss_\text{const}}{\partial \unconst{f}^{+}} = 0, \qquad
    \frac{\partial \Loss_\text{const}}{\partial \unconst{f}^{-}} = 1-\epsilon \sim 1, \nonumber
\end{align}
\ie, $\frac{\partial \Loss_\text{const}}{\partial \unconst{\V{f}}}\sim [0,1]^\top$.

\paragraph*{Comparisons}
While both methods control the features to increase the edge availability, the relation of gradient vectors $\|\frac{\partial \Loss_\text{const}}{\partial \unconst{\V{f}}}\| > \|\frac{\partial \Loss_\text{unconst}}{\partial \unconst{\V{f}}}\|$ always holds, which means our method strongly penalizes the incorrect estimates by MST by directly comparing the final estimation (\ie, tree graph) with the ground-truth edge availability, which highlights our key motivation---a \textbf{direct} control of the tree-constrained graph generation.

\section{Dataset Details}
\label{sec:supp_dataset}
We describe the details of the datasets used in our experiment. 

\subsection{Domain-specific datasets}

\paragraph*{Synthetic tree pattern dataset}
To prepare the synthetic dataset, we implement a generator of two-dimensional tree patterns based on the L-system~\cite{L-system-papa}, a formal language for describing the growth of the structural form. 
The L-system recursively applies rewriting rules to the current structure to simulate the growth of branching structures.

\begin{figure}[t]
	\centering
	\includegraphics[width=\linewidth]{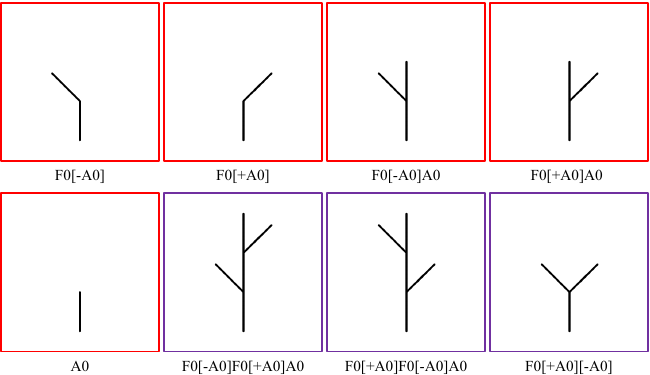}
	\caption{Atomic structures used for synthetic dataset generation. Three pre-defined initial structures are highlighted in purple. Eight pre-defined rewriting rules are used during the generation.}
	\label{fig:Atomic_structures}
\end{figure}

\begin{figure}[t]
	\centering
	\includegraphics[width=\linewidth]{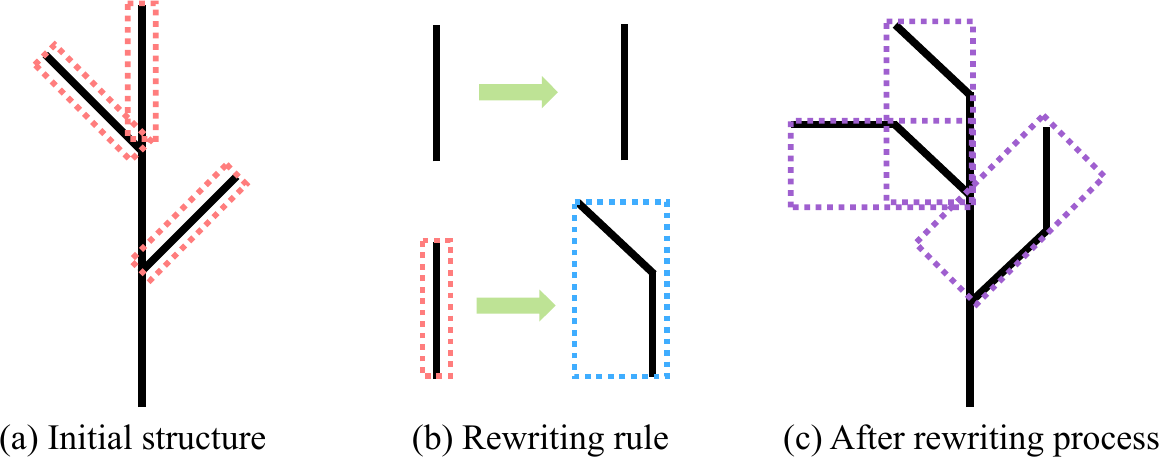}
	\caption{An example of the rewriting process. Suppose the initial structure is represented as \textbf{F0[+A0]F0[-A0]A0}. If a rewrite rule \textbf{F$\rightarrow$F{}; A{}$\rightarrow$F{}[-A{}]} is applied, \ie, \textbf{F} remains unchanged and \textbf{A} becomes \textbf{F{}[-A{}]}, the result of the rewrite process is \textbf{F0[+F1[-A1]]F0[-F1[-A1]]F1[-A1]}. The digits in the sequences indicate the number of times the rewrite is applied.}
	\label{fig:rewrite}
\end{figure}

\Fref{fig:Atomic_structures} shows the initial structures and the rewriting rules we used. At the beginning of the tree generation, an initial sequence is randomly chosen from the pre-defined sequences marked with a purple frame in the figure. At each iteration during the tree generation, the leaf edges (``\textbf{A}'' in the sequences) are replaced by a randomly chosen pattern from eight pre-defined ones. A simple example is shown in \fref{fig:rewrite}. We iterate the rewriting process a maximum of three times to generate a tree pattern.
We also add randomness to the branch length and joint angles in our dataset. We randomly choose a branch length of scaling $[0.5, 2.5]$ and joint angles of $[10^\circ,35^\circ]$. 

\paragraph*{Root dataset}
\label{sec:root-augmentataion}
For the root dataset, the structure of the early-growing roots of Arabidopsis is manually annotated.
The structures are annotated by placing points (\ie, graph nodes) on the root path, where the distance between neighboring points may vary depending on the annotator and the images. We, therefore, resample the graph nodes with the same intervals. Starting from keypoints with the degree $\neq 2$ (\ie, joints and leaf nodes), we sample nodes at intervals of $8$ pixels along continuous branch segments.

For data augmentation, we apply flipping, rotation, cropping, noise, lighting, and scaling on the original images.
Supposing the roots are almost aligned at seeding, we limit the range of rotation angles in $[-9^\circ,+9^\circ]$.

\paragraph*{Grapevine dataset}
\label{sec:grapevine-augmentataion}
We use the 3D2cut Single Guyot Dataset~\cite{Guyot} containing manual annotations on branch structures. We perform data augmentation with rotation angles in $[-15^\circ, +15^\circ]$ in the same manner as~\cite{Guyot}.
This dataset also contains the classification of nodes (four classes) and edges (five classes) related to biological meanings. Since the existing two-stage method~\cite{Guyot} estimates these categories, we follow the same setup for the two-stage baseline method (refer to the next section for detailed discussions). For the other methods, including our PlantPose implementation, we use only the binary class information (\ie, branch availability) for generalizability.

\subsection{Generalized datasets}

\subsubsection{Training datasets}

\bmhead*{Grapevine dataset~\textnormal{\cite{Guyot}}}
For the grapevine dataset, we used the same data described in the domain-specific datasets~\cite{Guyot}. Notably, the grapevine dataset has an average edge length of approximately 12.8 pixels, closely matching our 13-pixel sampling interval. This dataset also contains highly detailed annotations of plant biological joints, which are critical for capturing the structural semantics of plants. To preserve the dataset’s biological accuracy and annotation quality, we did not apply additional resampling to the grapevine dataset. 

We applied extensive augmentations for the grapevine images, including translation, flipping, rotation, brightness adjustment, aspect ratio modification, and scaling. This process expanded the training set tenfold, resulting in $12,020$ training images alongside $150$ validation and $151$ test images.

\bmhead*{LVD2021 leaf image dataset~\textnormal{\cite{LVD2021}}}
The Leaf Vein Dataset 2021 (LVD2021)~\cite{LVD2021} provides high-resolution RGB images of leaf veins with pixel-level annotations. 
The leaf images were captured against a white background, offering a clear visual contrast for leaf structure analysis.
We applied the same augmentation techniques used for the grapevine dataset. This process expanded the training set tenfold, resulting in $18,580$ training images alongside $232$ validation and $232$ test images. The class distribution of the leaf species in the train, validation, and test splits is summarized in \Tref{tab:lvd2021_classes}.

\begin{table}[tp]
\centering
\renewcommand{\arraystretch}{1.0} 
\caption{Leaf category distribution of the LVD2021 leaf image and vein mask datasets across train, validation, and test splits.}
\label{tab:lvd2021_classes}
\setlength{\tabcolsep}{3pt} 
\begin{tabularx}{0.88\textwidth}{l|c|c|c|l|c|c|c}
\toprule
\textbf{Leaf Category} & \textbf{Train} & \textbf{Val} & \textbf{Test} & \textbf{Leaf Category} & \textbf{Train} & \textbf{Val} & \textbf{Test} \\ \midrule
Forsythia Suspensa        & 83  & 13 & 7  & Sycamores              & 48  & 5  & 8  \\
Fructus Xanthii           & 32  & 1  & 6  & Lilac                  & 50  & 10 & 12 \\
Grape                     & 70  & 7  & 12 & Persimmon              & 7   & 2  & 2  \\
Morning Glory             & 80  & 5  & 8  & Smoke Tree             & 105 & 8  & 6  \\ \midrule
Apricot                   & 85  & 7  & 8  & Vitex Negundo Var      & 96  & 13 & 10 \\ 
Chenopodium Album         & 94  & 14 & 13 & Elm                    & 77  & 14 & 12 \\
Callistephus Chinensis    & 50  & 6  & 12 & Holly                  & 45  & 6  & 2  \\
Walnut                    & 69  & 9  & 10 & Poplar                 & 144 & 17 & 18 \\ \midrule
Maple Tree                & 109 & 16 & 9  & Chinese Redbud         & 83  & 16 & 15 \\
Amaranth                  & 104 & 10 & 10 & Hackberry              & 55  & 6  & 8  \\
Honeysuckle               & 52  & 3  & 6  & Crataegus Pinnatifida  & 68  & 8  & 4  \\
Sweet Potato              & 87  & 10 & 11 & Mirabilis Jalapa       & 47  & 8  & 10 \\
Cedar                     & 81  & 16 & 13 &                        &     &    &   \\ \midrule
Mulberry                  & 26  & 2  & -- & Virginia Creeper       & 1   & -- & -- \\
Thistle                   & 9   & -- & -- & Hibiscus               & 1   & -- & -- \\
\bottomrule
\end{tabularx}
\end{table}

\bmhead*{LVD2021 vein mask dataset~\textnormal{\cite{LVD2021}}}
The LVD2021 dataset also includes highly detailed vein masks for structural segmentation. From these vein masks, we extract three graphs by converting the masks into skeletons and identifying endpoints and joints based on the degree of each pixel.
The vein masks are particularly valuable because real-world tree structures are not limited to captured or synthetic plant images but also exist in simplified forms, such as sketches or abstract drawings. Including vein masks in the training regimen enhances the model's robustness to simplified structural inputs like sketches or abstract drawings.
The class distribution of the leaf species in the train, validation, and test splits is the same as the LVD2021 leaf image dataset.
Unlike the leaf images, the vein masks were treated as synthetic data without augmentation, yielding $1,858$ training images, $232$ validation images, and $232$ test images.

\bmhead*{MIPDB dataset~\textnormal{\cite{MIPDB}}}
The MIPDB dataset~\cite{MIPDB}, a maize image-phenotype database, contains over $28,000$ high-resolution images annotated using a precise point-line method. To prepare the dataset for graph-based analysis, we segmented the field images into single-plant views by clustering annotated points, ensuring that each resulting image contained only one complete plant. To prepare the dataset for graph-based analysis, we applied preprocessing steps (detailed in later sections) to convert the field images into single-plant tree-structured graphs. The final dataset consists of $12,992$ training images, $249$ validation images, and $250$ test images. Due to the dataset's large size and diversity, we did not apply data augmentation.

\bmhead*{Self-captured dataset}
We collected images of common plants, including cherry blossoms, ginkgo trees, pine trees, and a few flowers and grasses. The dataset includes $817$ images, divided into $652$ training, $82$ validation, and $83$ test images. Data augmentation expanded the training set to $13,040$ images.

\bmhead*{Web-sourced dataset}
We collected and organized additional data from publicly available images of various plants, including potted, coconut, and lemon trees. The dataset comprises $1,654$ images, divided into $1,320$ training, $166$ validation, and $168$ test images. The training set was augmented tenfold to $13,200$ images.
We summarize the class distribution of the web-sourced dataset across the train, validation, and test splits in \Tref{tab:web_sourced_classes}.

\begin{table}[tp]
\centering
\renewcommand{\arraystretch}{1.0} 
\caption{Plant category distribution of the Web-sourced dataset across train, validation, and test splits.}
\label{tab:web_sourced_classes}
\setlength{\tabcolsep}{4pt}
\begin{tabularx}{0.83\textwidth}{l|c|c|c|l|c|c|c}
\toprule
\textbf{Plant Category} & \textbf{Train} & \textbf{Val} & \textbf{Test} & \textbf{Plant Category} & \textbf{Train} & \textbf{Val} & \textbf{Test} \\ \midrule
Pomegranate      & 144 & 16 & 21 & Mango            & 156 & 20 & 19 \\
Plum             & 125 & 12 & 14 & Almond           & 2   & -- & -- \\
Coconut          & 525 & 67 & 69 & Mulberry         & 1   & -- & -- \\
Apricot          & 192 & 24 & 25 & Cherry           & 1   & -- & -- \\
Lemon            & 173 & 27 & 20 & Fig              & 1   & -- & --  \\
\bottomrule
\end{tabularx}
\end{table}

\paragraph*{Additional preprocessing details}
For the newly captured or collected images, namely the self-captured and web-sourced datasets, we manually annotate the branch paths.

For the MIPDB dataset, the original annotations were not provided as direct graphs but as sequential points along each branch. We convert them to a tree graph in a greedy manner to ensure a coherent tree structure. We first identify the longest branch as the initial tree structure. For every other branch, we compute the distances from its endpoints to each segment of the pre-computed tree structure and connect the branch endpoint closest to the pre-computed tree. By iteratively adding the newly connected branches to the tree and repeating this process, we ensure that the final graph becomes a single connected tree. 

For the LVD2021 leaf image and vein mask datasets, we first skeletonize the vein masks to extract veins as lines. By treating each pixel as a node and connecting adjacent nodes, we convert the masks to tree graphs that preserve the leaf vein topology.

\subsubsection{Out-of-domain test datasets}
The out-of-domain test datasets consist of four distinct sources; images are processed to have their nodes at $13$-pixel intervals, the same as the training datasets, for consistency. Details of each dataset are provided below.

\bmhead*{Thickened L-system dataset}
The thickened L-system dataset shares the same generation rules as the L-system synthetic dataset from the domain-specific datasets. To distinguish this dataset from the fine, curved lines in the LVD2021 vein mask dataset, we increased line thickness and employed straight branches.

\bmhead*{Root dataset}
The root dataset is identical to the root dataset used in the domain-specific datasets. Here, we combine all the data into a single test set for out-of-domain evaluation. To ensure consistency, we applied the same 13-pixel resampling process, reducing the maximum number of nodes from 117 to 71 per graph. This adjustment ensures uniform spacing between nodes, aligning with the structural standards of other datasets.

\bmhead*{Tree synthetic dataset}
We employed a more advanced modular tree framework (mtree)\footnote{\url{https://github.com/MaximeHerpin/modular_tree}, last accessed on December 20, 2024.}, which generates more complex, irregularly branching trees with variable thickness and realistic growth patterns. We rendered these synthetic tree images onto 1,408 natural backgrounds curated by environment-based keyword searches (\eg, forests, meadows, deserts, wetlands, savannahs) under varying conditions (\eg, sunny, shaded, foggy, or rainy). This diverse environmental context enhances the dataset's complexity, allowing the model to adapt to lighting, density, and environmental features.

\bmhead*{LVD with background dataset}
The LVD with background dataset augments the LVD2021 leaf images by placing them against 627 diverse natural backgrounds sourced from publicly available images. We applied various transformations (translation, flipping, rotation, brightness/scale adjustments) to introduce random variation in leaf appearance and positioning. The resulting dataset comprises 500 test images, providing an out-of-domain challenge for the model to adapt to varying backgrounds and leaf distortions.

\bmhead*{Unlabeled web-sourced test dataset}
We also test our PlantPose qualitatively for a variety of plant images sourced online, but drawn from plant species not present in the web-sourced training dataset, to assess our method for a completely unseen distribution. This dataset does not have ground-truth annotations. This dataset was curated specifically to evaluate the generalization capability of our model on out-of-training-distribution data.

\section{Details of Baseline Methods}
\label{sec:supp_baseline}
We describe the implementation details for the baseline methods: The \emph{two-stage} method and the method with the \emph{test-time constraint}. Note the implementation for the other baseline, the \emph{unconstrained} method, is identical to the original RelationFormer~\cite{Relationformer}.

\subsection{Two-stage baseline}
\label{sec:supp_two_stage}
Our experiment implements a two-stage baseline involving skeletonization and graph optimization. This baseline implementation is based on ViNet~\cite{Guyot}, a state-of-the-art plant skeleton estimation method. Since the implementation of \cite{Guyot} is not publicly available, we re-implement the method with reference to the descriptions in the paper. Through the re-implementation, we find room for improvement in the two-stage baseline method.
\Tref{tab:vinet} compares the performance of our two-stage implementation with a naive re-implementation of \cite{Guyot}. The SMD and TOPO scores are the same metrics used in the main paper, and we also compare the mean squared error (MSE) of the first-stage output of both methods. Our implementation achieves a better performance; thus, we use the improved version for our experiment.
In the following, we describe the implementation details.

\begin{table*}[tp]
\centering
\caption{Quantitative comparisons between our re-implementation of \cite{Guyot} and our two-stage baseline implementation. }
\label{tab:vinet}
\resizebox{\linewidth}{!}{
\begin{tabular}{c|c|ccc|c@{\hspace{2mm}}c}
\hline
\multirow{2}{*}{Method} & \multirow{2}{*}{SMD $\downarrow$} & \multicolumn{3}{c|}{TOPO score $\uparrow$} & \multicolumn{2}{c}{MSE $\downarrow$}\\
                      &                   & Prec.      & Rec.       & F1         & Node confidence  & Edge direction \\ \hline
Re-implementation of \cite{Guyot}    & $3.84\times 10^{-3}$      & $0.459$      & $0.365$      & $0.406$      & $6.95\times 10^{-3}$      & $1.01\times 10^{-2}$    \\ 
Our implementation of two-stage method            & $\bm{4.24\times 10^{-4}}$ & $\bm{0.677}$ & $\bm{0.589}$ & $\bm{0.630}$ & \bm{$1.19\times 10^{-3}$} & $\bm{2.67\times 10^{-3}}$   \\ \hline
\end{tabular}
}
\end{table*}

\paragraph*{First stage: Skeletonization}
Similar to ViNet~\cite{Guyot}, the first stage of our implementation outputs the prediction of node and edge positions as single-channel confidence maps and two-channel vector fields (hereafter referred to as node confidence maps and edge direction maps, respectively). This step is similar to a widespread human pose estimation method, \ie, OpenPose~\cite{OpenPose}, which jointly estimates the confidence of person keypoints and the Part Affinity Fields (\ie, two-channel vector fields).

\begin{figure*}[tp]
	\centering
	\includegraphics[width=\linewidth]{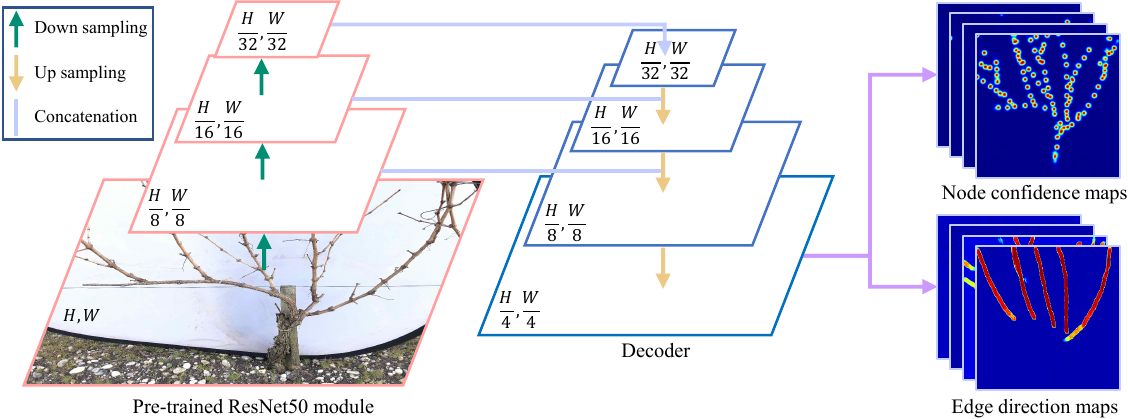}
	\caption{Network architecture for the first (skeletonization) stage of our two-stage baseline method.} 
	\label{fig:supp_two_stage_arch}
\end{figure*}

\begin{figure*}[tp]
	\centering
	\includegraphics[width=\linewidth]{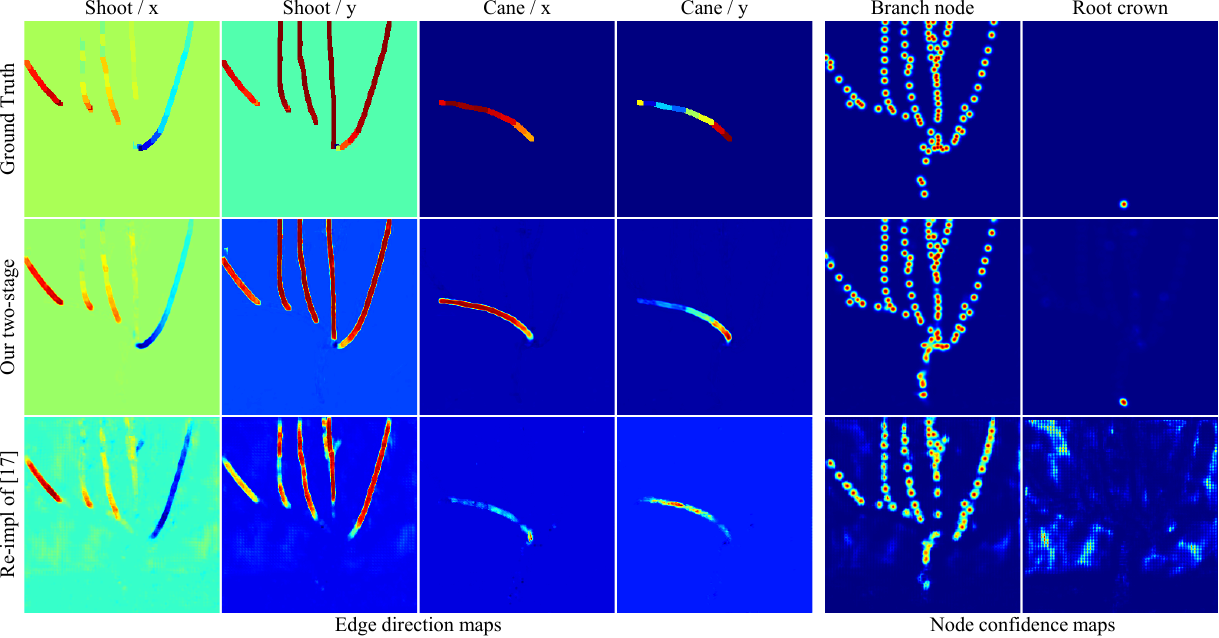}
	\caption{Visual comparisons between a re-implementation of \cite{Guyot} and our two-stage baseline implementation. Our implementation yields better node confidence and edge direction maps, which are the outputs of the first stage of these methods.}
	\label{fig:PAF_conf}
\end{figure*}

While ViNet~\cite{Guyot} uses a sequence of residual blocks followed by the Stacked Hourglass Network~\cite{supp_stacked} for this stage, we use a pre-trained ResNet50~\cite{ResNet} for image feature extraction. This is for a fair comparison to our PlantPose implementation, which also uses ResNet50 as the backbone\footnote{ResNet50 is actually used as the node detection module in RelationFormer (that is based on Deformable DETR~\cite{Deformable_DETR}), which is the basis of our PlantPose, and we inherited its implementation.}. 
We implement an architecture like the Feature Pyramid Network (FPN)~\cite{supp_FPN}, illustrated in \fref{fig:supp_two_stage_arch}, to decode the node \& edge maps from the image features.
\Fref{fig:PAF_conf} visually compares the estimated node \& edge maps, showing a better accuracy by our two-stage implementation.

The original ViNet estimates multiple classes of nodes (four classes) and edges (five classes) as different maps for the grapevine dataset. Compared to just using binary classes (\ie, a branch exists or not), our two-stage implementation also yields better estimation accuracies using the multiple classes (SMD in $4.2 \times 10^{-4}$ with multi-class and $1.4 \times 10^{-2}$ with binary classes). Therefore, we use the multi-class setup for our two-stage implementation of the grapevine dataset. For the other dataset, we use binary classification since we do not have specific class information.

\paragraph*{Second stage: Graph algorithm}
Given the node confidence and edge direction maps, ViNet~\cite{Guyot} first extracts the node positions, followed by the computation of the \emph{resistivity} between each node pair, defined using the edge directions and the Euclidean distance between nodes. The final estimates of the graph structure are generated using the Dijkstra algorithm, where the tree structure is obtained by computing the shortest paths from all nodes to the detected root crown. The resistivity is used as the edge cost for the Dijkstra algorithm.

For the second stage, we follow the method in \cite{Guyot} except for the graph algorithm used; namely, we compute MST instead of the shortest paths given by the Dijkstra algorithm, since using MST reduces the SMD metric to $4.2 \times 10^{-4}$, compared to $5.9 \times 10^{-4}$ using the Dijkstra algorithm for the grapevine dataset.

\begin{table*}[t]
\centering
\caption{Parameters used for the two-stage baseline method. $d$ denotes the distance threshold for the local maximum value search (\ie, non-maximum suppression) of node candidates. $\tau_m$ and $\tau_n$ are used as the thresholds for node detection from the confidence maps. For the detailed definitions, refer to the original paper~\cite{Guyot}.}
\label{tab:supp_params}
\resizebox{\linewidth}{!}{
\begin{tabular}{c|c|c|c|c|c|p{20mm}c}
\hline
\multirow{2}{*}{Dataset} & Image size   & Map size     & Node confidence & Edge direction & Node search distance & \multicolumn{2}{c}{Thresholds for node detection}  \\
                         & (W, H) [px]  & (W, H) [px]  &  diameter [px]  &  width [px]          & $d$ [px]    &  \centering $\tau_m$ & $\tau_n$        \\ \hline
Synthetic                & 512, 512     & 512, 512     & 4               & 5                    & 9     & \centering 0.97                   & 0.5         \\ 
Root                     & 570, 190     & 570, 190     & 3               & 5                    & 7     &\centering 0.99                   & 0.3         \\ 
Grapevine                & 1008, 756    & 256, 256     & 3               & 10                   & 25    &\centering 0.97                   & 0.5         \\ \hline
\end{tabular}
}
\end{table*}

\paragraph*{Detailed parameter settings}
The two-stage method involves heuristic parameters for node and edge detection. Therefore, we empirically select the best parameter sets for each dataset. \Tref{tab:supp_params} lists the detailed parameters. In particular, for the root dataset, we need to carefully tune some hyperparameters (namely, $d$, $\tau_m$, and $\tau_n$ in the table) to yield reasonable estimates, where the configurations yielding the best SMD scores are reported in the main paper. 

\subsection{Test-time constraint baseline}
\label{sec:supp_test_time}
For the test-time constraint baseline, we apply MST only in the inference phase, where the graph generator is trained using the same procedure as the unconstrained method. The MST used in this baseline method is identical to our proposed one. 

\section{Performance Analysis}
\label{sec:performance_analysis}
In this section, we present a detailed analysis of the performance of our proposed method. The effectiveness of our method is evaluated through comprehensive experiments in different scenarios. Specifically, we compare our method with the Auto-regressive (AR) model, and we also analyze the performance when our method is applied solely during the training process. 

\subsection{Comparison with auto-regressive method}
\label{sec:auto}
While we implement the tree-graph constraint on the state-of-the-art non-autoregressive graph generator, RelationFormer~\cite{Relationformer}, other choices of constrained graph generation are viable. 
Existing works aiming for tree-constrained graph generation, such as in molecule structure estimation~\cite{SpanningTreeMolecules, TreeMolecules_father, TreeMolecules_1}, use auto-regressive graph generation.
AR methods are a simpler choice for imposing the constraint since it is relatively straightforward to implement the tree-graph constraint in their graph development process. However, since the auto-regressive methods generate graph nodes and edges progressively, they are prone to breakdowns due to changes in the output order or errors during the generation. This tendency is particularly pronounced for relatively large graphs, including our setup.

To assess the potential of auto-regressive methods, we test the state-of-the-art transformer-based auto-regressive graph generator, Generative Graph Transformer (GGT)~\cite{GGT}. \Tref{tab:GGT} compares our method and several variances of GGT on the synthetic tree pattern dataset. The results show that the accuracy of GGT falls short compared to our method, although the vanilla GGT (the top row) mostly outputs tree graphs ($92~\%$) without explicitly imposing the tree-graph constraint. We identified that errors by GGT occurring at a particular step in the auto-regressive generation process continuously cause errors in the sequence of the following generations. The GGT was initially designed for small datasets, specifically for graphs with $\left | V \right | \le 10$. For our setup, where $\left | V \right | \geq 100$, generating these long sequences in a specific order presents a significant challenge.

\begin{table}[t]
\centering
\caption{Comparisons of different graph generation models (\ie, RelationFormer~\cite{Relationformer} and GGT~\cite{GGT}) on the synthetic dataset. }
\label{tab:GGT}
\begin{tabular}{c|c|ccc|c}
\hline
 \multirow{2}{*}{Method}& \multirow{2}{*}{SMD $\downarrow$} & \multicolumn{3}{c|}{TOPO score $\uparrow$} & Tree rate \\
                                             &                           & Prec.     & Rec.     & F1       & [\%]   \\ \hline
 GGT~\cite{GGT}                & $2.71 \times 10^{-3}$     & 0.635     & 0.537    & 0.582    & 92.06     \\
 GGT w/ test-time constraint   & $4.13 \times 10^{-3}$     & 0.620     & 0.545    & 0.580    & 98.10     \\
 GGT w/ SFS layer              & $2.80 \times 10^{-3}$     & 0.652     & 0.584    & 0.616    & 99.63    \\ \hline 
 RelationFormer~\cite{Relationformer} w/ SFS layer (Ours)& $4.78 \times 10^{-6}$     & 0.986     & 0.968    & 0.977    & 100.0      \\ \hline
\end{tabular}
\end{table}

\subsection{Effectiveness of tree-constraint during training}
To assess whether our SFS layer (positively) affects the training process itself or not, we evaluate our method \emph{without} using the SFS layer and the MST algorithm during the inference phase, \ie, introducing constraint only during the training phase (called  \emph{train-time constraint} hereafter). \Tref{tab:traintime} summarizes the performances. Inducting the tree constraint during the training phase mostly outperforms the methods without constraints, meaning that the improvement by our method is based on network improvement by the loss propagated via the SFS layer. 
We also checked the change in the accuracy metric during training, and found our method consistently achieved better accuracy from the beginning of the training. 

\begin{table}[t]
\centering
\caption{Quantitative results with additional baseline, \emph{train-time constraint}.} 
\label{tab:traintime}
\begin{tabular}{c|c|c|c|ccc|c}
\toprule
\multirow{2}{*}{Dataset}   & \multirow{2}{*}{Method} & \multirow{2}{*}{SFS} & \multirow{2}{*}{SMD $\downarrow$} & \multicolumn{3}{c|}{TOPO score $\uparrow$} & Tree rate \\
                           &                      &                &                           & Prec. & Rec. & F1              & [\%]   \\ \midrule 
\multirow{4}{*}{Synthetic} & Unconstrained        &                & $1.43 \times 10^{-5}$     &0.978         &0.929         &0.953            &36.2      \\ 
                           & Test-time &                & $6.26 \times 10^{-6}$     &0.977         &0.953         &0.965           &\textbf{100.0}     \\
                           & Train-time & \checkmark     & $8.44 \times 10^{-6}$     &\textbf{0.987}&0.954         &0.970           &56.5     \\
                           & Ours                 & \checkmark     & \bm{$4.78 \times 10^{-6}$}&0.986         &\textbf{0.968}&\textbf{0.977}  &\textbf{100.0}     \\ \midrule
\multirow{4}{*}{Root}      & Unconstrained        &                & $1.19 \times 10^{-4}$     &0.831         &0.633         &0.719           &35.9     \\
                           & Test-time  &                & $1.52 \times 10^{-4}$     & 0.829         &0.771        &0.799           &\textbf{100.0}     \\
                           & Train-time & \checkmark     & \bm{$7.81 \times 10^{-5}$}&0.853         &0.619         &0.718          &37.2     \\
                           & Ours                 & \checkmark     & $8.82 \times 10^{-5}$     &\textbf{0.861}&\textbf{0.807}&\textbf{0.833}   &\textbf{100.0}  \\ \midrule
\multirow{4}{*}{Grapevine} & Unconstrained        &                & $1.45 \times 10^{-4}$     &0.963         &0.559         &0.708           &0.0      \\
                           & Test-time  &                & $1.47 \times 10^{-4}$     &0.896         &0.840         &0.867            &\textbf{100.0}     \\
                           & Train-time & \checkmark     & $1.30 \times 10^{-4}$     &\textbf{0.965}&0.566         &0.713           &0.0     \\
                           & Ours                 & \checkmark     & \bm{$1.03 \times 10^{-4}$}&0.899         &\textbf{0.843}&\textbf{0.870}  &\textbf{100.0}     \\ \bottomrule
\end{tabular}
\end{table}

\section{Other Design Choices}
\label{sec:supp_ablation}
The experiments in the main paper already provide some ablation studies, namely, comparisons of our method with 1) graph generation without constraint (\emph{unconstrained}), and 2) a method without using MST in the training loop (\emph{test-time constraint}). Here, we delve further into the potential design choices of our PlantPose model.

\subsection{Other graph generators}
Although the proposed module, the SFS layer, can be easily integrated into graph generators other than RelationFormer~\cite{Relationformer}, we found that no methods but our PlantPose implementation achieve satisfactory results. Here, we discuss the results of the implementation of our method to the AR graph generator, GGT~\cite{GGT}, which achieves the second-best accuracy for multiple datasets following the state-of-the-art RelationFormer.

\Tref{tab:GGT} in the last section compares the GGT with and without the tree-graph constraint. Compared to the GGT with test-time MST, using our SFS layer on top of GGT improves both SMD and TOPO scores\footnote{GGT w/ SFS layer does not achieve $100$~[\%] tree rate because it sometimes fails to generate any graphs.}, although the accuracy is insufficient in practice due to the drawback of AR-based generation processes discussed above. Using the newer RelationFormer model significantly improves the estimation accuracy, which implies that our SFS layer will benefit from the future development of graph generation models.

\subsection{Using node distances for edge cost in MST}
Although our proposed method uses the edge non-existence probabilities $\{\unconst{y}^{-}_{\ijsub}\}$ as the edge cost for the MST algorithm, inspired by the two-stage method that uses node distance for the edge cost computation, we multiply the Euclidean distance between nodes by our original edge cost.

As a result, SMD with modified edge cost does not improve accuracy (it achieves the same SMD as our method in the Grapevine dataset).
A possible reason for this is that the graph generator itself can take the node distance into account when estimating graph edges.
Therefore, we simply use the edge non-existence probabilities $\{\unconst{y}^{-}_{\ijsub}\}$ as the edge cost for our method. 

\begin{table}[ht]
\centering
\caption{Ablation for $\Lambda$.}
\label{tab:ablation}
\begin{tabular}{c|c|ccc}
\hline
 \multirow{2}{*}{$\Lambda$}& \multirow{2}{*}{SMD $\downarrow$} & \multicolumn{3}{c}{TOPO score $\uparrow$}  \\
                              &                           & Prec.     & Rec.     & F1         \\ \hline
$2$ $(\exp(-\Lambda) = 1.4\times 10^{-1})$         & $1.51 \times 10^{-4}$     & $0.871$     & $0.803$    & $0.836$         \\ 
$5$ $(\exp(-\Lambda) = 6.7\times 10^{-3})$         & $1.27 \times 10^{-4}$     & $0.866$     & $0.799$    & $0.831$         \\ 
$10$ $(\exp(-\Lambda) = 4.5\times 10^{-5})$         & $\mathbf{1.03 \times 10^{-4}}$     & $\mathbf{0.899}$     & $\mathbf{0.843}$    & $\mathbf{0.870}$         \\ 
$100$ $(\exp(-\Lambda) = 3.7\times 10^{-44})$       & $1.07 \times 10^{-4}$     & $0.886$    & $0.830$    & $0.857$        \\ \hline
\end{tabular}
\end{table}

\subsection{\texorpdfstring{Ablation for $\Lambda$}{Ablation for Lambda}}
An important hyperparameter in our method is $\Lambda$, which controls the level of suppression for unwanted features.
Here, we report an ablation study for this parameter using the grapevine dataset. 
\Tref{tab:ablation} shows that our choice ($\Lambda=10$) achieves better, while the changes in $\Lambda$ do not significantly affect the overall accuracy as long as $\exp(-\Lambda)$ is close enough to zero. This result indicates that our method is robust to the hyperparameter setting.

\section{Additional Visual Results}
\label{sec:supp_results}

\subsection{Results on domain-specific training}
\label{sec:supp_results_specific}
\begin{figure*}[tp]
	\centering
  \includegraphics[width=\textwidth,height=\dimexpr\textheight-2\baselineskip\relax,keepaspectratio]{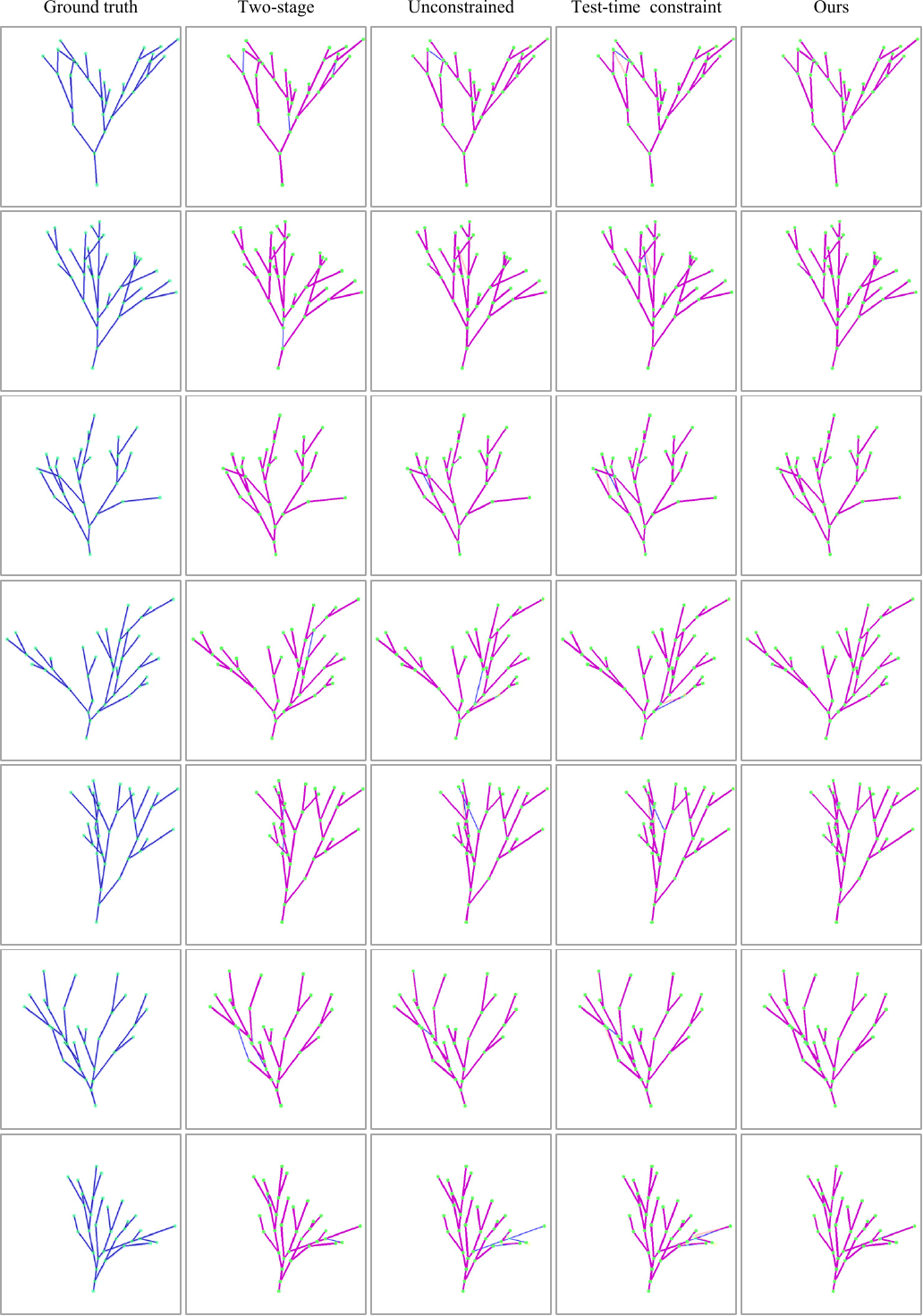}    
	\caption{Additional results for the synthetic branch pattern dataset (domain-specific training).}
    \label{fig:supp_result_syn}
\end{figure*}

\begin{figure*}[tp]
	\centering
  \includegraphics[width=\textwidth,height=\dimexpr\textheight-2\baselineskip\relax,keepaspectratio]{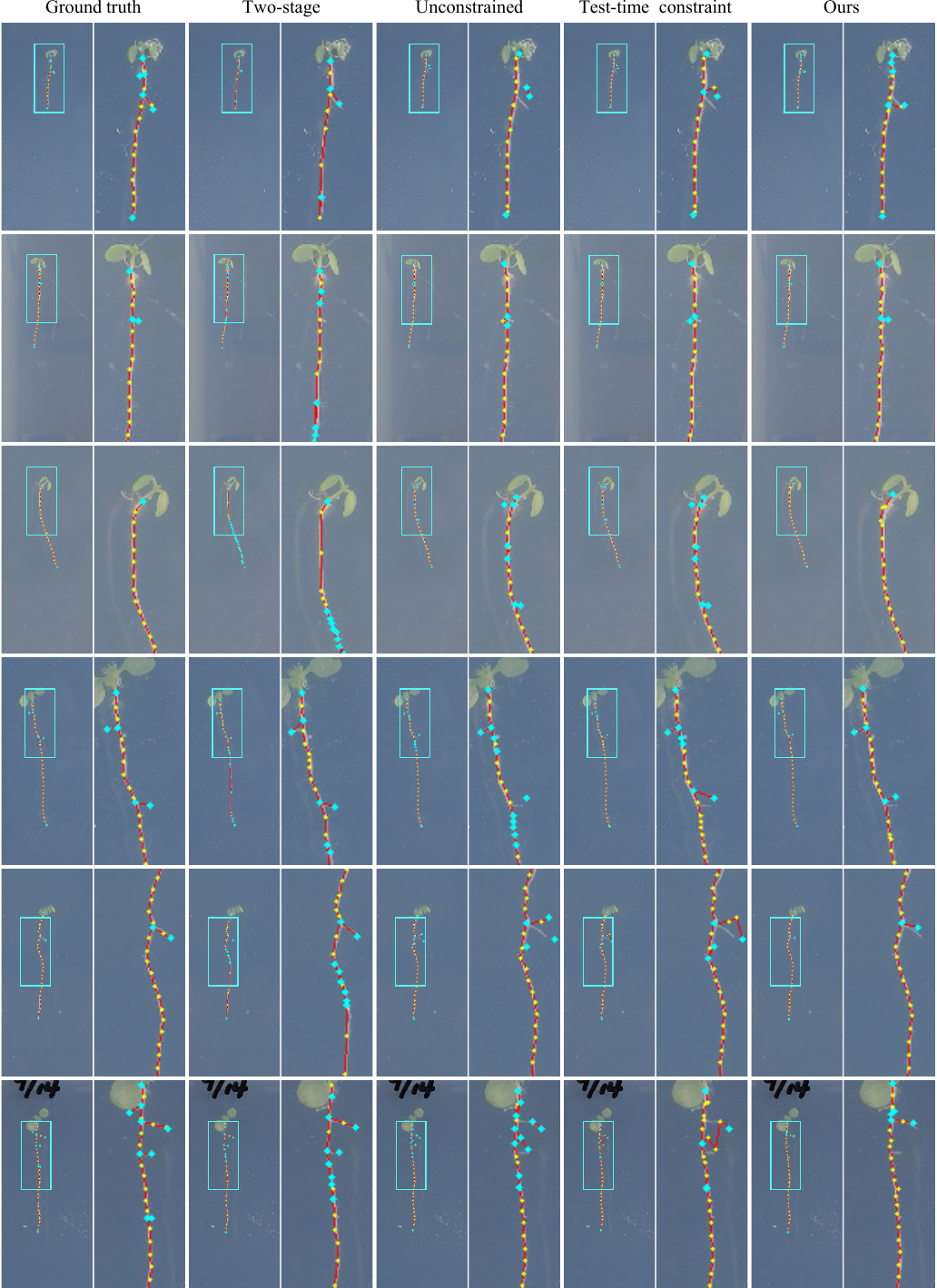}    
	\caption{Additional results for the root dataset (domain-specific training).}
    \label{fig:supp_result_root}
\end{figure*}

\begin{figure*}[tp]
	\centering
  \includegraphics[width=\textwidth,height=\dimexpr\textheight-2\baselineskip\relax,keepaspectratio]{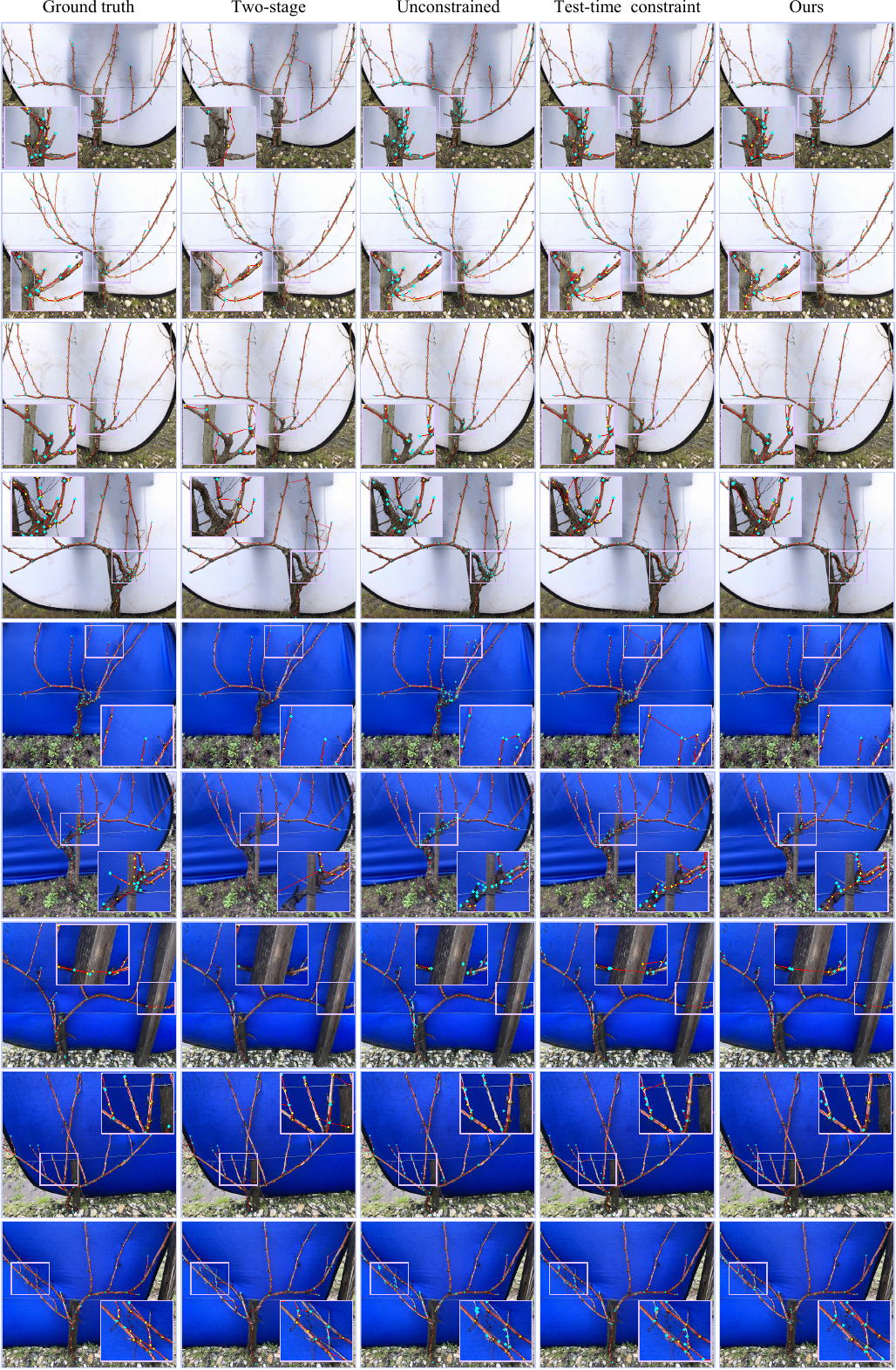}    
	\caption{Additional results for the grapevine dataset (domain-specific training).}
    \label{fig:supp_result_grapevine1}
\end{figure*}

\begin{figure*}[tp]
	\centering
  \includegraphics[width=\textwidth,height=\dimexpr\textheight-2\baselineskip\relax,keepaspectratio]{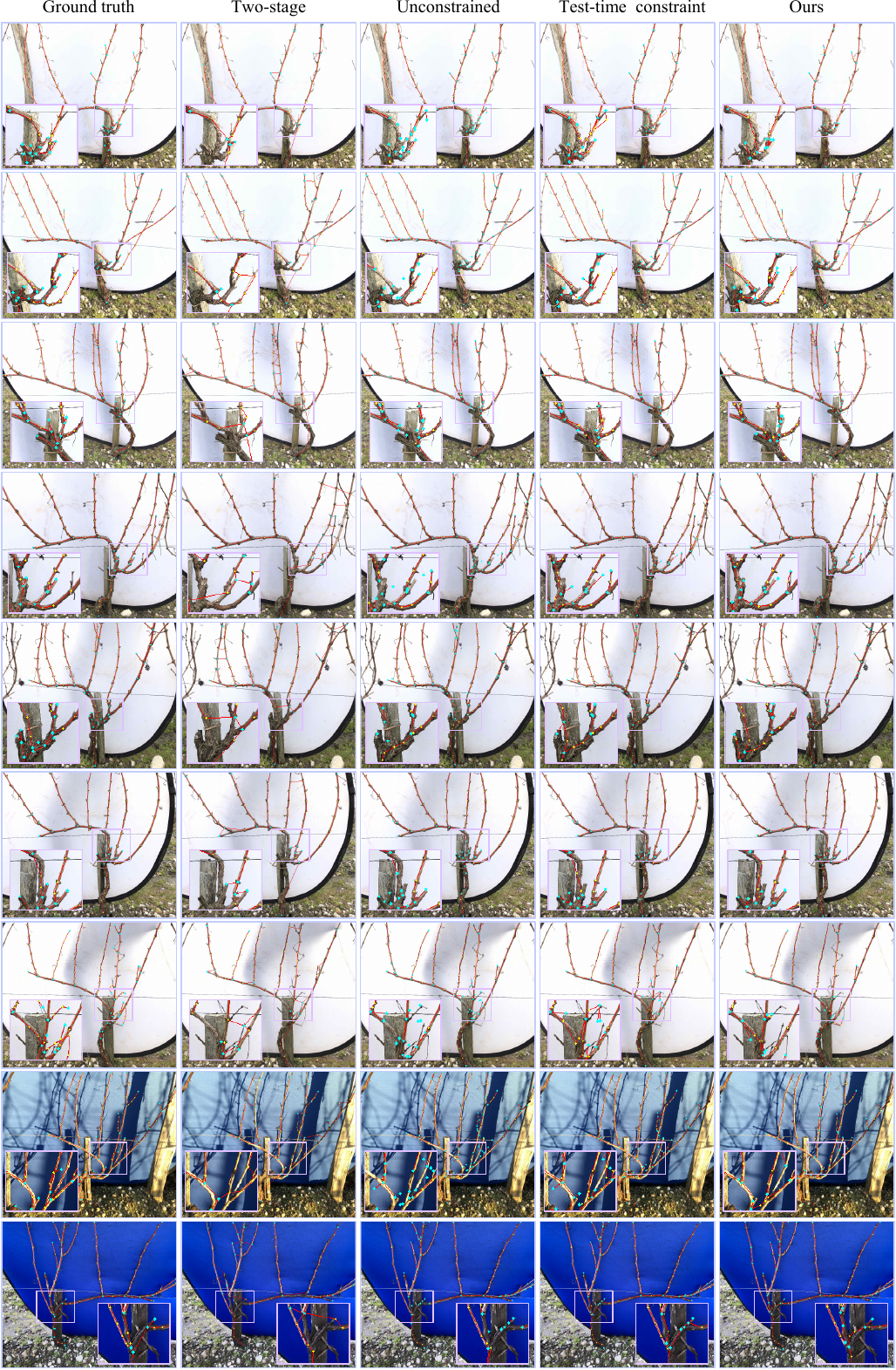}    
	\caption{Additional results for the grapevine dataset (domain-specific training, cont'd).}
    \label{fig:supp_result_grapevine2}
\end{figure*}

\begin{figure*}[tp]
	\centering
  \includegraphics[width=\textwidth,height=\textheight,keepaspectratio]{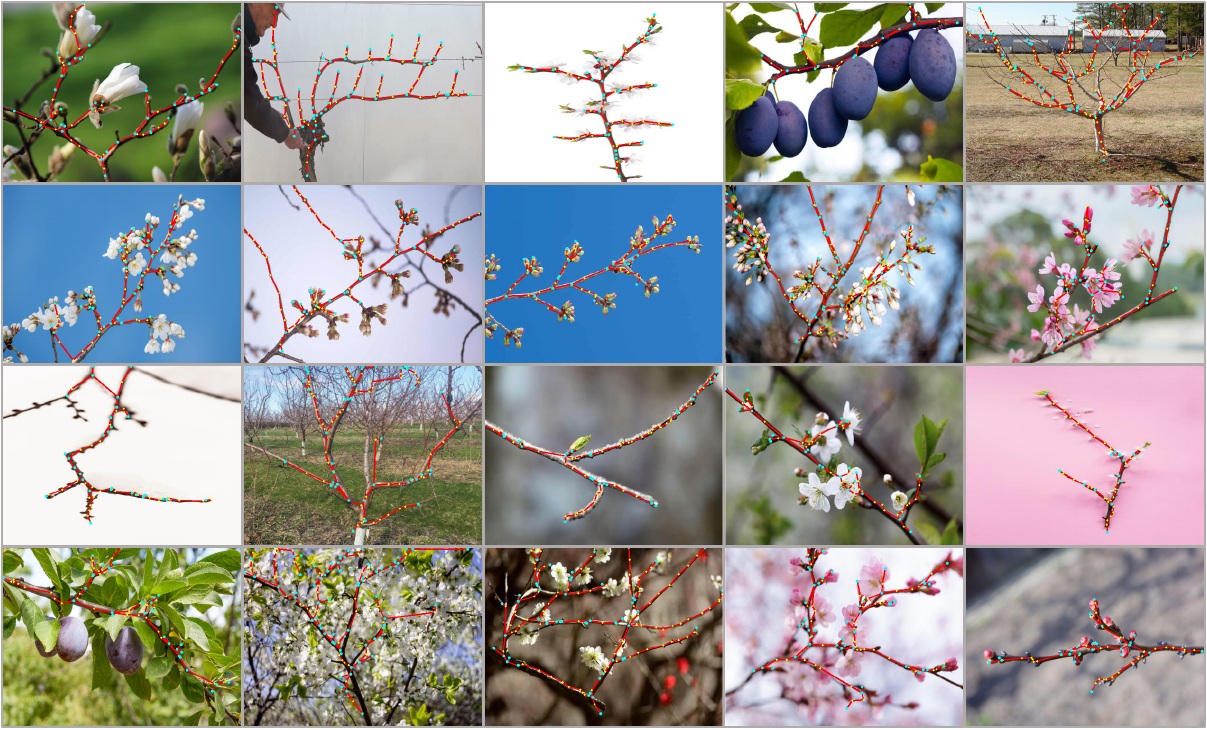}    
	\caption{Additional results for the out-of-domain test dataset (trained on the grapevine dataset).}
    \label{fig:supp_result_domain}
\end{figure*}

We first present the results of domain-specific training.  
Figures~\ref{fig:supp_result_syn} and \ref{fig:supp_result_root} show the additional results for the synthetic and root datasets, respectively.
Figures~\ref{fig:supp_result_grapevine1} and \ref{fig:supp_result_grapevine2} show the results for the grapevine dataset.
Figure~\ref{fig:supp_result_domain} shows the results for out-of-domain testing.

These results consistently demonstrate the high-fidelity estimation of plant skeletons by our PlantPose, which uses the SFS layer that incorporates the constraints while training graph generation models.

\subsection{Results on generalized training}
\label{sec:supp_results_generalized}
\begin{figure*}[tp]
	\centering
  \includegraphics[width=\textwidth,height=\dimexpr\textheight-2\baselineskip\relax,keepaspectratio]{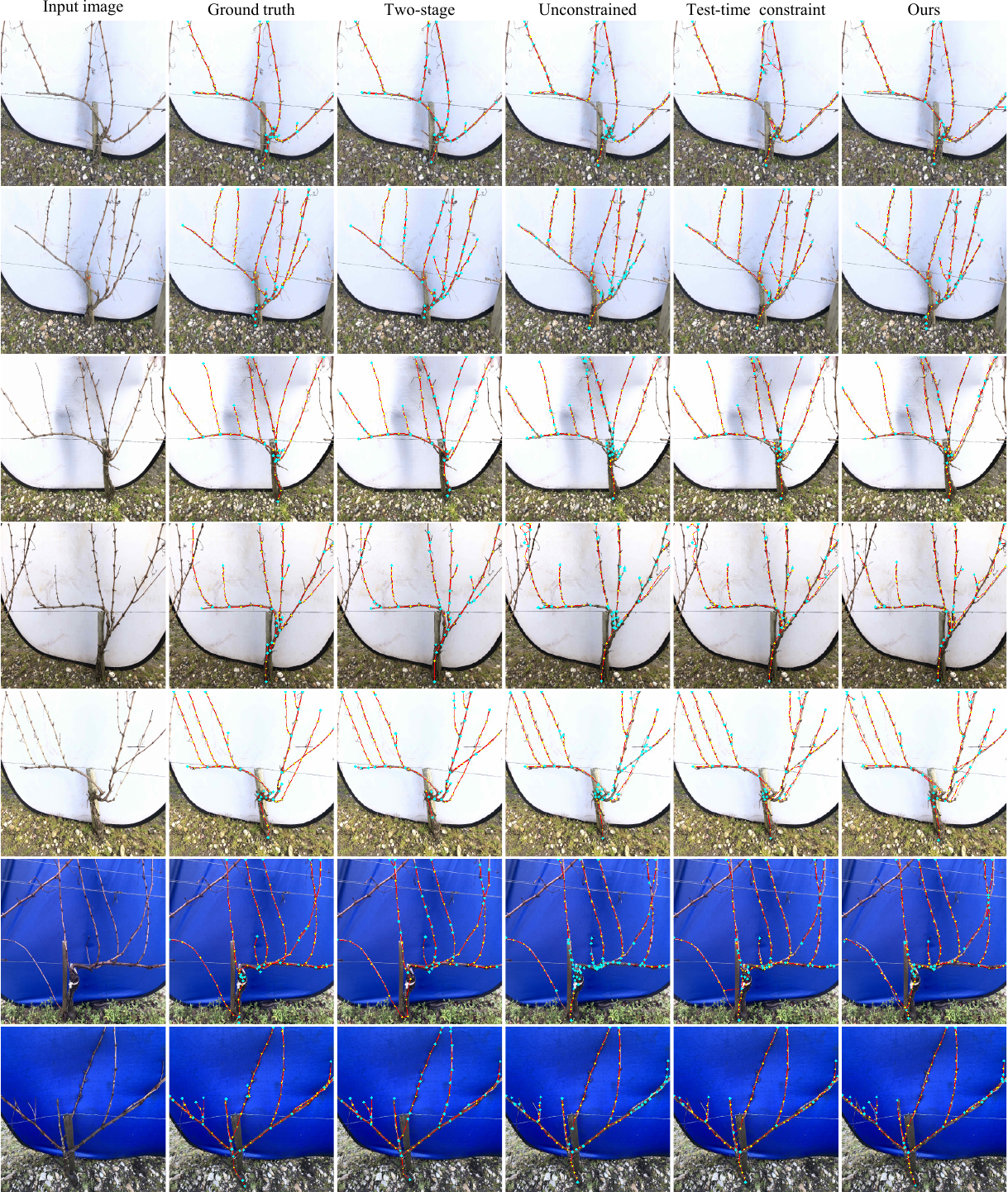}    
	\caption{Additional results for the grapevine dataset (generalized training).}
    \label{fig:supp_result_domain_guyot}
\end{figure*}

\begin{figure*}[tp]
	\centering
  \includegraphics[width=\textwidth,height=\dimexpr\textheight-2\baselineskip\relax,keepaspectratio]{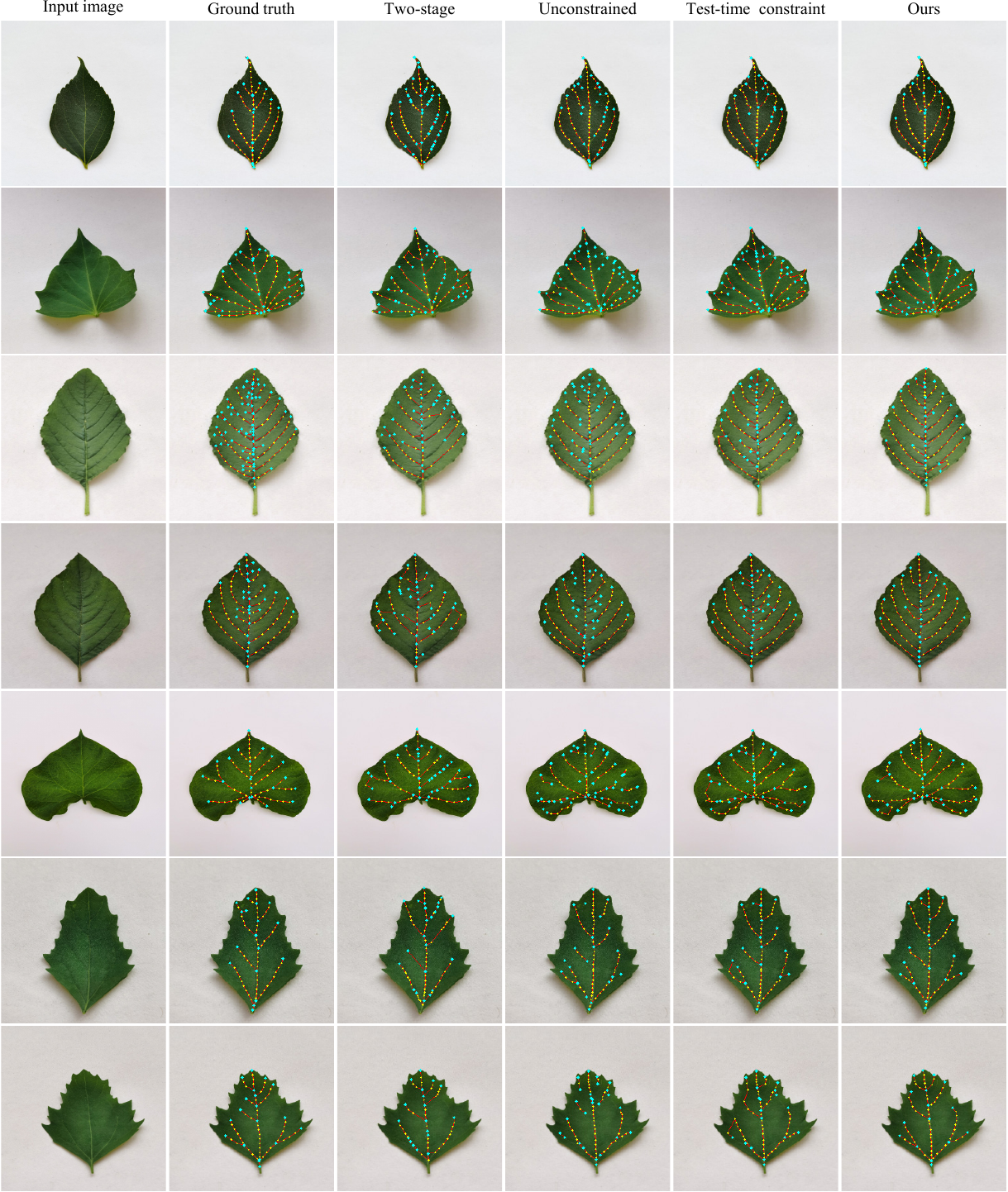}    
	\caption{Additional results for the LVD2021 leaf image dataset (generalized training).}
    \label{fig:supp_result_domain_leaf}
\end{figure*}

\begin{figure*}[tp]
	\centering
  \includegraphics[width=\textwidth,height=\dimexpr\textheight-2\baselineskip\relax,keepaspectratio]{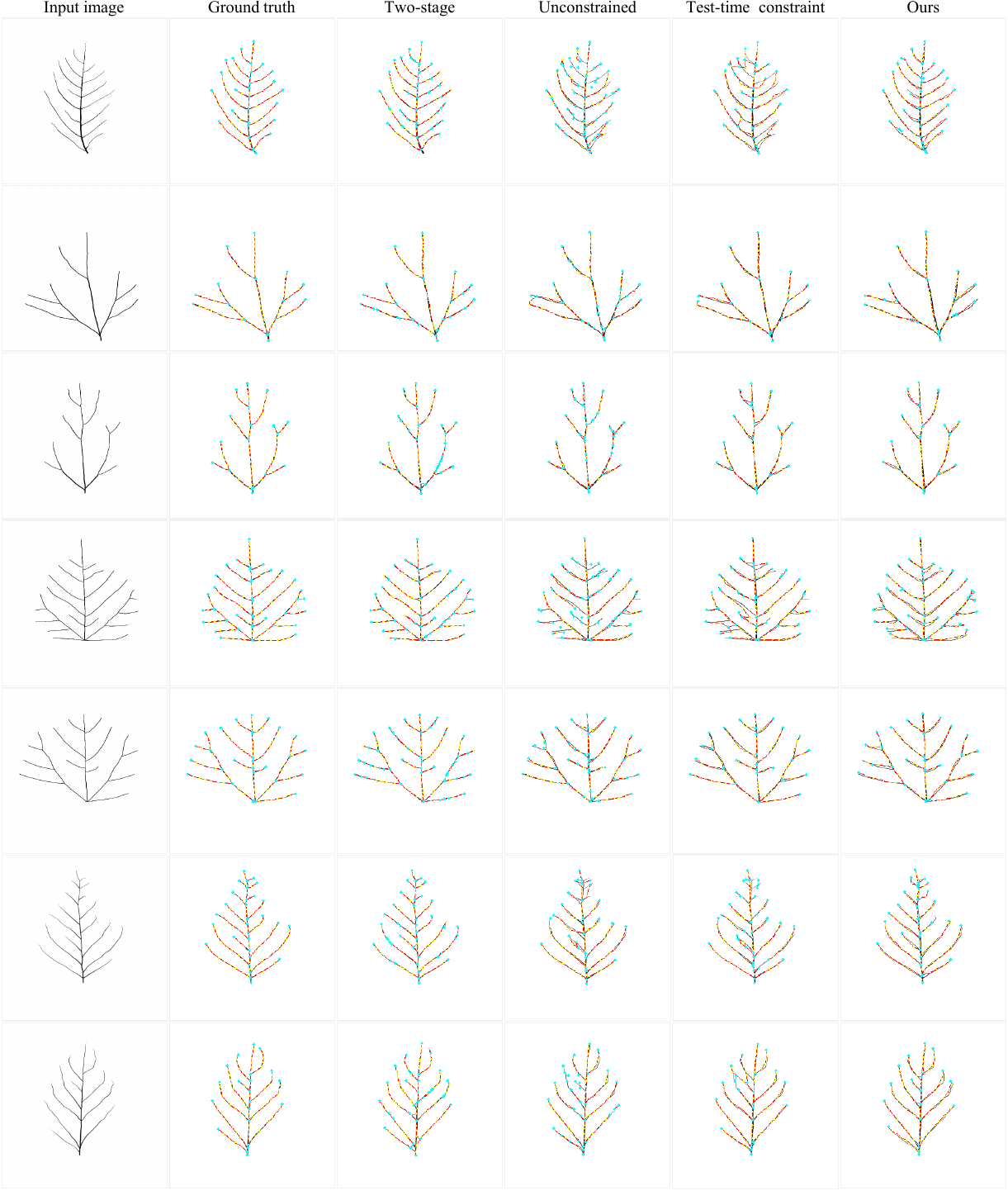}    
	\caption{Additional results for the LVD2021 leaf vein mask dataset (generalized training).}
    \label{fig:supp_result_domain_leaf_mask}
\end{figure*}

\begin{figure*}[tp]
	\centering
  \includegraphics[width=\textwidth,height=\dimexpr\textheight-2\baselineskip\relax,keepaspectratio]{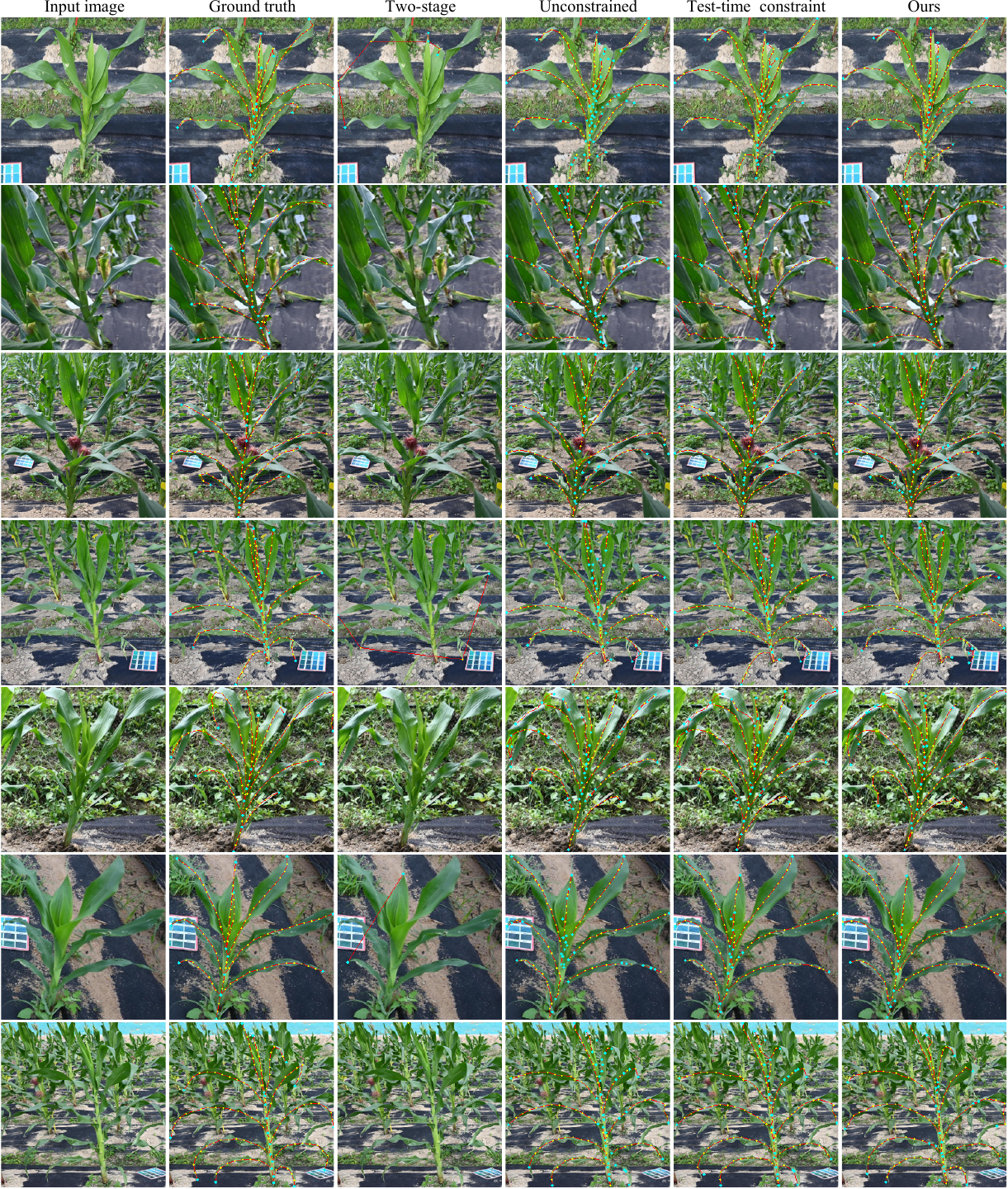}    
	\caption{Additional results for the MIPDB dataset (generalized training).}
    \label{fig:supp_result_domain_MIPDB}
\end{figure*}

\begin{figure*}[tp]
	\centering
  \includegraphics[width=\textwidth,height=\dimexpr\textheight-2\baselineskip\relax,keepaspectratio]{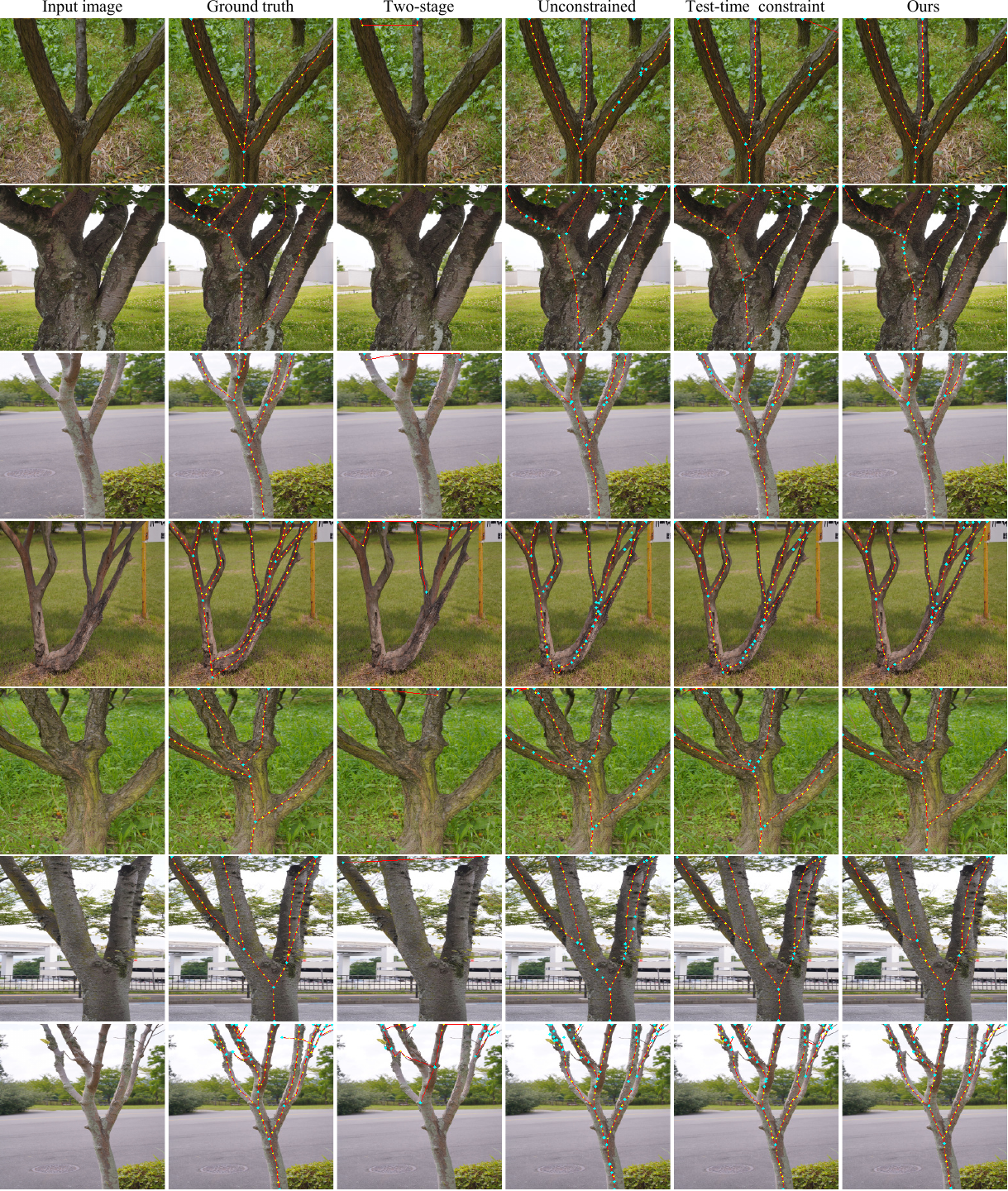}    
	\caption{Additional results for the self-captured dataset (generalized training).}
    \label{fig:supp_result_domain_toda_cam}
\end{figure*}

\begin{figure*}[tp]
	\centering
  \includegraphics[width=\textwidth,height=\dimexpr\textheight-2\baselineskip\relax,keepaspectratio]{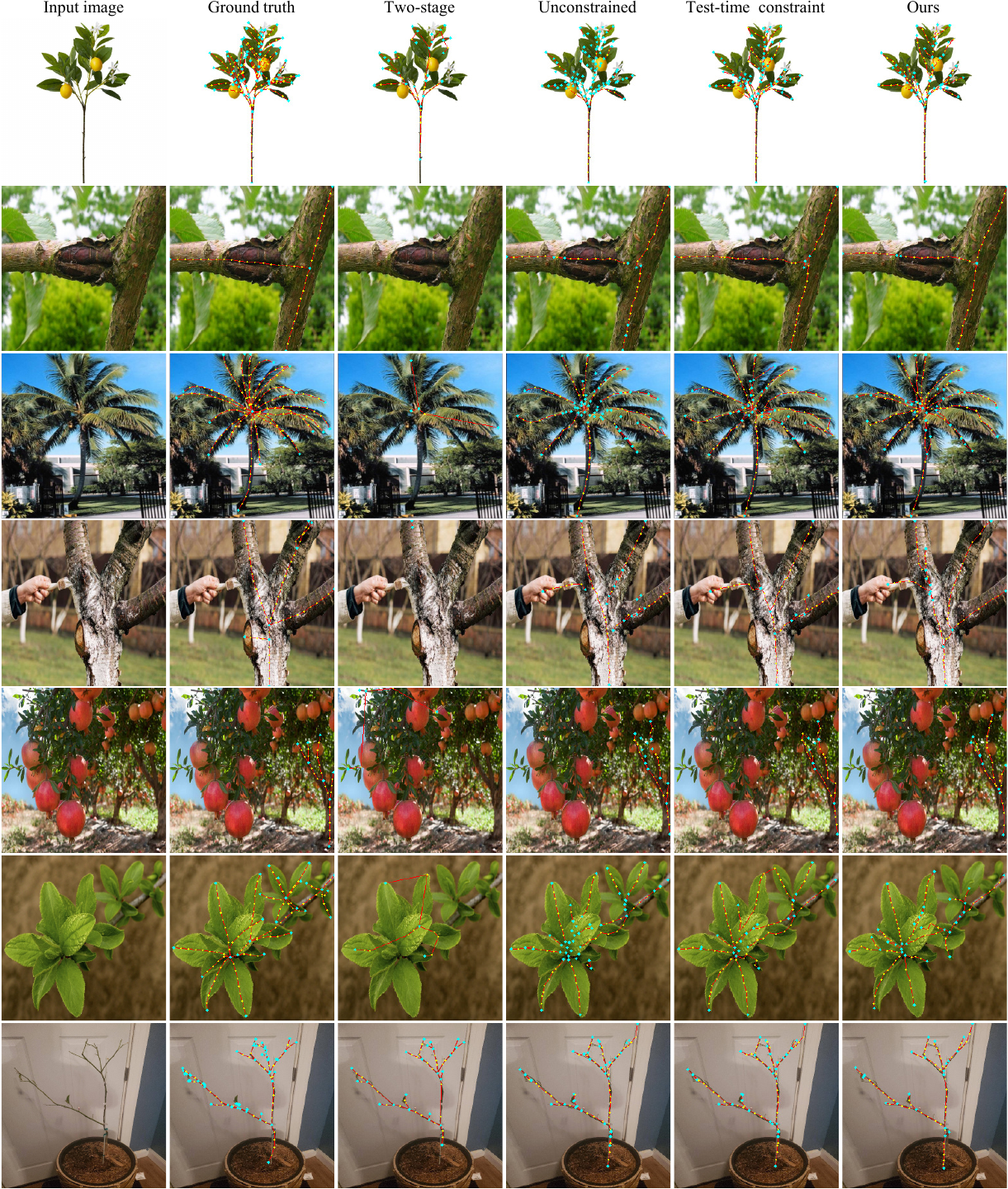}    
	\caption{Additional results for the web-sourced dataset (generalized training).}
    \label{fig:supp_result_domain_toda_world}
\end{figure*}

\begin{figure*}[tp]
	\centering
  \includegraphics[width=\textwidth,height=\dimexpr\textheight-2\baselineskip\relax,keepaspectratio]{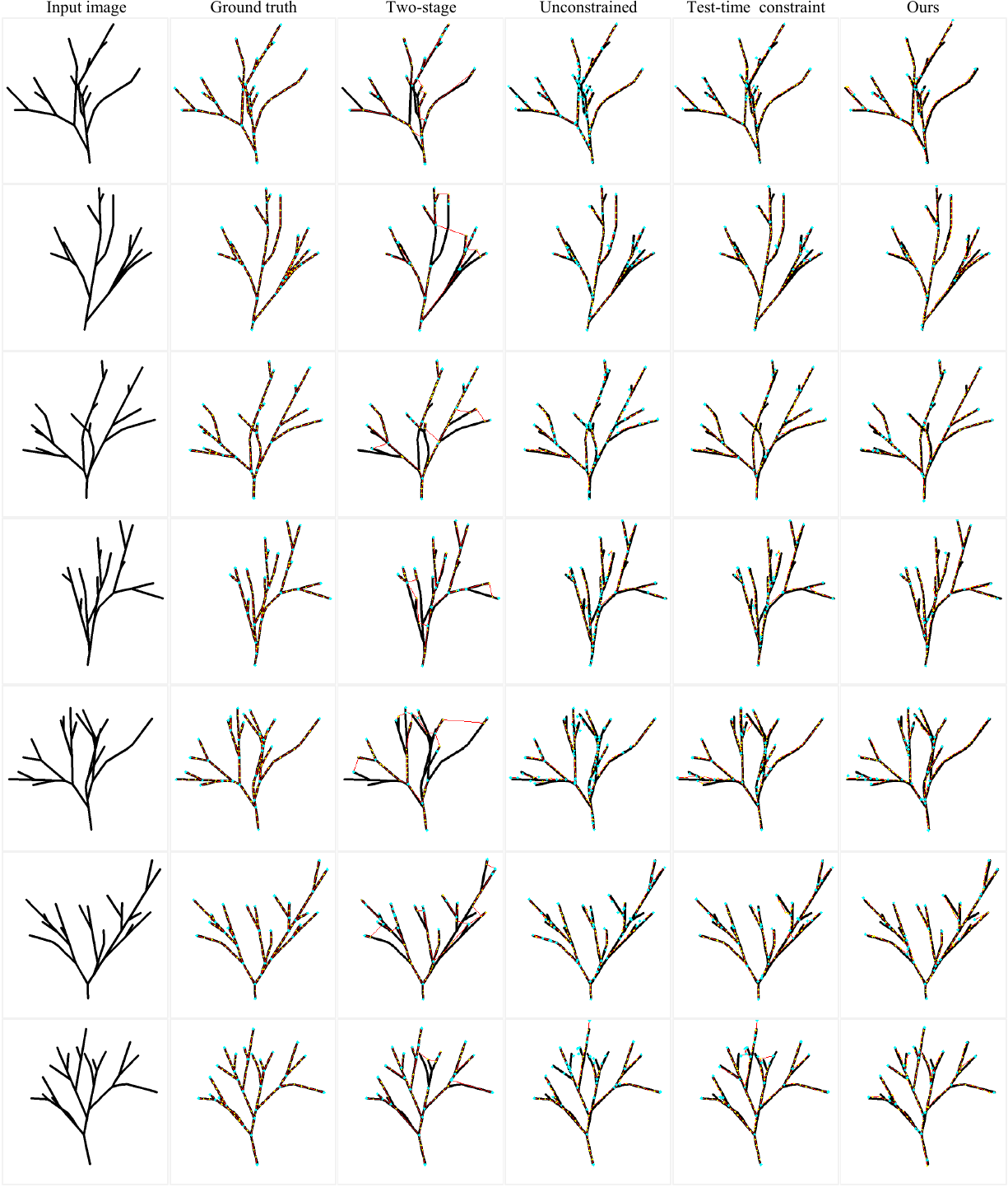}    
	\caption{Additional results for the thickened L-system dataset (generalized training).}
    \label{fig:supp_result_domain_lsystem}
\end{figure*}

\begin{figure*}[tp]
	\centering
  \includegraphics[width=\textwidth,height=\dimexpr\textheight-2\baselineskip\relax,keepaspectratio]{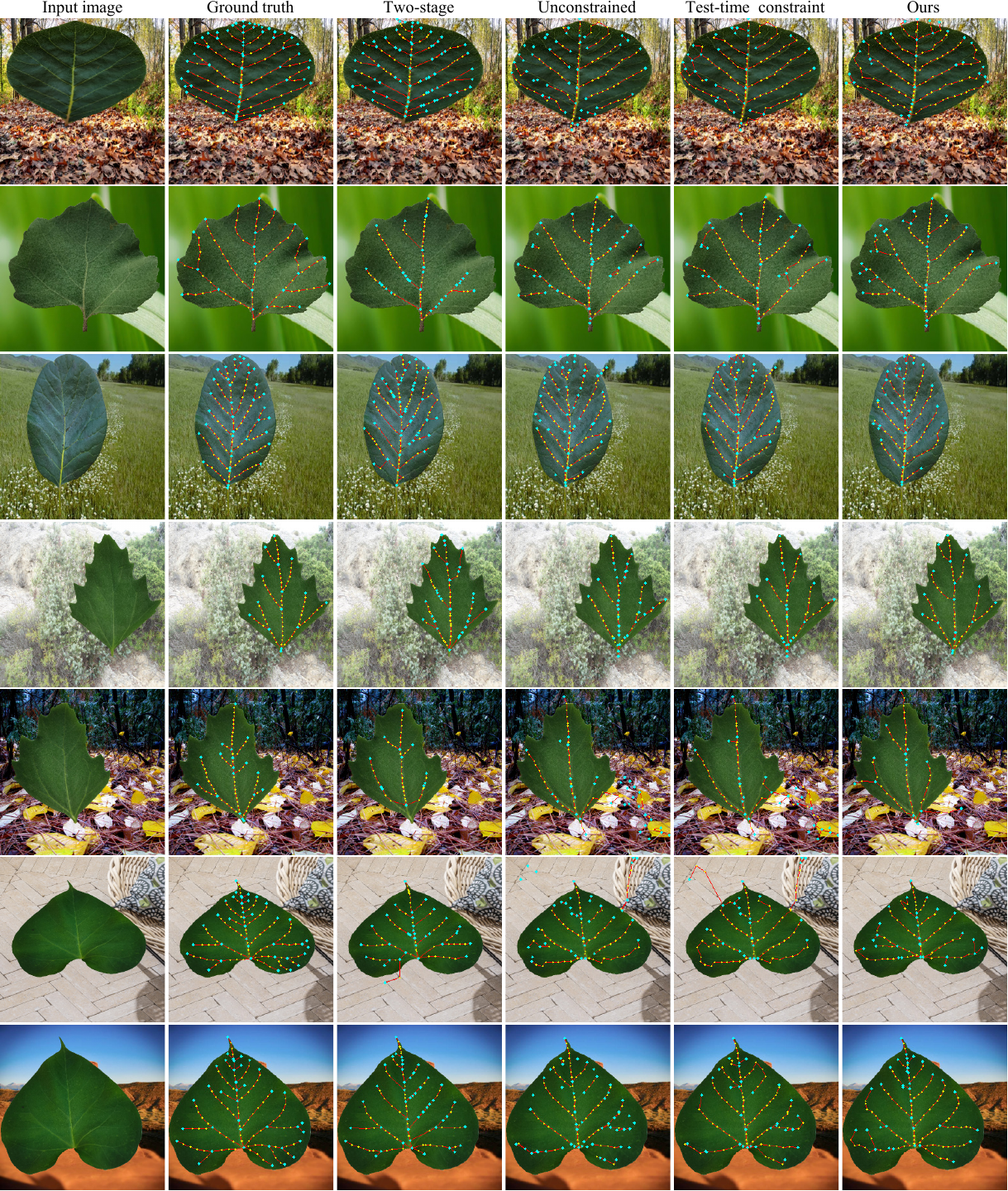}    
	\caption{Additional results for the LVD with background dataset (generalized training).}
    \label{fig:supp_result_domain_leaf_bkg}
\end{figure*}

\begin{figure*}[tp]
	\centering
  \includegraphics[width=\textwidth,height=\dimexpr\textheight-2\baselineskip\relax,keepaspectratio]{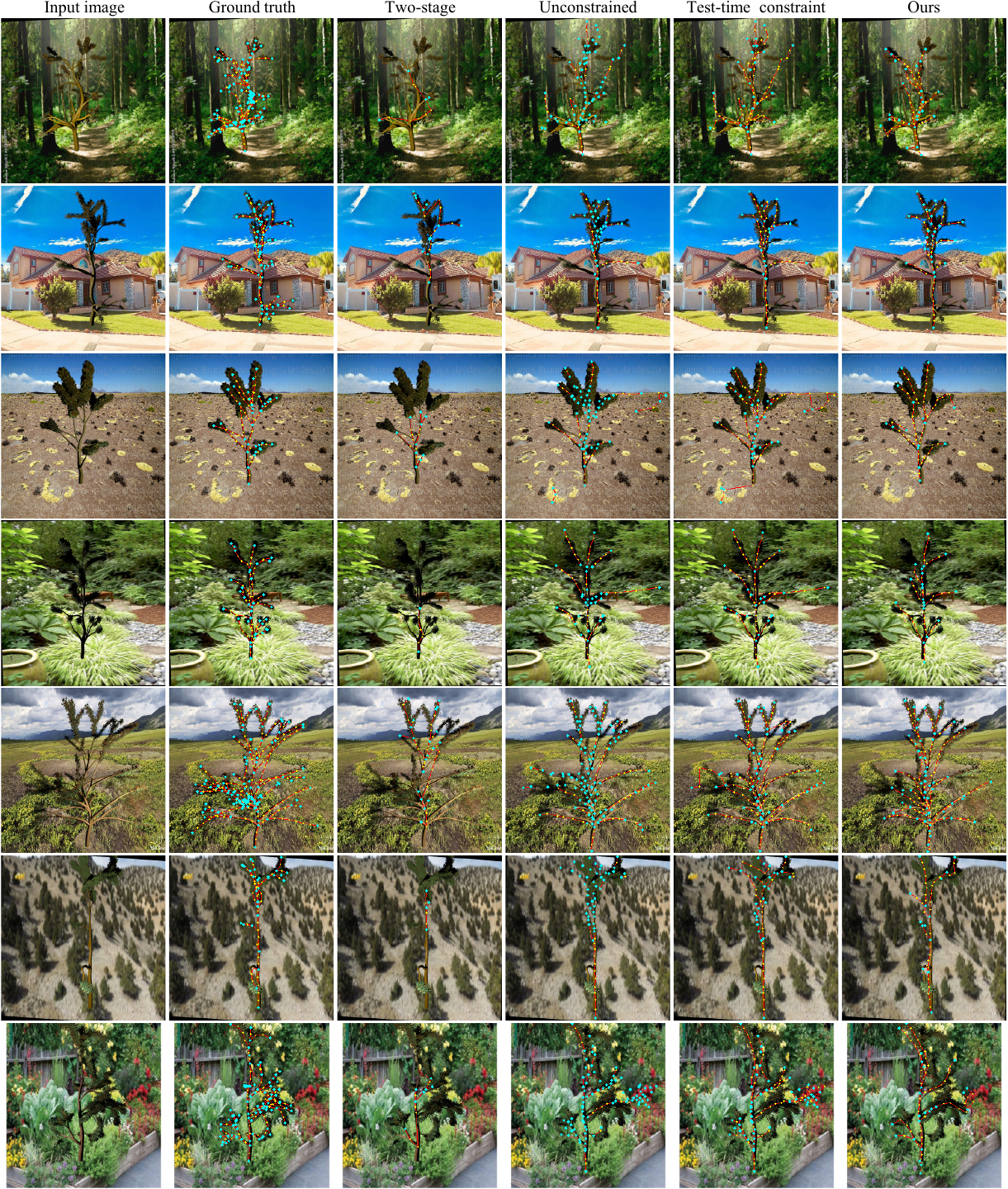}    
	\caption{Additional results for the tree synthetic dataset (generalized training).}
    \label{fig:supp_result_domain_mtree}
\end{figure*}

\begin{figure*}[tp]
	\centering
  \includegraphics[width=\textwidth,height=\dimexpr\textheight-2\baselineskip\relax,keepaspectratio]{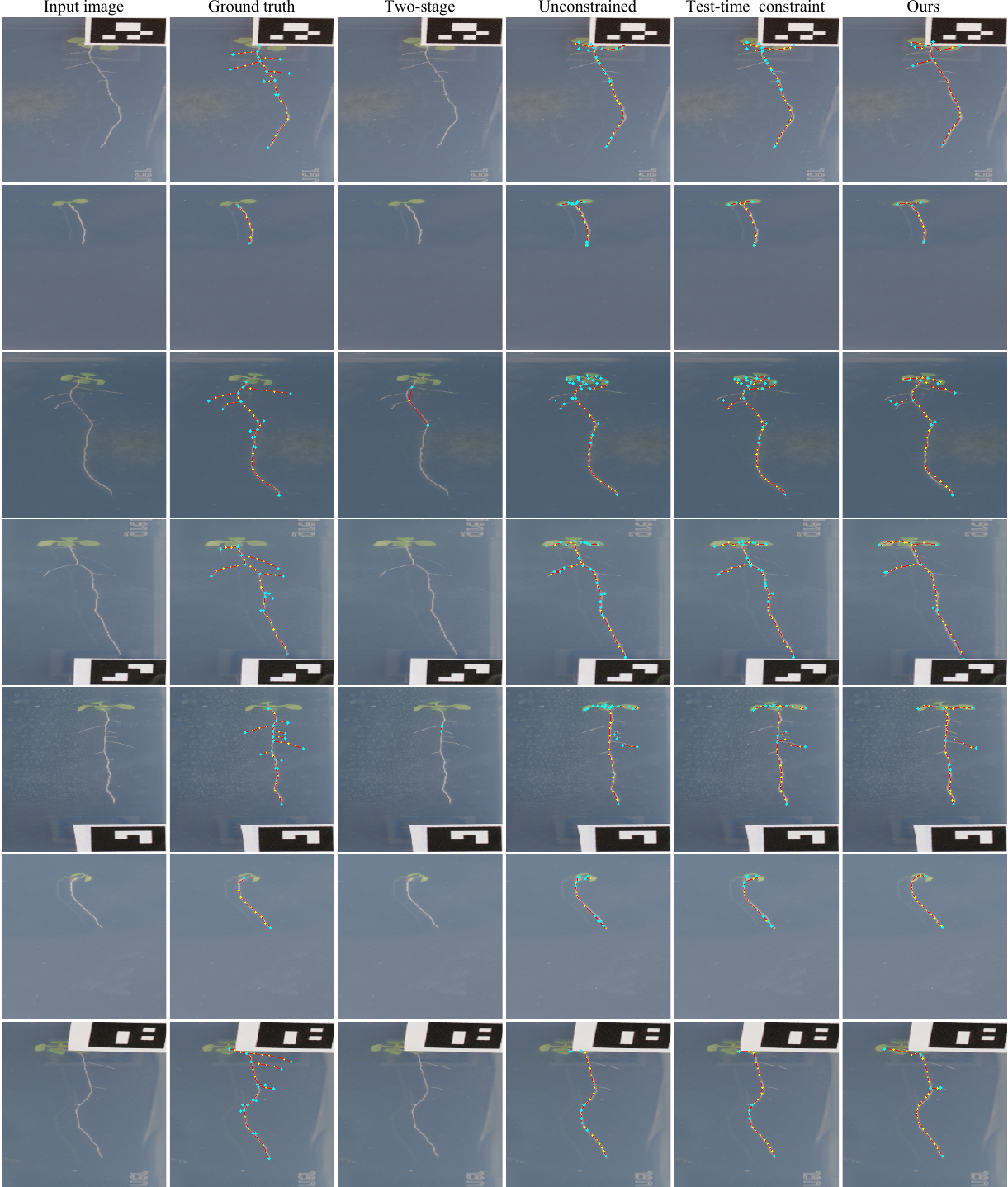}    
	\caption{Additional results for the root dataset (generalized training).}
    \label{fig:supp_result_domain_root}
\end{figure*}

\begin{figure*}[tp]
	\centering
  \includegraphics[width=\textwidth,height=\dimexpr\textheight-2\baselineskip\relax,keepaspectratio]{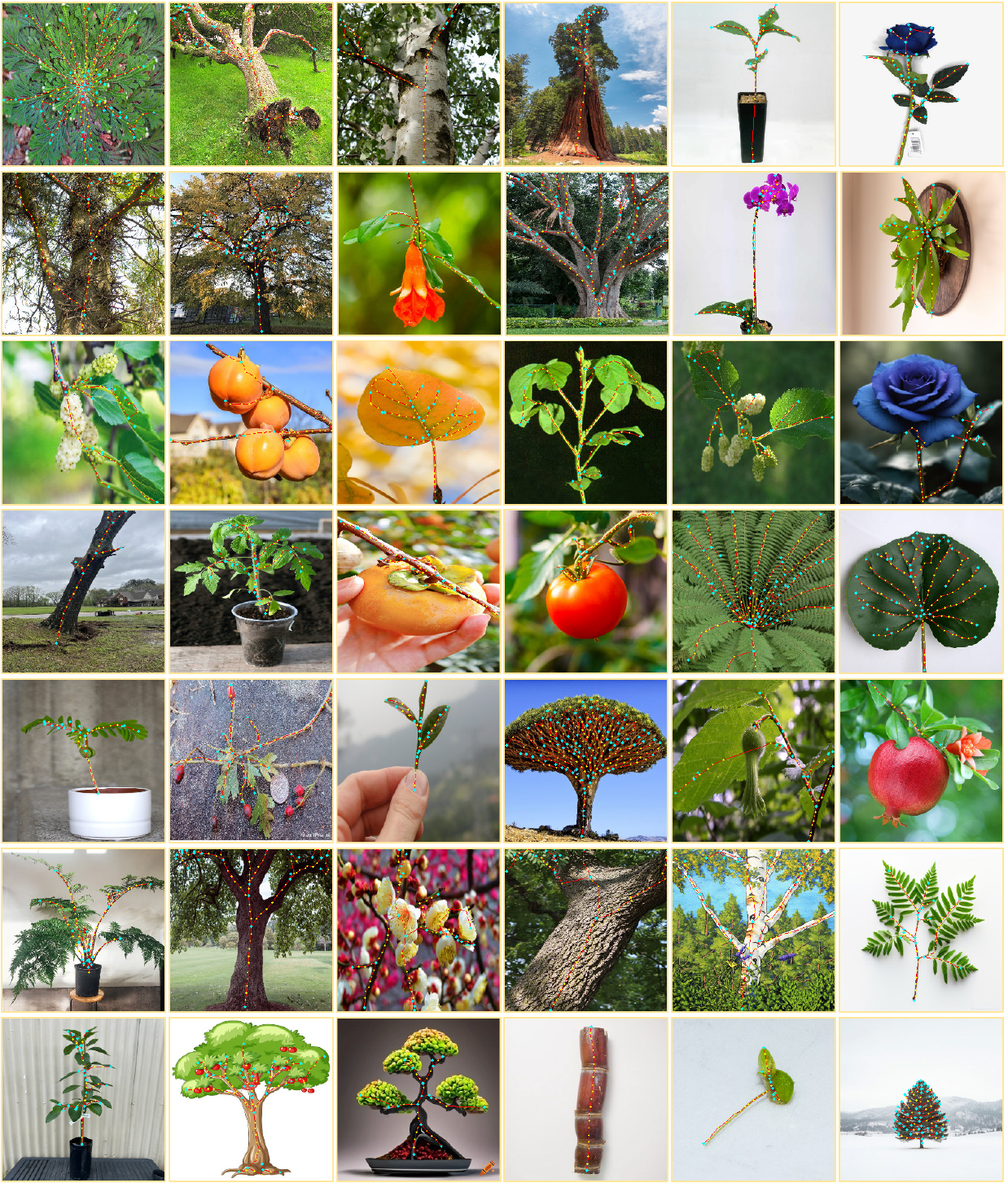}    
	\caption{Additional results for the unlabeled web test dataset (generalized training).}
    \label{fig:supp_result_domain_google1}
\end{figure*}

\begin{figure*}[tp]
	\centering
  \includegraphics[width=\textwidth,height=\dimexpr\textheight-2\baselineskip\relax,keepaspectratio]{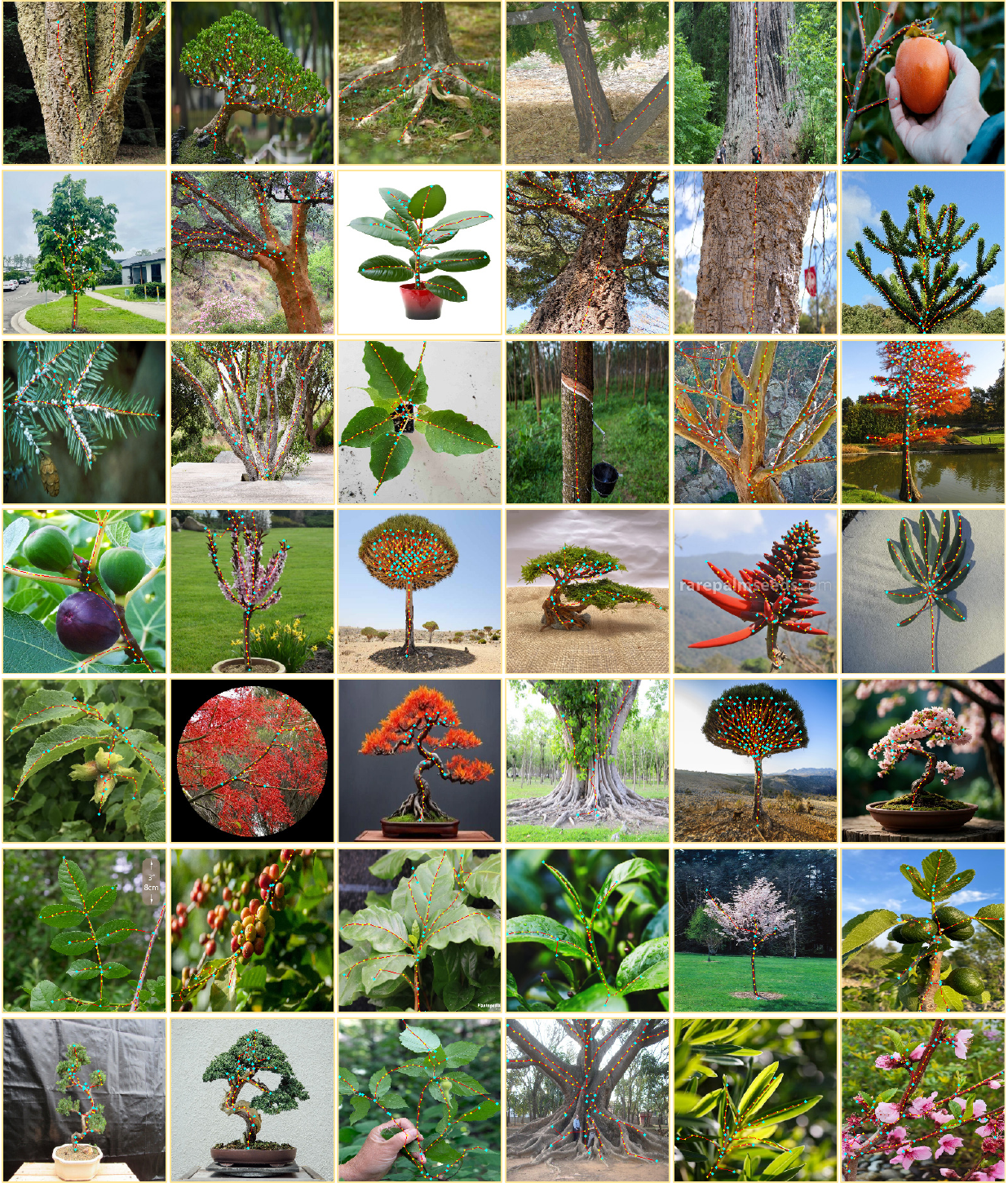}    
	\caption{Additional results for the unlabeled web test dataset (generalized training, cont'd).}
    \label{fig:supp_result_domain_google2}
\end{figure*}

We present the results of the training using generalized datasets. \Cref{fig:supp_result_domain_guyot,fig:supp_result_domain_leaf,fig:supp_result_domain_leaf_mask,fig:supp_result_domain_MIPDB,fig:supp_result_domain_toda_cam,fig:supp_result_domain_toda_world} illustrate the visual outputs for the six training datasets, showcasing the diversity and structural complexity captured by our model. \Cref{fig:supp_result_domain_lsystem,fig:supp_result_domain_leaf_bkg,fig:supp_result_domain_mtree,fig:supp_result_domain_root} display the results for the four out-of-domain test datasets, which assess the model's ability to generalize to unseen structure and contexts. Finally, \Cref{fig:supp_result_domain_google1,fig:supp_result_domain_google2} demonstrate the performance on the unlabeled web test dataset, which further evaluates the model's capability to handle out-of-training-distribution data.

These results consistently demonstrate the high-fidelity estimation of plant skeletons by our PlantPose, which generalizes effectively across diverse datasets, including real-world, synthetic, and out-of-training-distribution images. Notably, the model accurately captures structural variations and environmental contexts, highlighting its robustness and generalization capability.

\clearpage
\begin{figure}[tp]
    \includegraphics[width=\linewidth]{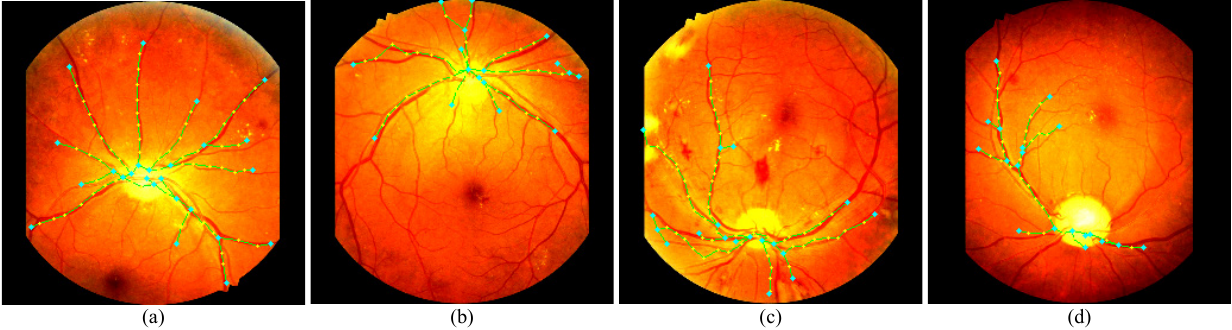}
    \caption{Qualitative results on retinal vessel images. PlantPose, trained only on plant skeletons, recovers most major vessel trees after simple preprocessing (contrast adjustment and rotation).}
    \label{fig:vessel}
\end{figure}

\subsection{Generalization beyond plants}
While the scope is beyond the plant skeleton estimation, we further assess the generalization ability of our model for images other than plants. We tested the pre-trained generalized PlantPose model on retinal vessel images (RETA~\cite{reta_data} dataset). The visual results in \fref{fig:vessel} show that the majority of vessels were reconstructed as valid tree graphs, while a few thin vessels were only partially predicted. These results show the potential usability of our method and the pre-trained model for broader domains that require tree graphs, possibly with or without fine-tuning.

\section{Analysis of Evaluation Metrics}

\begin{figure}[tp]
    \includegraphics[width=\linewidth]{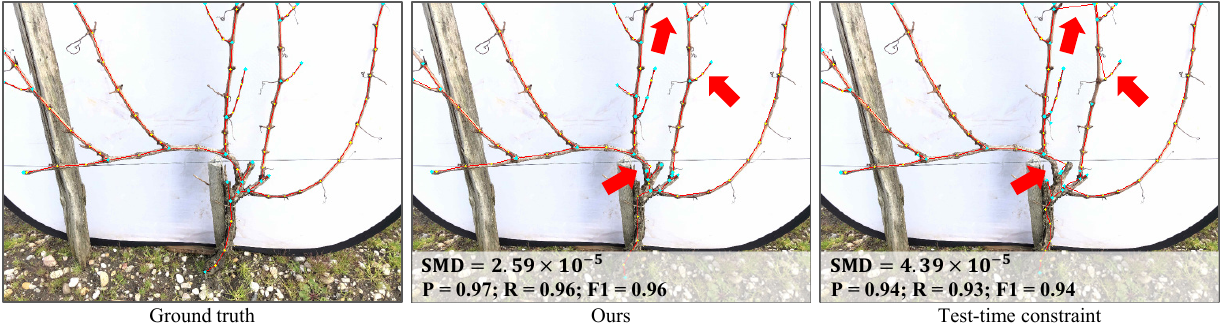}
    \caption{Comparison of our method and test-time MST constraint under generalized training. 
    Both predictions obtain similar TOPO scores ($F1=0.94$ vs.\ $0.96$), indicating comparable reachability.
    However, as highlighted by red arrows, the baseline introduces several spurious or misplaced edges, which are not penalized by TOPO but cause a substantial increase in SMD (almost $2\times$ higher). 
    This example demonstrates that SMD is a more discriminative measure of structural fidelity, capturing global geometric deviations that TOPO may overlook.}
    \label{fig:smd_vs_topo}
\end{figure}

In the experimental results, we report SMD and TOPO scores commonly used in image-to-graph generation studies. While our method achieves promising results, it is not intuitive what each metric evaluates, which has not been investigated so far. 

To briefly investigate the SMD and TOPO scores, we show an illustrative example in \fref{fig:smd_vs_topo}. The figure compares the cases where TOPO is similar, but SMD reveals large differences between our method and the test-time constraint baseline. In this case, the TOPO score overlooks the apparent mistake in the structure. On the other hand, SMD decreases approximately $40\%$ by our method, which reflects the difference in the shape (implicitly including structure and position) of the reconstructed graph.

Overall, SMD is considered the key indicator of skeleton fidelity in our setup. While TOPO provides complementary evidence, it may not always evaluate structural errors adequately. We hope that providing better evaluation metrics would be an important open problem for image-to-graph generation studies.

\section{Details of Graph Size}

\begin{figure}[tp]
    \includegraphics[width=\linewidth]{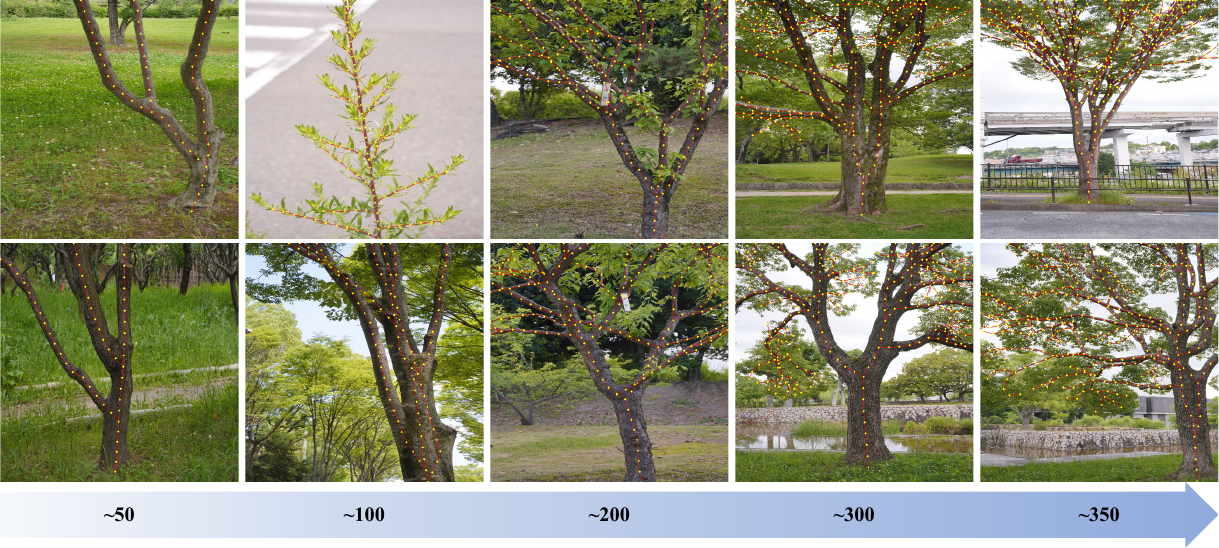}
    \caption{Illustration of scalability with respect to node count. 
    From left to right: examples with $\sim$50, $\sim$100, $\sim$200, $\sim$300, and $\sim$350 nodes.
    Simpler graphs (grass-like or single-trunk plants) have sparse nodes, while more complex trees include dense branching structures. }
    \label{fig:node_scalability}
\end{figure}

We use the maximum cap of $384$ nodes for the experiments. To further provide the justification and investigate the scalability of our method, we describe the details of how the graph sizes have been determined, and discuss the potential of extending our method for larger scales.

\paragraph{Input Resolution} 
We adopt a fixed input resolution of $256\times256$ across all 10 datasets to ensure reproducibility and training stability, while emphasizing \emph{topological fidelity} (connectivity, junctions, and endpoints) rather than fine texture details. 
This choice is consistent with prior image-to-graph works: RelationFormer~\cite{Relationformer} uses $64\times64$ for the Toulouse road dataset and $128\times128$ for the 20 US Cities dataset, RoadTracer~\cite{roadtracer} slices city-scale $4096\times4096$ maps into $256\times256$ tiles, and RNGDet~\cite{RNGDet}, RNGDet++~\cite{RNGDet++}, as well as SAM-Road++~\cite{SAM_road++} adopt $128$--$256$ pixels patches as their default input. 

\paragraph{Node Cap} 
We cap the maximum number of nodes per graph at $384$, which is not a fundamental limitation but a standardized setting under our sampling interval of $13$ pixels at $256\times256$. 
This interval was derived from the observed average inter-junction spacing in grapevine images ($\approx 12.8$ px $\approx 1/20$ of the width). 
It corresponds to a sampling interval $s \approx 13$ pixels derived from the observed mean inter-junction spacing in grapevine ($\approx 12.8$ px $\approx 1/20$ of the width).
Under uniform density, the capacity per tile scales as
\begin{equation}
    N_{max}(H, W, s) \approx \left\lfloor \frac{H}{s} \right\rfloor \cdot \left\lfloor \frac{W}{s} \right\rfloor,
\end{equation}
resulting in $s{=}13$ at $H{=}W{=}256$, giving $\approx 3.88 \times 10^2$ (rounded to $384$ for scheduling).

\paragraph{Scalability} 
\Fref{fig:node_scalability} shows how graphs of different sizes appear in our datasets. 
From left to right, the examples become progressively more complex: Starting with simple grass-like plants or single trunks ($\sim$50 nodes), then adding visible full plant structures ($\sim$100–200 nodes), and finally reaching fully developed trees with dense branches ($\sim$300–350 nodes). 
This trend reflects our dataset statistics: The average node count in the \textit{Self-captured dataset} is $156$, while in the \textit{Web-sourced dataset} is $96$.
When node count grows, the cost of pairwise edge classification increases quadratically ($O(N^2)$), leading to higher memory and runtime. 
Future work may address this by (i) pruning candidate edges to local neighborhoods, (ii) tiling and stitching subgraphs, or (iii) using memory-efficient attention mechanisms (\eg, FlashAttention~\cite{flashattention}) to handle denser graphs without altering our formulation.

\end{appendices}

\clearpage
{\small
\bibliography{egbib}
}

\end{document}